\documentclass[nohyperref]{article}

\usepackage{microtype}
\usepackage{graphicx}
\usepackage{subfigure}
\usepackage{booktabs} 

\usepackage{hyperref}

\usepackage[accepted]{icml2022}

\usepackage{amsmath}
\usepackage{amssymb}
\usepackage{mathtools}
\usepackage{amsthm}

\usepackage[capitalize,noabbrev]{cleveref}

\theoremstyle{plain}
\newtheorem{theorem}{Theorem}[section]

\theoremstyle{definition}
\newtheorem{definition}[theorem]{Definition}

\theoremstyle{remark}

\usepackage[textsize=tiny]{todonotes}
\usepackage[utf8]{inputenc} 
\usepackage[T1]{fontenc} 
\usepackage{hyperref} 
\usepackage{url} 
\usepackage{booktabs} 
\usepackage{amsfonts} 
\usepackage{nicefrac} 
\usepackage{microtype} 
\usepackage{xcolor}         

\usepackage{placeins}
\usepackage{graphicx}
\usepackage{subfigure}
\usepackage{bm}
\usepackage{booktabs}
\usepackage{threeparttable}
\usepackage{multirow}
\usepackage{amsmath}
\usepackage{wrapfig}
\usepackage{enumitem}

\icmltitlerunning{Supervised Off-Policy Ranking}

\begin{document}

\twocolumn[
\icmltitle{Supervised Off-Policy Ranking}




\begin{icmlauthorlist}
\icmlauthor{Yue Jin}{1}
\icmlauthor{Yue Zhang}{2}
\icmlauthor{Tao Qin}{3}
\icmlauthor{Xudong Zhang}{1}
\icmlauthor{Jian Yuan}{1}
\icmlauthor{Houqiang Li}{2}
\icmlauthor{Tie-Yan Liu}{3}
\end{icmlauthorlist}

\icmlaffiliation{1}{Department of Electronic Engineering, Tsinghua University, Beijing, China}
\icmlaffiliation{2}{Department of Electronic Engineering and Information Science, University of Science and Technology of China, Hefei, China}
\icmlaffiliation{3}{Microsoft Research Asia, Beijing, China}

\icmlcorrespondingauthor{Tao Qin}{taoqin@microsoft.com}

\icmlkeywords{Machine Learning, ICML}

\vskip 0.3in
]



\printAffiliationsAndNotice{}  

\begin{abstract}
Off-policy evaluation (OPE) is to evaluate a target policy with data generated by other policies. Most previous OPE methods focus on precisely estimating the true performance of a policy. We observe that in many applications, (1) the end goal of OPE is to compare two or multiple candidate policies and choose a good one, which is a much simpler task than precisely evaluating their true performance; and (2) there are usually multiple policies that have been deployed to serve users in real-world systems and thus the true performance of these policies can be known. Inspired by the two observations, in this work, we study a new problem, supervised off-policy ranking (SOPR), which aims to rank a set of target policies based on supervised learning by leveraging off-policy data and policies with known performance. We propose a method to solve SOPR, which learns a policy scoring model by minimizing a ranking loss of the training policies rather than estimating the precise policy performance. The scoring model in our method, a hierarchical Transformer based model, maps a set of state-action pairs to a score, where the state of each pair comes from the off-policy data and the action is taken by a target policy on the state in an offline manner. Extensive experiments on public datasets show that our method outperforms baseline methods in terms of rank correlation, regret value, and stability. Our code is publicly available at GitHub~\footnote{\url{https://github.com/SOPR-T/SOPR-T}}.
\end{abstract}

\section{Introduction}
Off-policy evaluation (OPE), which aims to estimate the online/true performance~\footnote{We use online performance and true performance interchangeably in this paper.} of a policy using only pre-collected historical data generated by other policies, is critical to many real-world applications, in which evaluating or deploying a poorly performed policy online might be prohibitively expensive (e.g., in trading, advertising, traffic control) or even dangerous (e.g., in robotics, autonomous vehicles, drug trials).

Existing OPE methods can be roughly categorized into three classes: distribution correction based methods~\cite{thomas2015high, liu2018breaking,hanna2019importance,xie2019towards, nachum2019dualdice, zhang2020gendice, nachum2020reinforcement, yang2020off, kostrikov2020statistical}, model estimation based methods~\cite{mannor2004bias, thomas2016data, hanna2017bootstrapping, zhang2021autoregressive}, and Q-estimation based methods~\cite{Le2019,munos2016safe, harutyunyan2016q, precup2000eligibility, farajtabar2018more}. While those methods are based on different assumptions and with different formulations, most of them (1) focus on precisely estimating the expected return of a target policy using pre-collected historical data, and (2) perform unsupervised estimation without directly leveraging the online performance of previously deployed policies.

We notice that there are some mismatches between the settings of those OPE methods and the OPE problem in real-world applications. First, in many applications, we do not need to estimate the exact true performance of a target policy. Instead, what we need is to compare the true performance of a set of candidate policies and identify the best one which will then be deployed into real-world systems. That is, correct ranking of policies rather than precise return estimation is the end goal of off-policy evaluation.
Second, in real-world applications, we usually know the true performance of some polices that have been deployed into real-world systems and interacted with real-world users. Such information is not well exploited in today’s OPE methods.

Based on these observations, in this work, we define a new problem, supervised off-policy ranking (SOPR), which is different from previous OPE in two aspects. First, 
SOPR aims to correctly rank new policies, rather than accurately estimate their expected returns.
Second, SOPR leverages the true performance of a set of pre-collected policies in addition to off-policy data. 

A straightforward way to rank policies is to first use a scoring model to score each policy and then rank them based on their scores. Following this idea, we propose a supervised learning-based method to train a scoring model. The scoring model takes logged states in historical data and the actions taken by a policy offline as raw features. Considering that there might be plenty of data, we design a hierarchical Transformer encoder (TE) based model to learn representations at different levels.
Specifically, we first adopt k-means clustering to cluster similar states.
Within each cluster, a low-level TE is employed to encode each state-action pair.
At high level, another TE is employed to encode each cluster.
Following each TE, an average pooling operator is adopted to aggregate information in the corresponding level.
Finally, a multi-layer perceptron (MLP) maps the overall representation to a score. 
The hierarchical TE and MLP are jointly trained to correctly rank the pre-collected policies with known performance. 
Our method is named as SOPR-T, where ``T'' stands for our TE-based model.

We evaluate SOPR-T on public datasets for offline reinforcement learning (RL)~\cite{fu2020d4rl}. Experiments on multiple tasks and different datasets demonstrate that SOPR-T outperforms representative baseline methods in terms of both rank correlation, regret value, and stability.

The main contributions of this work are as follows:
\begin{itemize}[leftmargin=*]
\item We point out that OPE should focus on policy ranking rather than exactly estimating the returns of policies, which simplifies the OPE problem.
\item According to our knowledge, we are the first to introduce supervised learning into OPE. We take a preliminary step towards supervised OPE in this work and define the problem of supervised off-policy ranking. We hope that our work can inspire more solid works along this direction.
\item We propose a hierarchical Transformer encoder based model for policy ranking. Experiments demonstrate the effectiveness and stability of our method. Our code and data have been released at GitHub.
\end{itemize}

\section{Supervised Off-Policy Evaluation/Ranking}
In this section, we first give some notations and then formally describe the problems of supervised off-policy evaluation and supervised off-policy ranking.

We consider OPE in a Markov Decision Process (MDP) defined by a tuple $(S, A, T, R, \gamma)$.
$S$ and $A$ denote state and action space. $T(s'|s,a)$ and $R(s,a)$ are state transition distribution and reward function.
$\gamma \in (0,1]$ is a discount factor.
The expected return of a policy $\pi$ is defined as
$V(\pi) = \mathbb{E}[\sum\nolimits_{t=0}^{T} {{\gamma ^t}} {r_t}]$, where $a_t \sim \pi(\cdot|s_t)$, $r_t \sim R(s_t,a_t)$, and $T$ is the time horizon of the MDP.

The goal of OPE is to evaluate a policy without running it in the environment.
Traditional OPE methods estimate the expected return of a policy leveraging a pre-collected dataset
$\mathcal{D}=\{\tau_i\}_{i=1}^N$ composed of $N$ trajectories generated by some other policies (usually called behavioral policy), where $\tau_i = s_0^i, a_0^i, r_0^i, \cdots, s_T^i, a_T^i, r_T^i$.

Apart from the pre-collected dataset, we can also collect some previously deployed policies whose true performance is available.
For instance, in many real-world applications, such as advertising and recommendation systems, we can get the true performance of previously deployed polices by observing and counting user clicks.
Thus, the true performance of these policies is available.
Clearly, those information is helpful for OPE, but was ignored in previous OPE methods. 
In this work, we define a new kind of OPE problem, the supervised OPE problem, as below.
\begin{definition}
Supervised off-policy evaluation: given a pre-collected dataset $\mathcal{D}=\{\tau_i\}_{i=1}^N$ with $N$ trajectories and $M$ pre-collected policies $\{\pi_i\}_{i=1}^M$ with known performance, estimate the performance of a target policy or a set of target policies without interacting with the environment.
\end{definition}

Comparing with previously studied OPE problems, our new problem has more available information, the set of pre-collected policies $\{\pi_i\}_{i=1}^M$ with known performance, which can serve as supervised signal while we learn an OPE model. We name this problem ``supervised" OPE as we have label information (i.e., the true performance) for the policies available in training. In contrast, previous OPE problems can be deemed unsupervised learning problem, since they do not directly learn from the online performance of pre-collected policies. 

In addition, in real-world applications we often need to compare a set of candidate policies and choose a good one from them.
Thus, what we really need is the correct ordering of a set of policies, rather than the exact performance value of each policy. 
Formally, we define a variant of the supervised OPE problem, which is called supervised off-policy ranking.
 \begin{definition}
Supervised off-policy ranking: given a pre-collected dataset $\mathcal{D}=\{\tau_i\}_{i=1}^N$ with $N$ trajectories and $M$ pre-collected policies $\{\pi_i\}_{i=1}^M$ together with their performance ranking, rank a set of target policies without interacting with the environment.
\end{definition}
It is not difficult to see that policy ranking is relatively easier than policy evaluation, since accurate evaluation inherently implies correct ranking, but correct ranking does not need accurate evaluation. Considering the practical value and simplicity of policy ranking, we focus on supervised off-policy ranking (SOPR) in the following parts of this paper.

\section{Our Method}
We propose a method for SOPR in this section, which involves training a scoring model to score and rank a set of  policies. We start with introducing policy representation (Section ~\ref{sec:fea}), and then loss function design and training pipeline (Section ~\ref{sec:algo}).

\subsection{Policy Representation}\label{sec:fea}
Policies can be of very different forms: a policy can be a set of designed rules, a linear function of states, or deep neural networks, e.g., convolutional neural networks, recurrent neural networks, and attention-based networks.
To map a policy to a score, the first question to answer is how to represent different forms of policies.

To handle all kinds of policies, we should leverage their shared points rather than differences. Obviously, we cannot use the internal parameters of policies, because different policies may have different numbers of parameters and some policies do not have parameters. Fortunately, all the policies for the same task or game share the same input space, state space $S$, and the same output space, action space
$A$. Therefore, we propose to learn policy representations upon state-action pairs.

Let $\mathcal{D}_s =\{s_i\}_{i=1}^{N_s}$ denote a set of states included in the pre-collected dataset $\mathcal{D}$, where $N_s$ is the number of these states. For a policy $\pi$ to be ranked, let it take actions for all the states in $\mathcal{D}_s$ and obtain a dataset $\mathcal{D}_\pi=\{(s_i, a_i^{\pi})\}_{i=1}^{N_s}$, where $s_i \in \mathcal{D}_s, a_i^{\pi} \sim \pi(\cdot|s_i) $. Now, any policy $\pi$ can be represented by a dataset $\mathcal{D}_\pi$ in the same format.
\footnote{For the simplicity of description, we consider deterministic policies here. For stochastic policies, we can use the distribution or the statistics over actions for a state $s$ to create the dataset $\mathcal{D}_\pi$.}

Now the question becomes how to map a set of state-action pairs/points\footnote{A state-action pair can be deemed a point in a high-dim space.}, $\mathcal{D}_\pi=\{(s_i, a_i^{\pi})\}_{i=1}^{N_s}$, to a score that indicates the performance of policy $\pi$. Following~\cite{kool2018attention,nazari2018reinforcement, bello2016neural, xin2020multi, vinyals2015pointer}, we design a Transformer encoder (TE)~\cite{vaswani2017attention} based model to encode all those points. We (1) first project each point into an embedding vector, (2) use a few self-attention layers to get high-level representations of those points, (3) aggregate them by average pooling to get a representation of the dataset (and thus the policy), and (4) finally adopt a linear projection to map the vector to a score.

\begin{figure}[!t]
    \centering
    \includegraphics[width=1
    \columnwidth]{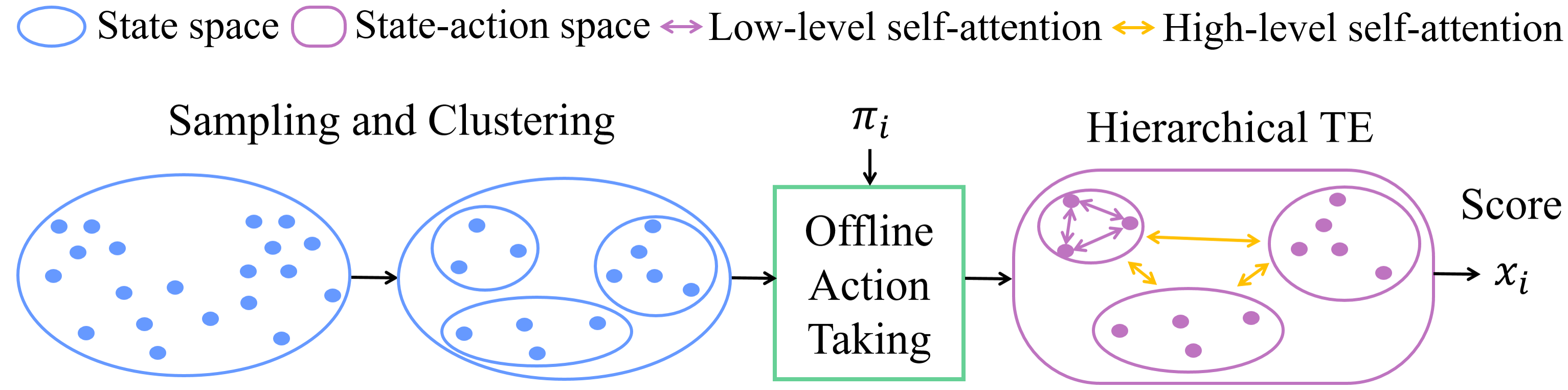}
    \caption{Illustration of SOPR-T method.}
    \label{pipline}
\end{figure}

\begin{figure}[!t]
    \centering
    \includegraphics[width=0.8
    \columnwidth]{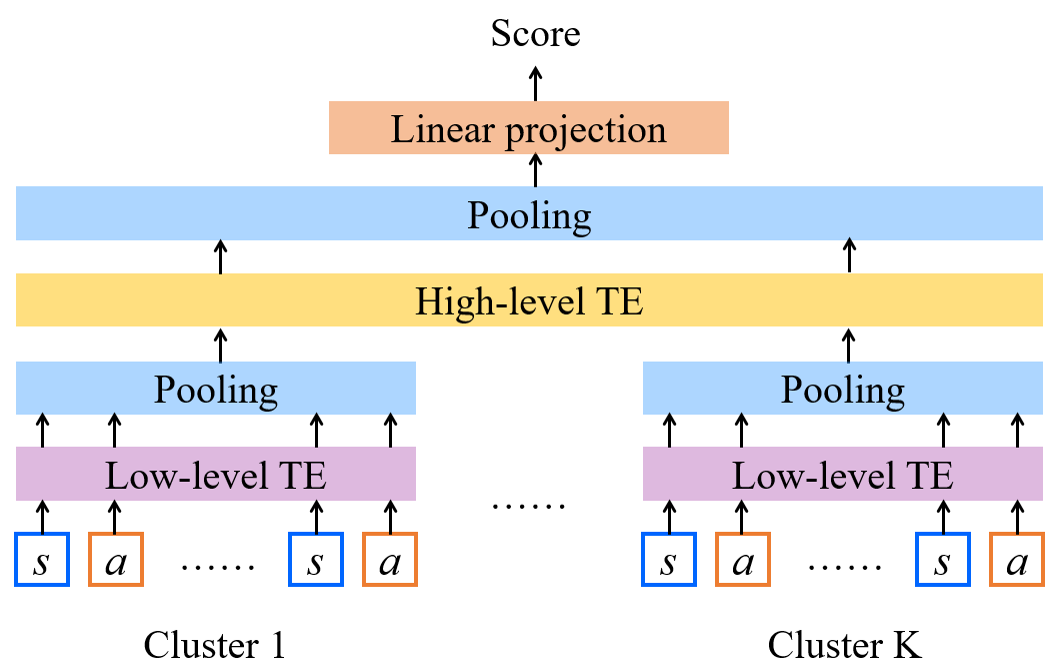}
    \caption{Architecture of hierarchical Transformer encoder based scoring model.}
    \label{structure}
\end{figure}

A computational challenge is that there could be millions of states in the pre-collected historical data. It is impossible for Transformer to handle such a large scale of data. Our solution is to down sample the states and encode the state-action pairs in a hierarchical way as follows:
\begin{enumerate}[leftmargin=*]
    \item Randomly sample a subset $D_s$ of states from $\mathcal{D}_s$ and cluster them into $K$ clusters by k-means clustering.
    \item Let a policy $\pi$ take actions over the states in $D_s$ and obtain a set of state-action pairs.
    \item Use a low-level TE to encode all the state-action pairs in a cluster and get a vector representation for each cluster by average pooling.
    \item Use a high-level TE to encode all the clusters and get a vector representation for the policy by average pooling.
    \item Use a linear projection to map the vector in the above step to a score.
\end{enumerate}
Since our scoring function is based on Transformer, we name our method SOPR-T.
Figure~\ref{pipline} illustrates the pipeline of SOPR-T.
Figure~\ref{structure} shows the architecture of our proposed hierarchical TE based policy scoring model.

\subsection{Pairwise Loss and Learning Algorithm} \label{sec:algo}
As aforementioned, the goal of SOPR is to rank a set of policies correctly. We adopt a pairwise ranking loss following~\cite{burges2005learning}: 
\begin{equation}
\begin{aligned}
    &l(\pi_i,\pi_j;\theta) = - \left[y_{ij} \log(\frac{1}{1+e^{-(f(\pi_i,D_s;\theta) - f(\pi_j,D_s;\theta))}}) \right.\\
    & \left. + (1-y_{ij}) \log(1-\frac{1}{1+e^{-(f(\pi_i,D_s;\theta) - f(\pi_j,D_s;\theta))}}) \right],
\end{aligned}
\end{equation}
where $f(\pi,D_s;\theta)$ denotes the scoring function with parameter $\theta$, $\pi_i$ and $\pi_j$ are two policies with a performance ranking $y_{ij}$. $y_{ij}=0$ means $\pi_i$ is worse than $\pi_j$,  $y_{ij}=1$ means $\pi_i$ is better than $\pi_j$, and $y_{ij}=0.5$ means the two policies perform similarly.
To minimize the loss, the scoring function needs to correctly rank two policies, i.e., consistent with the ranking of their true performance.

The complete learning algorithm of SOPR-T is shown in Algorithm \ref{algo}.  Note that, in training, we sample multiple subsets of states in a manner similar to mini-batch training. In inference, we sample multiple subsets and compute a score using the well-trained scoring function over each subset. The final score of a test policy is the average score over those subsets.

\begin{algorithm}[!t]
\caption{ Training procedure of SOPR-T}
\begin{algorithmic}
\raggedright
    \STATE \textbf{Input}: State dataset $\mathcal{D}_s$, training policy set $\{\pi_i\}_{i=1}^M$, and pairwise ranking label $\{y_{ij}\}_{i,j=1, i \neq j}^M$.
    \STATE \textbf{Initialize}: Scoring model $f(\pi,D_s;\theta)$.
    \FOR {iteration $t$ in $\{1, \cdots, T\}$}
        \STATE Sample a subset $D_s^t$ of states from $\mathcal{D}_s$.\\
        \STATE Compute the score $f(\pi_i,D_s^t;\theta)$ as described in Section~\ref{sec:fea} for all the $M$ policies.
        \STATE Perform gradient update to minimize the ranking loss $l(\pi_i,\pi_j;\theta)$ for each pair of policy $\pi_i$ and $\pi_j$.
    \ENDFOR
\end{algorithmic}
\label{algo}
\end{algorithm}

\subsection{Discussions} \label{sec:dis}

One may notice that we use only the states in the pre-collected historical data $\mathcal{D}$, while most previous OPE methods use both the states and immediate rewards $R(s,a)$ in $\mathcal{D}$. There are several considerations for only using states. First, a new policy to be evaluated usually takes actions different from the historical data.
For a state with a new action $a'$, $R(s,a')$ is unknown if such a state-action pair is not contained in the historical data.  
Thus, we do not directly use immediate rewards in this work. 
Second, in the setting of SOPR, we have a set of training policies with known performance. 
The performance information of those policies is more reliable and can be used as a direct signal for supervised learning, as compared with immediate rewards. This is because the performance information comes from online interactions with real-world systems and directly indicates the performance of a policy. 
Third, consider the following formulation of the expected return of a policy,
\begin{equation} \label{ return with stationary distribution}
    V(\pi) = \mathbb{E} [d^\pi(s,a) R(s,a)],
\end{equation}
where $d^\pi(s,a)$ denotes the stationary distribution of state-action pairs under $\pi$.
Note that, only $d^\pi(s,a)$ depends on policy $\pi$, while $R(s,a)$ is the same across all policies for a task. 
Therefore, to rank different policies for a task, the more important part is $d^\pi(s,a)$, rather than the immediate rewards.
Of course, if we can work out a good solution to accurately predict and well leverage immediate rewards, this will further improve the accuracy of supervised off-policy evaluation/ranking. Such a problem is beyond the scope of this paper. We leave it to future work.

Although we focus on supervised off-policy ranking in this work, our proposed scoring model can be easily applied to supervised OPE. 
For this purpose, we need the true performance of training polices, and only performance ranking is not enough. 
Given the true performance of training policies, we can train the scoring model by minimizing the gap between the true performance and the predicted performance of a policy.

\section{Experiments}
We compare SOPR-T with different kinds of baseline OPE algorithms on various tasks, including evaluating policies learned by different algorithms, in different games, and with different settings of datasets. 
Below we first introduce experimental settings, including tasks, training/validation/test sets, baselines, and evaluation metrics. Then, we present experimental results regarding performance comparison among different algorithms as well as further studies of our algorithm. 

\subsection{Experimental Settings}
\paragraph{Tasks}
We evaluate SOPR-T and baseline OPE algorithms on D4RL datasets~\footnote{\url{https://github.com/rail-berkeley/d4rl}} \cite{fu2020d4rl} which are widely used in offline RL studies.
Overall, we use 12 tasks covering three Mujoco games, i.e.,  Hopper-v2, HalfCheetah-v2, and Walker2d-v2, and four types of off-policy datasets for each game, i.e., expert, medium, medium-replay (m-replay for short), and full-replay (f-replay for short).
The expert and medium datasets are collected by a single expert and medium policy, respectively.
The two policies are trained with Soft Actor-Critic (SAC) algorithm \cite{haarnoja2018soft} online, and the medium policy only achieves 1/3 performance of the expert policy.
The full-replay and medium-replay datasets contain all the data collected during the training of the expert and medium policy, respectively.

\paragraph{Training Set and Validation Set}
Training set and validation set consist of policies and their rank labels.
The policies are collected during online SAC training.
For each game, we collect 50 policy snapshots during the training process and get their performance by online evaluation using 100 Monte-Carlo rollouts in the real environment.
After the training process is finished, we randomly select 30 policies to form training policy set and another 10 policies to form validation policy set.
The remaining 10 policies are used to form a test policy set. We will provide a detailed description of the test policy set later.
Note that, in the training phase of SOPR-T, only the rank of the policies are used as labels.

\paragraph{Test Set}
In each task, we use two kinds of test policy sets to simulate two kinds of OPE cases.

In Test Set I, we investigate the capability of SOPR-T and baseline OPE algorithms to rank and select good policies in offline RL settings.
Specifically, Test Set I is composed of policies collected by running 3 popular offline RL algorithms, BEAR \cite{Kumar2019}, CQL \cite{kumar2020conservative}, and CRR \cite{wang2020critic}.
The implementation of the three algorithms is based on a public codebase \footnote{\url{https://github.com/takuseno/d3rlpy}}.
Note that, these algorithms adopt different network architectures.
Thus, the types of policy models are not all the same.
For each task, we run each algorithm until convergence and collect policy snapshots during training.
To get the ground truth rank labels of the performance of these policies, we also evaluate the performance of each policy using 100 Monte-Carlo rollouts in the real environment, and use the rank of the performance as rank labels.
Then, we mix these policies and select 10 policies whose performance are evenly spaced in the performance range of all policies.
In this way, the selected polices have diverse performance.
In addition, mixing policies generated by different algorithms to form a test policy set is aligned with the practical cases where the sources of policies are various and unknown.

In Test Set II, we investigate the capability of SOPR-T to rank policies that are approximately within the distribution of training policies.
Because in practice, such as production development and update, the updated policies and the previously evaluated policies usually have many common properties.
Thus, it can be assumed that the policies to be evaluated and the policies that have been evaluated are in two similar distributions.
To simulate this case in experiments, we leverage the 10 policies mentioned in the description regarding the training policy set to form Test Set II.
Because the 10 policies and the training policies are uniformly sampled from the same policy set, they are approximately within the same distribution.
As these policies are learned online, we name Test Set II online learned policies.


\paragraph{Baselines}
We compare SOPR-T with four representative baseline OPE algorithms. 

1. Fitted Q-Evaluation (FQE) \cite{Le2019, kostrikov2020statistical}, which is a Q-estimation based OPE method. It learns a neural network to approximate the Q-function of the target policy by leveraging Bellman expectation backup operator \cite{sutton2018reinforcement} iteratively based on the off-policy data.

2. Model-based estimation (MB) \cite{kostrikov2020statistical, fu2021benchmarks}. It estimates the environment model including a state transition distribution and a reward function.
The expected return of the target policy is estimated using the returns of Monte-Carlo rollouts in the modeled environment.

3. Weighted importance sampling (IW) \cite{kostrikov2020statistical, fu2021benchmarks}, which is a distribution correction-based method. It leverages importance sampling to correct the weight of the reward regarding the data from the behavior data distribution to the target data distribution, and adopts weight normalization.

4. DualDICE \cite{nachum2019dualdice}, which is also based on distribution correction, but without directly using importance weights. It learns to estimate the state-action stationary distribution correction. 

We leverage a popular implementation~\footnote{\url{https://github.com/google-research/google-research/tree/master/policy_eval}} of these algorithms.

\paragraph{Evaluation Metrics}
We evaluate SOPR-T and baseline OPE algorithms with two metrics, i.e., Spearman’s rank correlation coefficient and normalized Regret@k, to reflect their performance of ranking candidate policies, which is aligned with a related work \cite{Paine2020}.
Specifically, Spearman's rank correlation is the Pearson correlation between the ground truth rank sequence and the evaluated rank sequence of the candidate policies.
Normalized Regret@k is the normalized performance gap between the truly best policy of all candidate policies and the truly best policy of the ranked top k policies, i.e., $\text{Regret@k} = (V_{\text{max}}-V_{\text{max\_topk}})/(V_{\text{max}}-V_{\text{min}})$, where $V_{\text{max}}$ and $V_{\text{min}}$ are the ground truth performance of the truly best and the truly worst policy of all candidate policies, respectively. $V_{\text{max\_topk}}$ is the ground truth performance of the truly best policy of the ranked top k policies.
We use $k=3$ in our experiments.

\subsection{Performance on Offline Learned Policies} \label{exp_testset_1}
We first evaluate SOPR-T and baseline OPE algorithms on Test Set I, i.e., the offline learned policies. 
In the evaluation process, we use 3 random seeds for each experiment.

\begin{figure}[!t]
\centering
\subfigure[]
{
    \begin{minipage}[t]{0.47\linewidth}
	\centering
    \includegraphics[width = \linewidth]{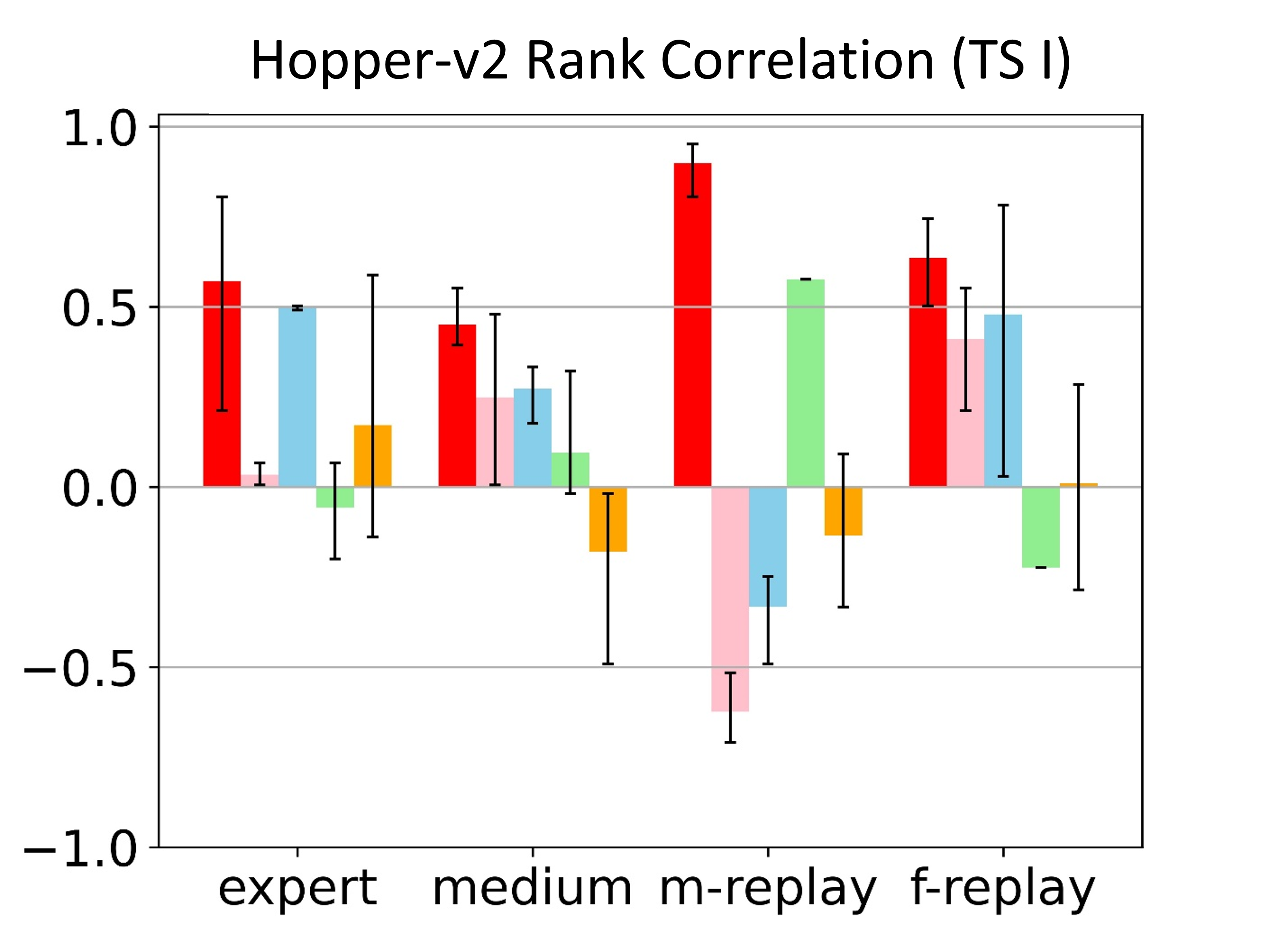} \label{hopper_RC}
    \end{minipage}
}
\subfigure[]
{
    \begin{minipage}[t]{0.47\linewidth}
	\centering
    \includegraphics[width = \linewidth]{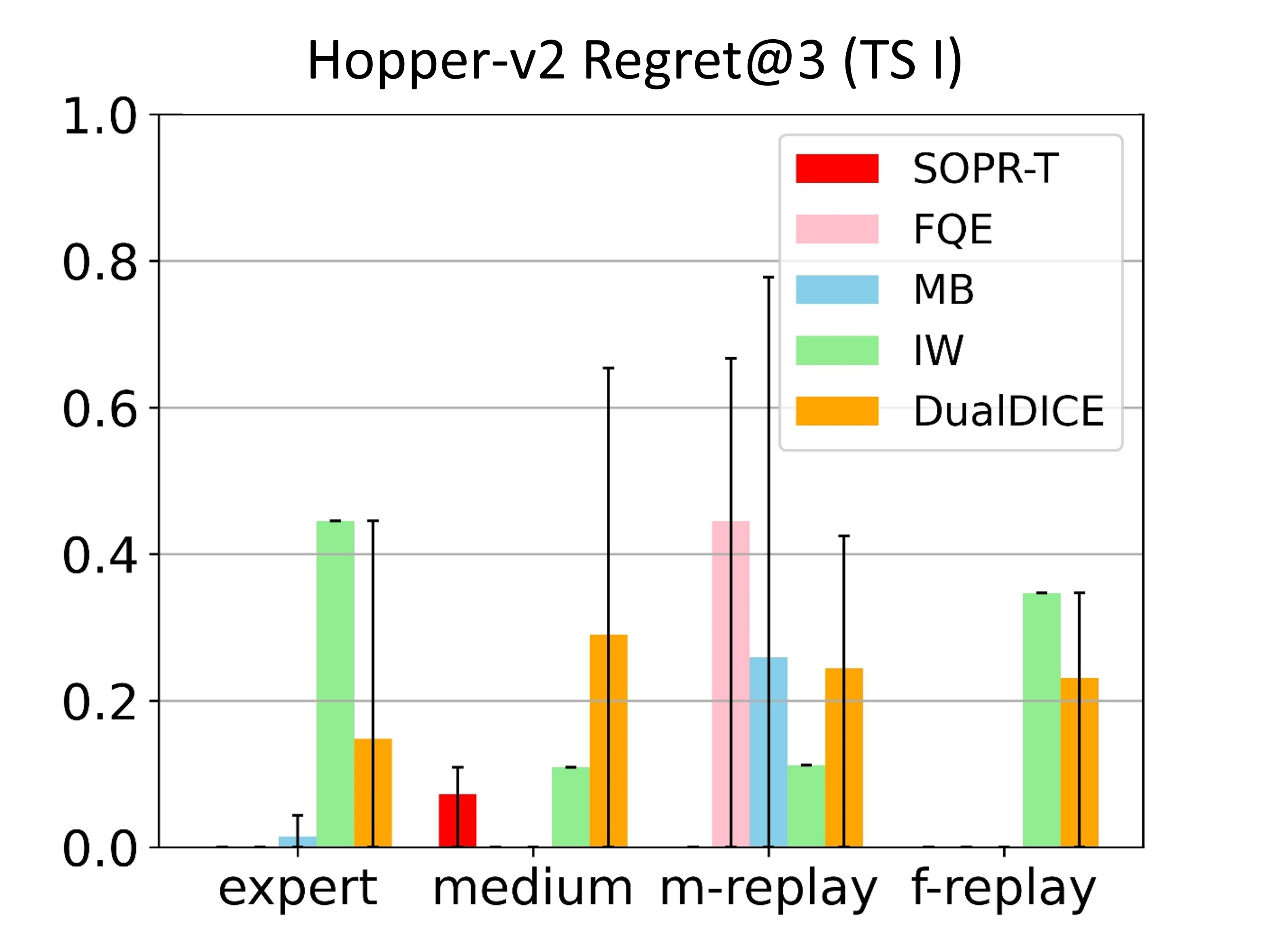}\label{hopper_Rt}
    \end{minipage}
}

\subfigure[]
{
    \begin{minipage}[t]{0.47\linewidth}
	\centering
    \includegraphics[width = \linewidth]{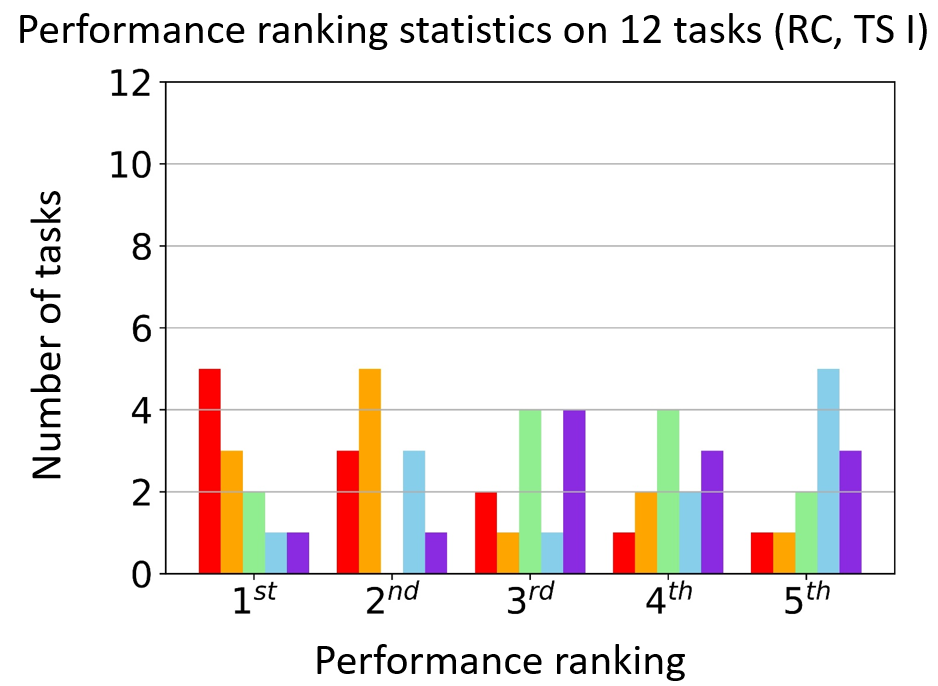}\label{all_RC}
    \end{minipage}
}
\subfigure[]
{
    \begin{minipage}[t]{0.455\linewidth}
	\centering
    \includegraphics[width = \linewidth]{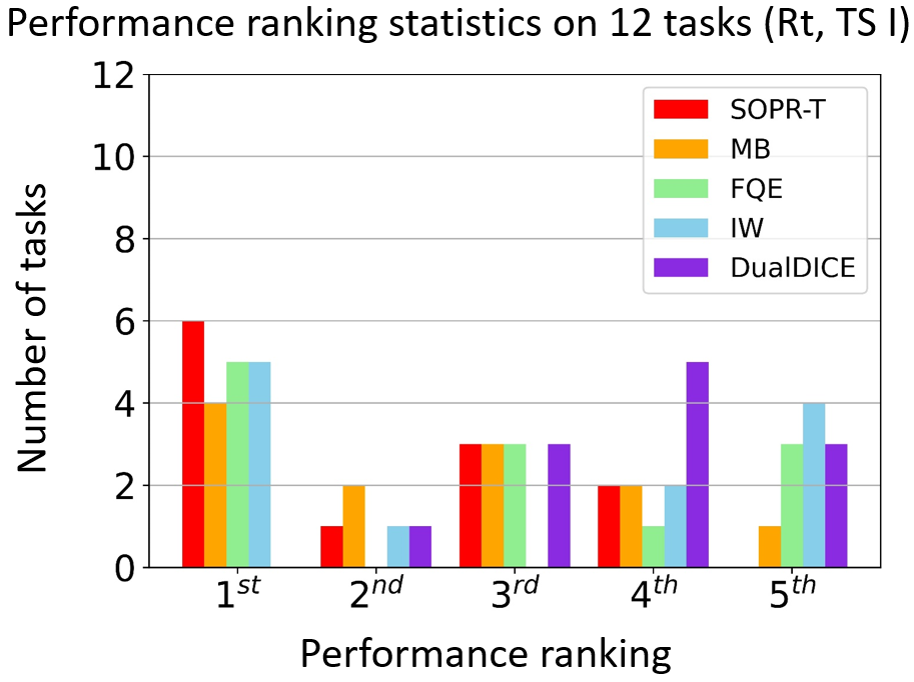}\label{all_Rt}
    \end{minipage}
}

\subfigure[]
{
    \begin{minipage}[t]{0.47\linewidth}
	\centering
    \includegraphics[width = \linewidth]{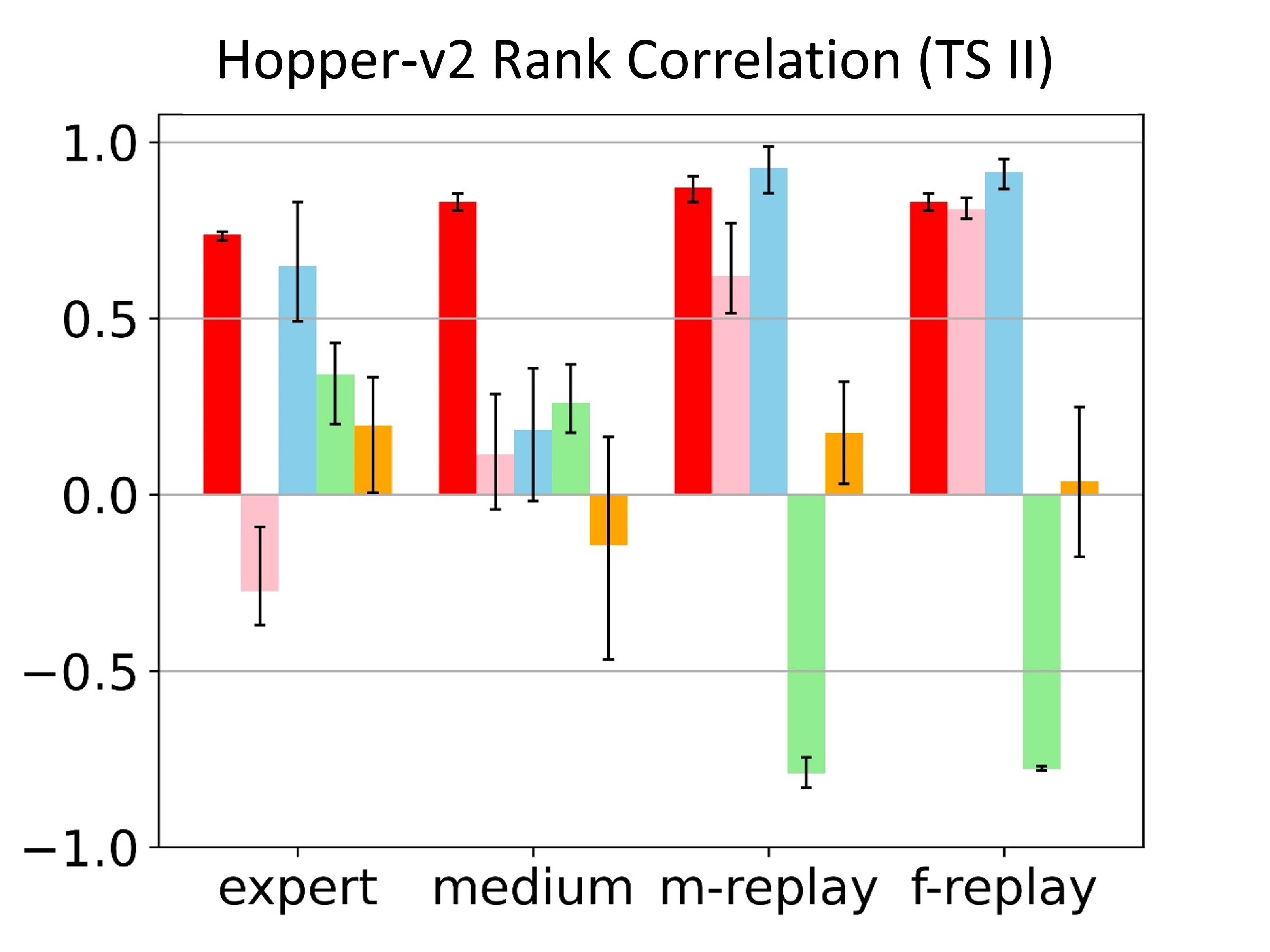}
    \end{minipage}
}
\subfigure[]
{
    \begin{minipage}[t]{0.47\linewidth}
	\centering
    \includegraphics[width = \linewidth]{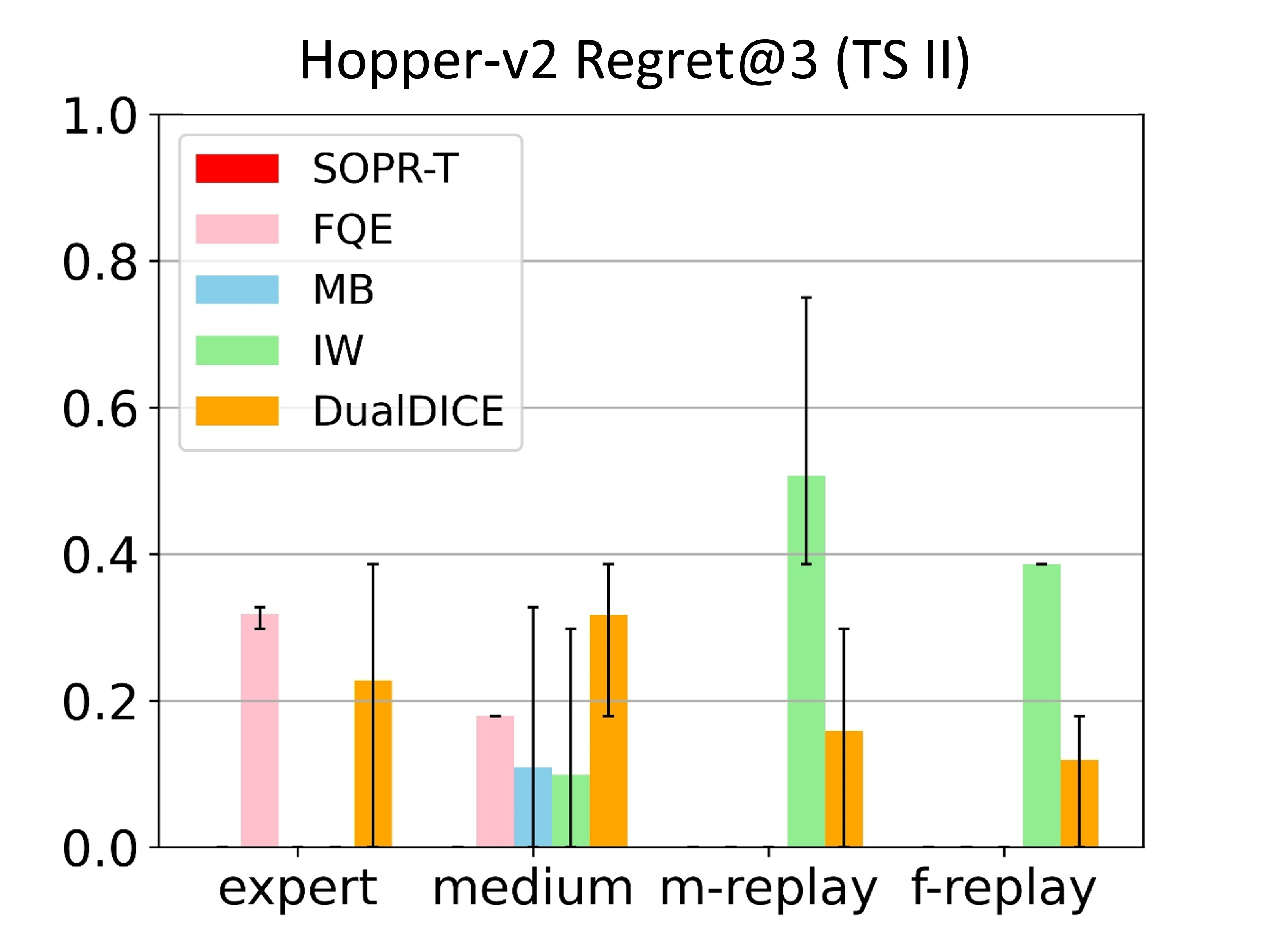}
    \end{minipage}
}

\subfigure[]
{
    \begin{minipage}[t]{0.47\linewidth}
	\centering
    \includegraphics[width = \linewidth]{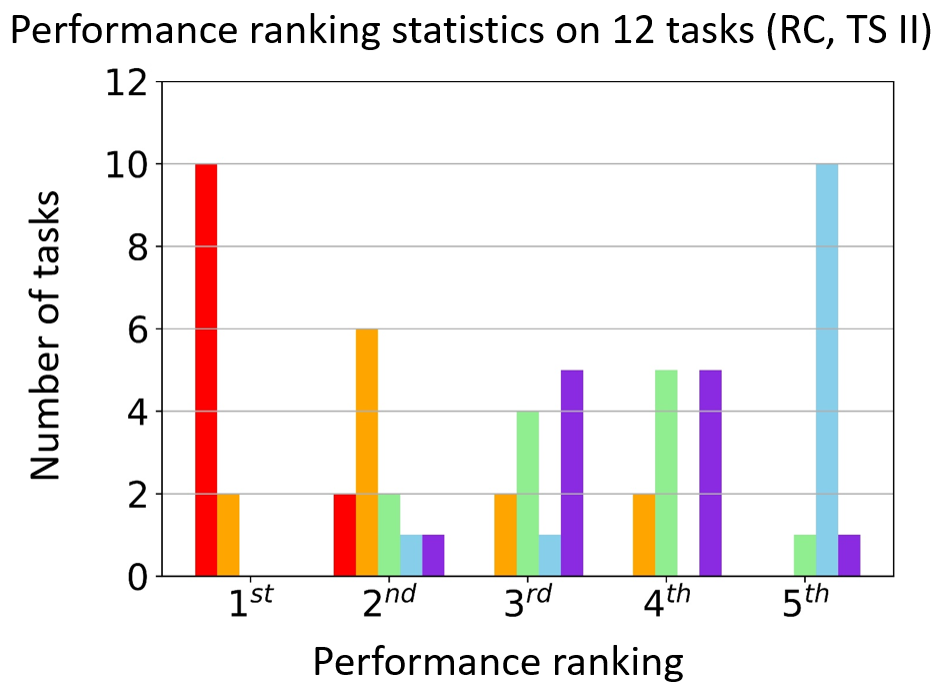}
    \end{minipage}
}
\subfigure[]
{
    \begin{minipage}[t]{0.465\linewidth}
	\centering
    \includegraphics[width = \linewidth]{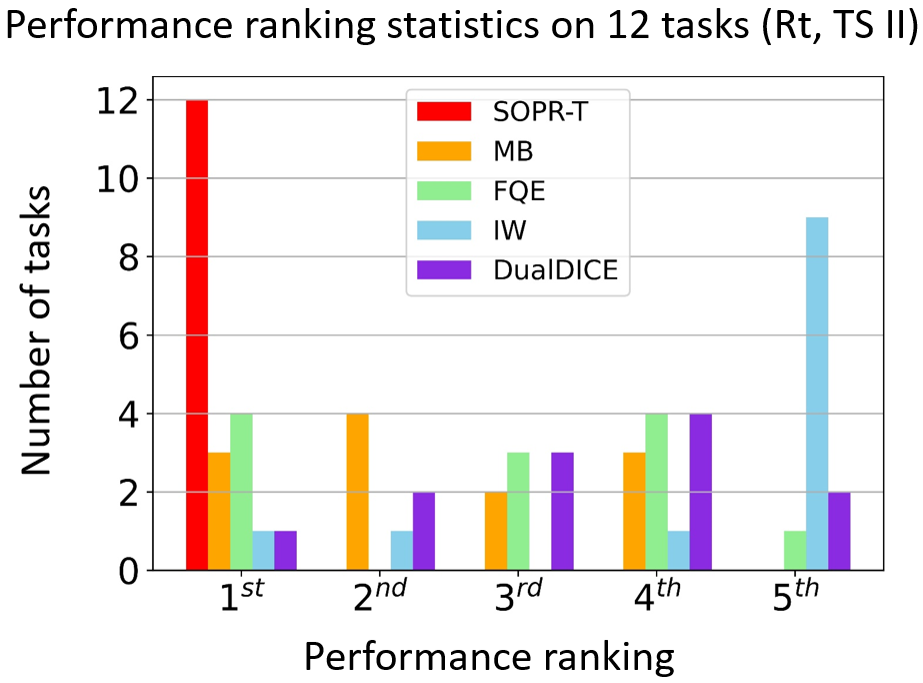}
    \end{minipage}
}
\caption {
Performance comparison.
Left: rank correlation (abbreviate to RC). Right: regret@3 (abbreviate to Rt).
Top 2 rows: Test Set I (offline learned policies). 
Bottom 2 rows: Test Set II (online learned policies).
Row 1 and Row 3: performance comparison on 4 types of datasets of the Hopper game. 
Row 2 and Row 4: performance ranking statistics of 5 algorithms on 12 tasks.
}
\label{result_1}
\end{figure}

Due to space limitation, we only present the results on the Hopper game (top row of Figure~\ref{result_1}), and an overall performance ranking of 5 algorithms on 12 tasks (second row of Figure~\ref{result_1}) here. Results on other games can be found in Appendix \ref{A.3}.
As can be seen from the results shown in Figure~\ref{hopper_RC} and Figure~\ref{hopper_Rt}, SOPR-T achieves higher rank correlation coefficient and smaller regret value than baseline algorithms, which means SOPR-T can rank different policies with higher accuracy and also figure out good policies from the candidate policies.
In addition, SOPR-T performs the most stably, which does not have negative rank correlation results in all the tasks, whereas each of the baseline OPE algorithms has one or more negative rank correlation results.
Though in Walker2d and Halfcheetah (shown in Appendix \ref{A.3}), SOPR-T does not hold consistent superiority, it still performs the most stably.

Figure~\ref{all_RC} and Figure~\ref{all_Rt} present the overall performance ranking statistics of five algorithms (SOPR-T and four OPE baselines) on 12 tasks.
The results in Figure~\ref{all_RC} indicate that SOPR-T has the top performance of rank correlation on 5 tasks. In Figure~\ref{all_Rt}, SOPR-T has the top performance of regret value on 6 tasks. Among all the algorithms, the number of tasks on which SOPR-T achieves the top performance in terms of both rank correlation and regret value is the highest. The results demonstrate the overall advantage of SOPR-T.

\subsection{Performance on Online Learned Policies}\label{exp_testset_2}
Then, we evaluate SOPR-T and OPE baselines on Test Set II.
Because these policies are collected during online training rather than offline trained with the datasets, they are irrelevant to the datasets, which is aligned with many practical cases where the policies to be evaluated are not designed or learned based on the dataset.

We also run SOPR-T and each baseline OPE algorithm with 3 different seeds. Results on the Hopper game are shown in the third row of Figure~\ref{result_1} and other games in Appendix \ref{A.4}.
The results demonstrate that SOPR-T outperforms baseline OPE algorithms dramatically on almost all the tasks.
Performance ranking statistics of all algorithms on 12 tasks are shown in the bottom row of Figure~\ref{result_1}.
As can be seen from the results, SOPR-T achieves the top performance of rank correlation on 10 tasks and the top performance of regret value on all the 12 tasks.

\begin{figure}[!t]
    \centering
    \includegraphics[width=0.5
    \columnwidth]{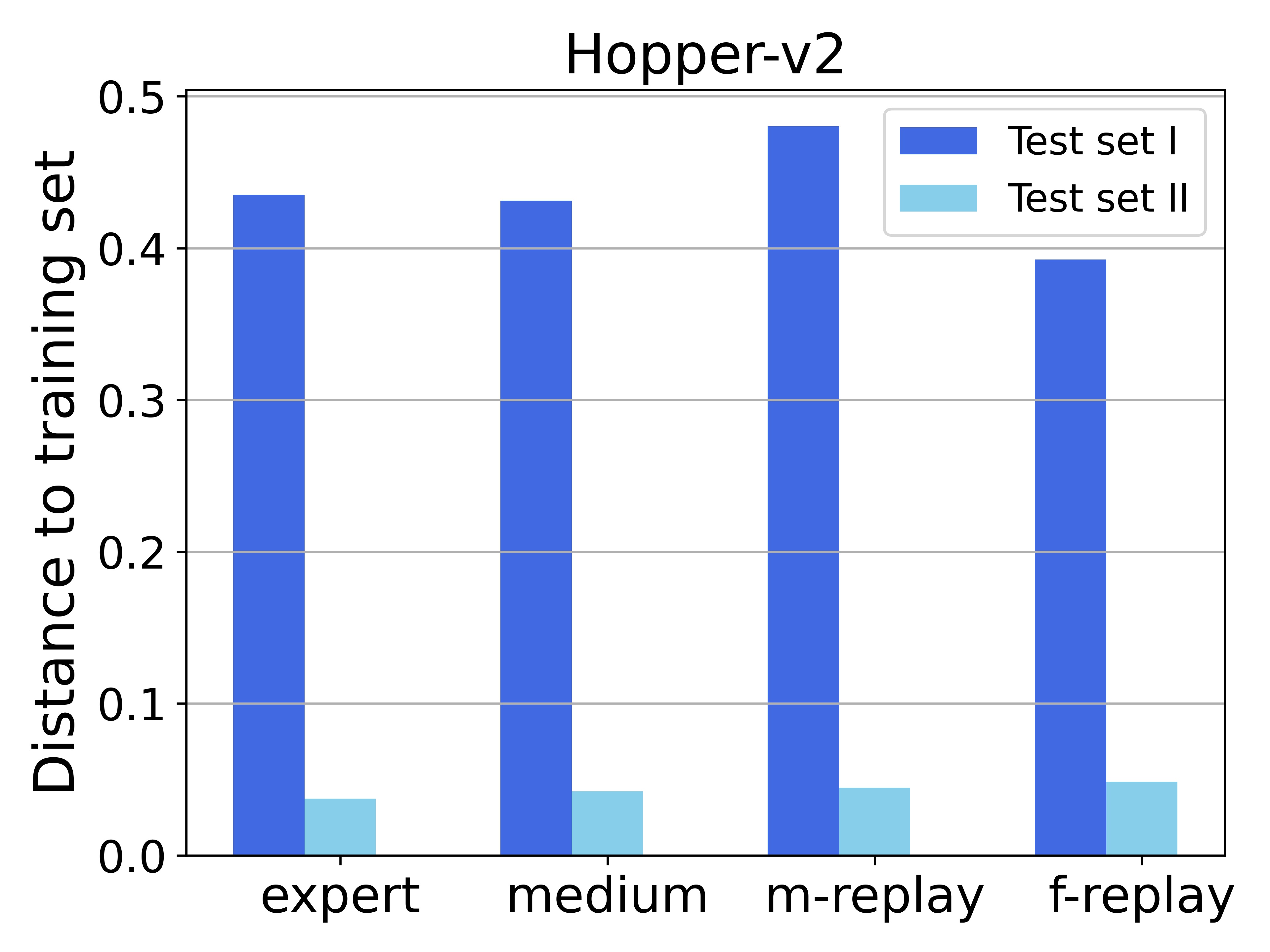}
    \caption{Distance between training and test policy sets.}
\label{mse}
\end{figure}

Compared with the performance of SOPR-T on ranking the offline learned policies, the performance on ranking the online learned policies improves.
We consider that the performance of SOPR-T is affected by the distribution difference between the training policy set and test policy set. In this respect, we measure the distribution distance between the two policy sets by:
\begin{equation}\label{eq_distance}
\begin{aligned}
    & D(\left.\{\pi_i^{\text{train}}\}_{i=1}^{M_{\text{train}}} , \{\pi_j^{\text{test}}\}_{j=1}^{M_{\text{test}}}\right|\mathcal{D}_s) \\
    &= \frac{1}{M_{\text{test}}}\sum_{j=1}^{M_{\text{test}}} \min_i (\frac{1}{N_s} \sum_{n=1}^{N_s} ||a_n^{\pi_j^{\text{test}}} - a_n^{\pi_i^{\text{train}}} ||_2^2 ) , 
\end{aligned}
\end{equation}
where $N_s$ is the number of states in the dataset, $a_n^{\pi} \sim \pi(\cdot|s_n), s_n \in \mathcal{D}_s$, $M_{\text{train}}$ and $M_{\text{test}}$ are the number of policies in the training set and test set, respectively.
Distance results are shown in Figure~\ref{mse} and Figure~\ref{5_2_supp_2} in Appendix \ref{A.4}.
As can be seen from the results,
the distance between Test Set I and the training policy set is much larger than the distance between Test Set II and the training policy set.


\subsection{Further Studies}\label{exp_further}
In this part, we further conduct a set of experiments to better understand our algorithm.
\paragraph{Effect of Data Size}
We investigate the sensitivity of SOPR-T and baseline OPE algorithms to the size of dataset.
We set the number of data (states used in SOPR-T, tuples used in baseline OPE algorithms) as 4k, 8k, 16k, and 32k.
Note that, in the original datasets, the size of data is not identical among different tasks, but all of them contain more than 200k data.

We sample different amounts of data from the original dataset in trajectory form.
For the expert and medium datasets (composed of data collected by a single policy), trajectories are sampled randomly.
For the medium-replay dataset (composed of all data during training SAC in sequence), we fetch data sequentially.
Note that, in this experiment, we use three kinds of datasets, i.e., expert, medium, and medium-replay, as we suppose the first part of data in the full-replay dataset is similar to the medium-replay dataset. 
We still test three games as before, and thus there are 9 tasks (different games or different datasets) in total in this part of experiments.

\begin{figure}[!h]
\centering
\subfigure[]
{
    \begin{minipage}[t]{0.47\linewidth}
	\centering
    \includegraphics[width = \linewidth]{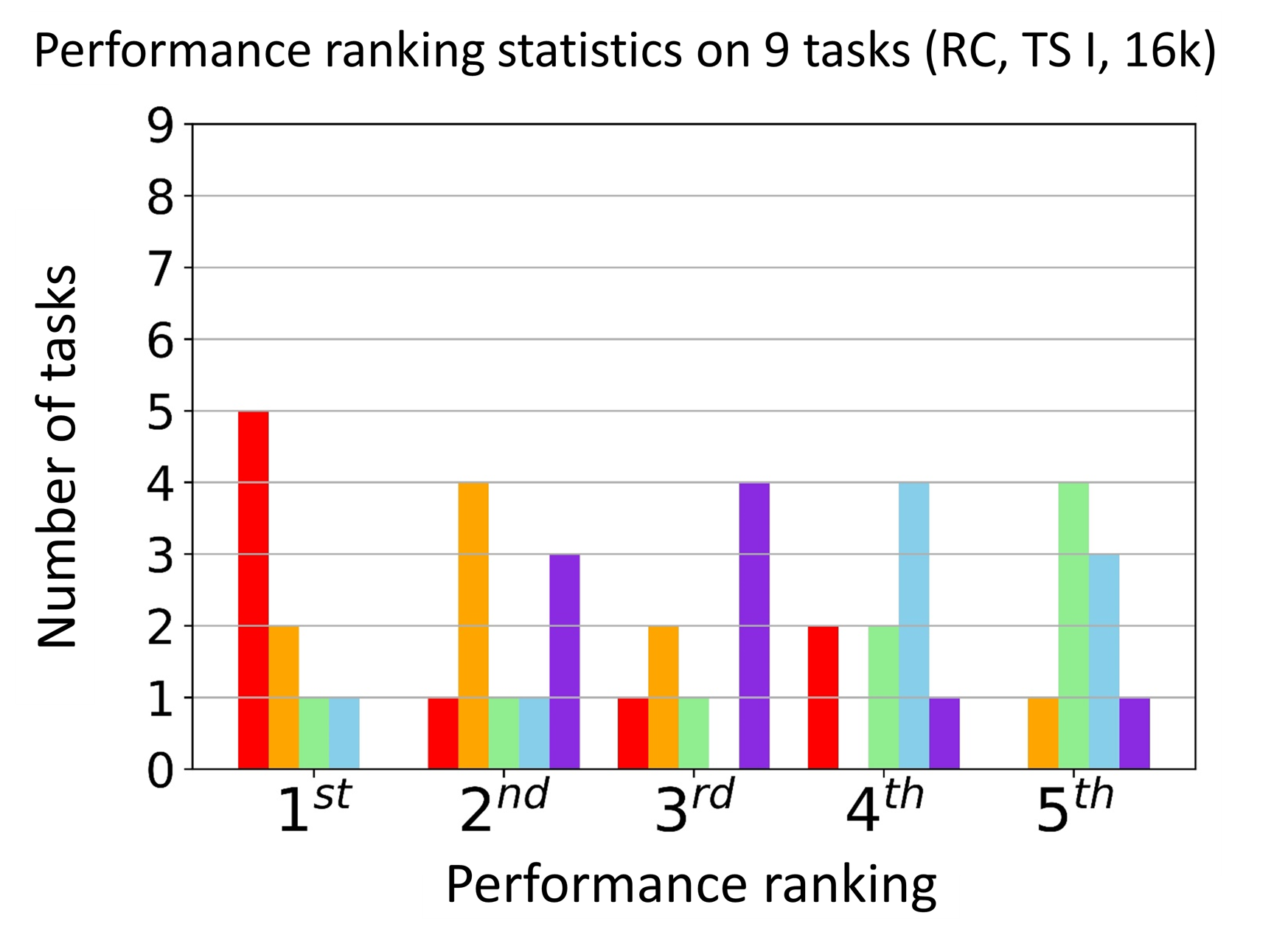}
    \end{minipage}
}
\subfigure[]
{
    \begin{minipage}[t]{0.47\linewidth}
	\centering
    \includegraphics[width = \linewidth]{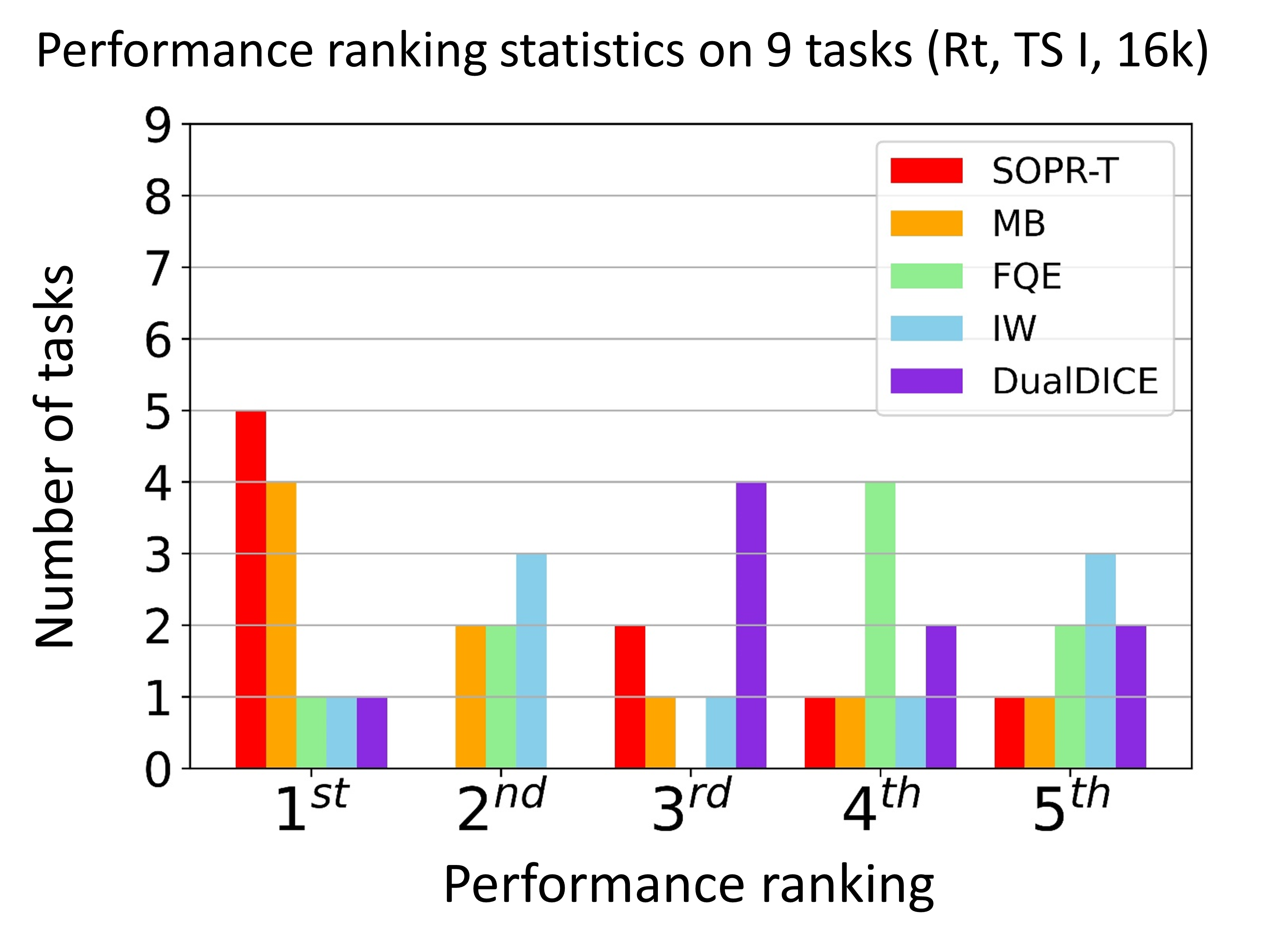}
    \end{minipage}
}

\subfigure[]
{
    \begin{minipage}[t]{0.47\linewidth}
	\centering
    \includegraphics[width = \linewidth]{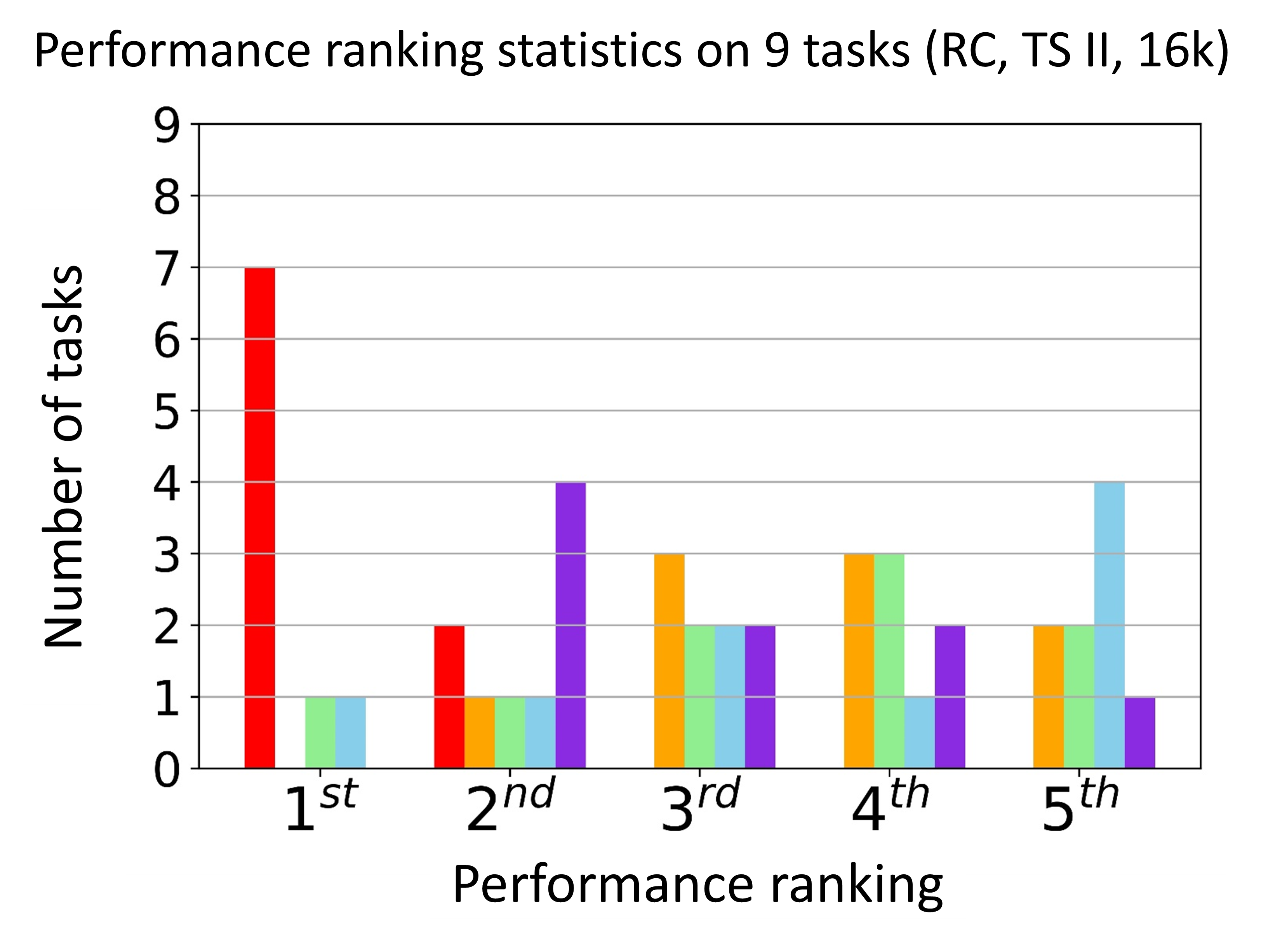}
    \end{minipage}
}
\subfigure[]
{
    \begin{minipage}[t]{0.47\linewidth}
	\centering
    \includegraphics[width = \linewidth]{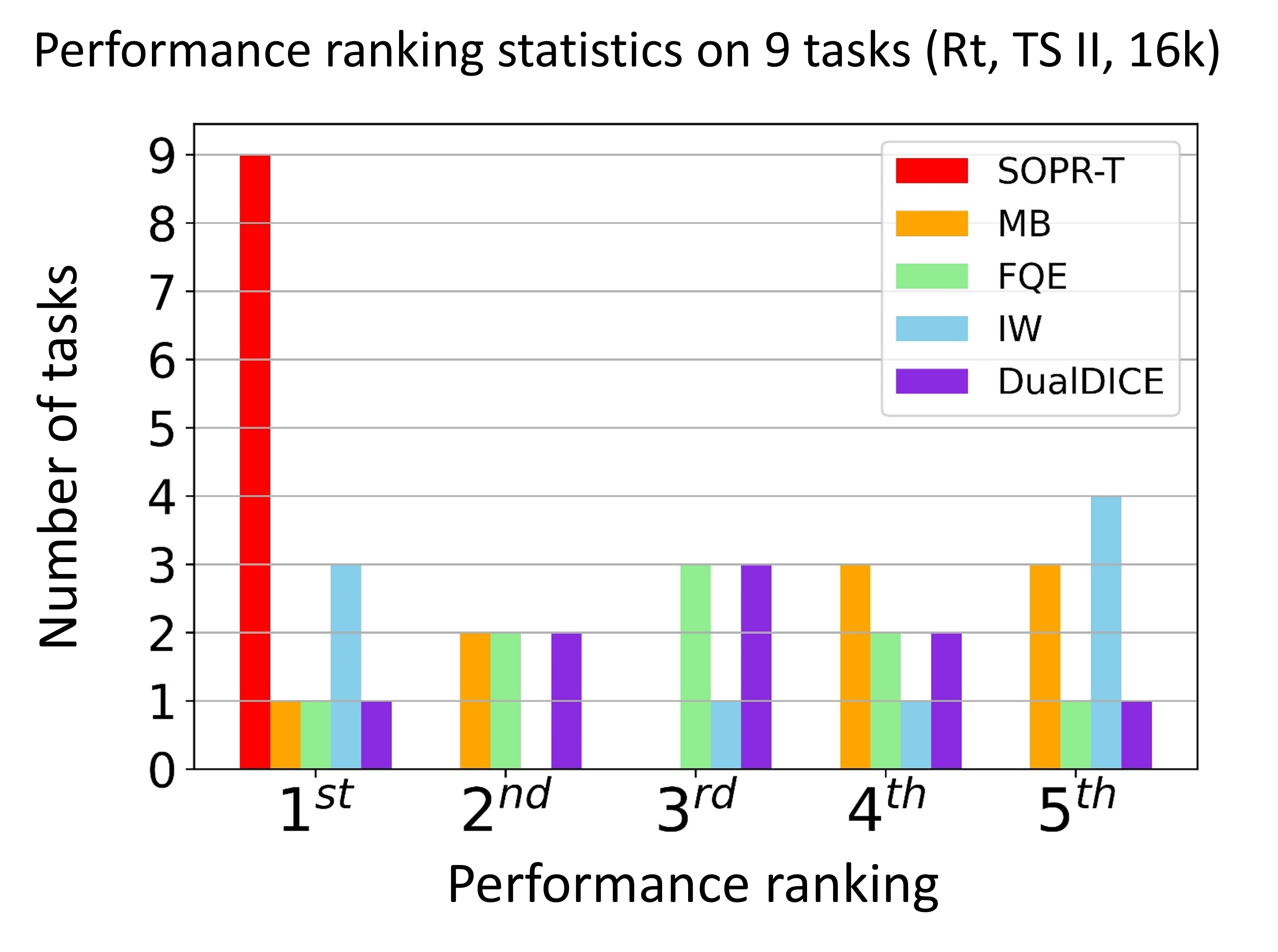}
    \end{minipage}
}
\caption {
Performance ranking statistics of 5 algorithms when the size of dataset is 16k.
Left: rank correlation. Right: regret@3.
Top row: Test Set I.
Bottom row: Test Set II.
}
\label{16k_data}
\end{figure}

Due to space limitation, here we only show performance ranking statistics of all the algorithms on 9 tasks corresponding to 16k data in Figure~\ref{16k_data}.
Detailed performance of each algorithm on each individual task
can be found in Figure~\ref{5_3_supp_16k_1} and Figure~\ref{5_3_supp_16k_2} in Appendix \ref{A.5}.
The results indicate that when using much less data, SOPR-T still outperforms baseline OPE algorithms.

\begin{figure}[!h]
\centering
\subfigure[]
{
    \begin{minipage}[t]{0.47\linewidth}
	\centering
    \includegraphics[width = \linewidth]{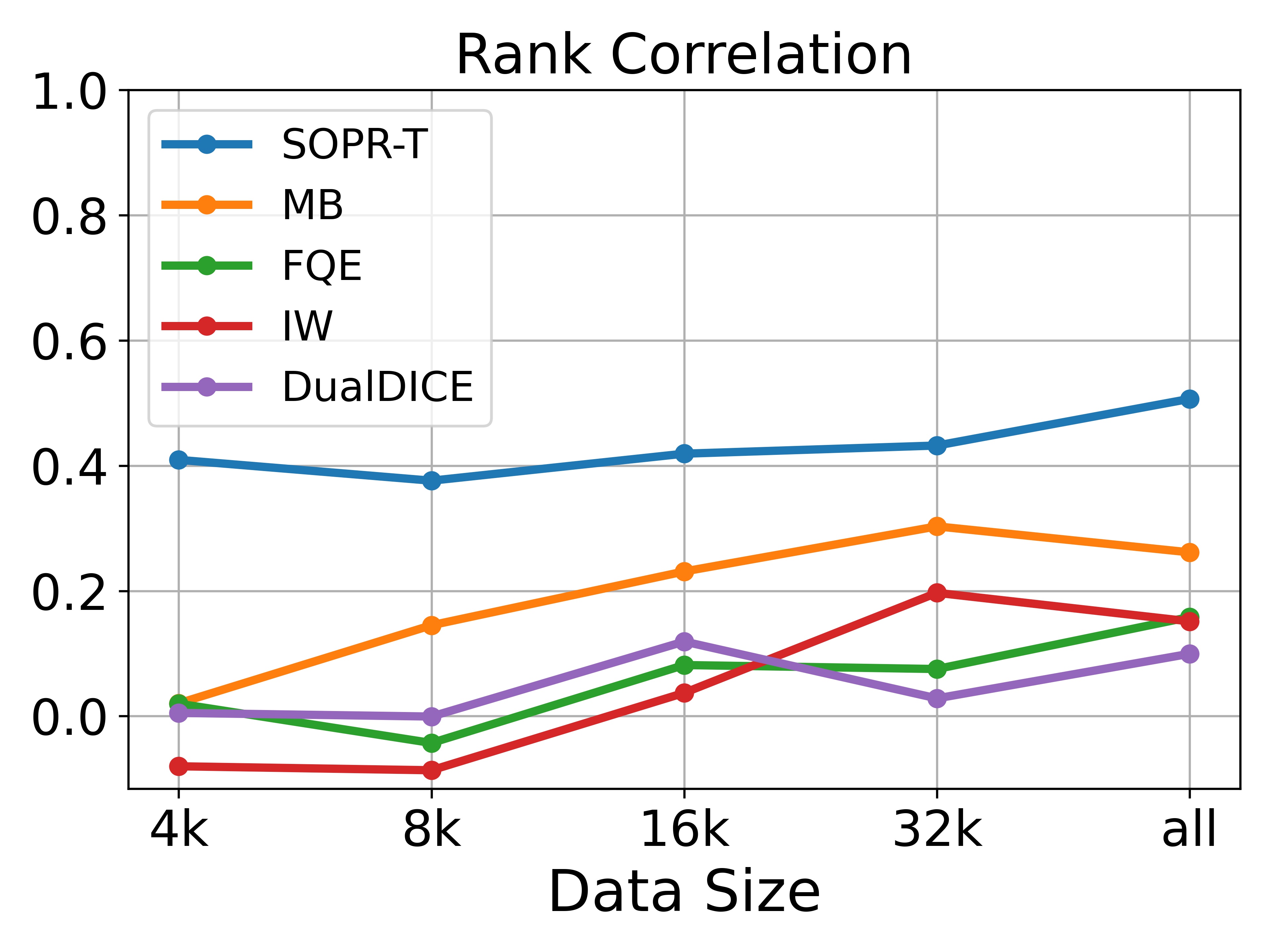}
    \end{minipage}
}
\subfigure[]
{
    \begin{minipage}[t]{0.47\linewidth}
	\centering
    \includegraphics[width = \linewidth]{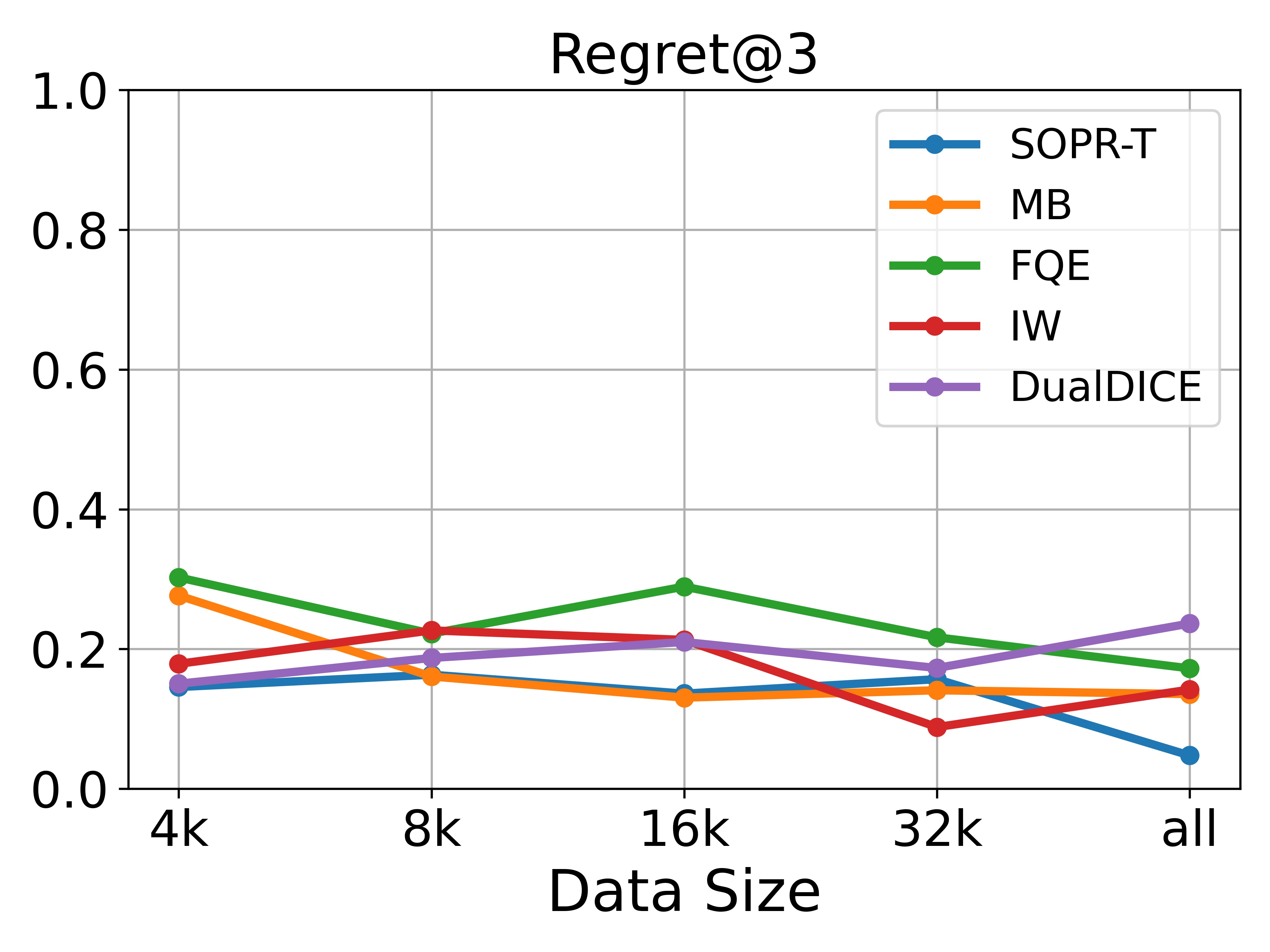}
    \end{minipage}
}

\subfigure[]
{
    \begin{minipage}[t]{0.47\linewidth}
	\centering
    \includegraphics[width = \linewidth]{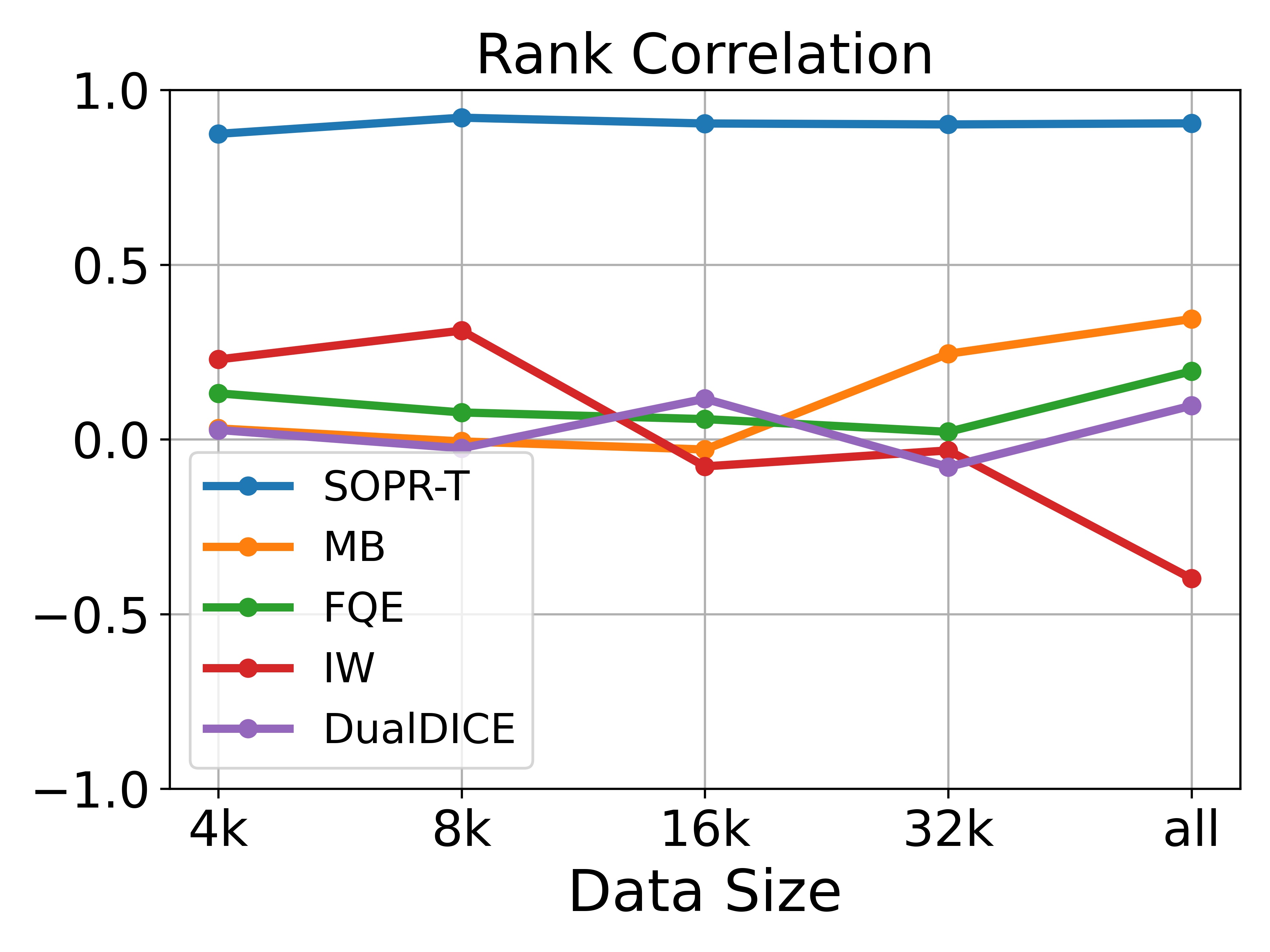}
    \end{minipage}
}
\subfigure[]
{
    \begin{minipage}[t]{0.47\linewidth}
	\centering
    \includegraphics[width = \linewidth]{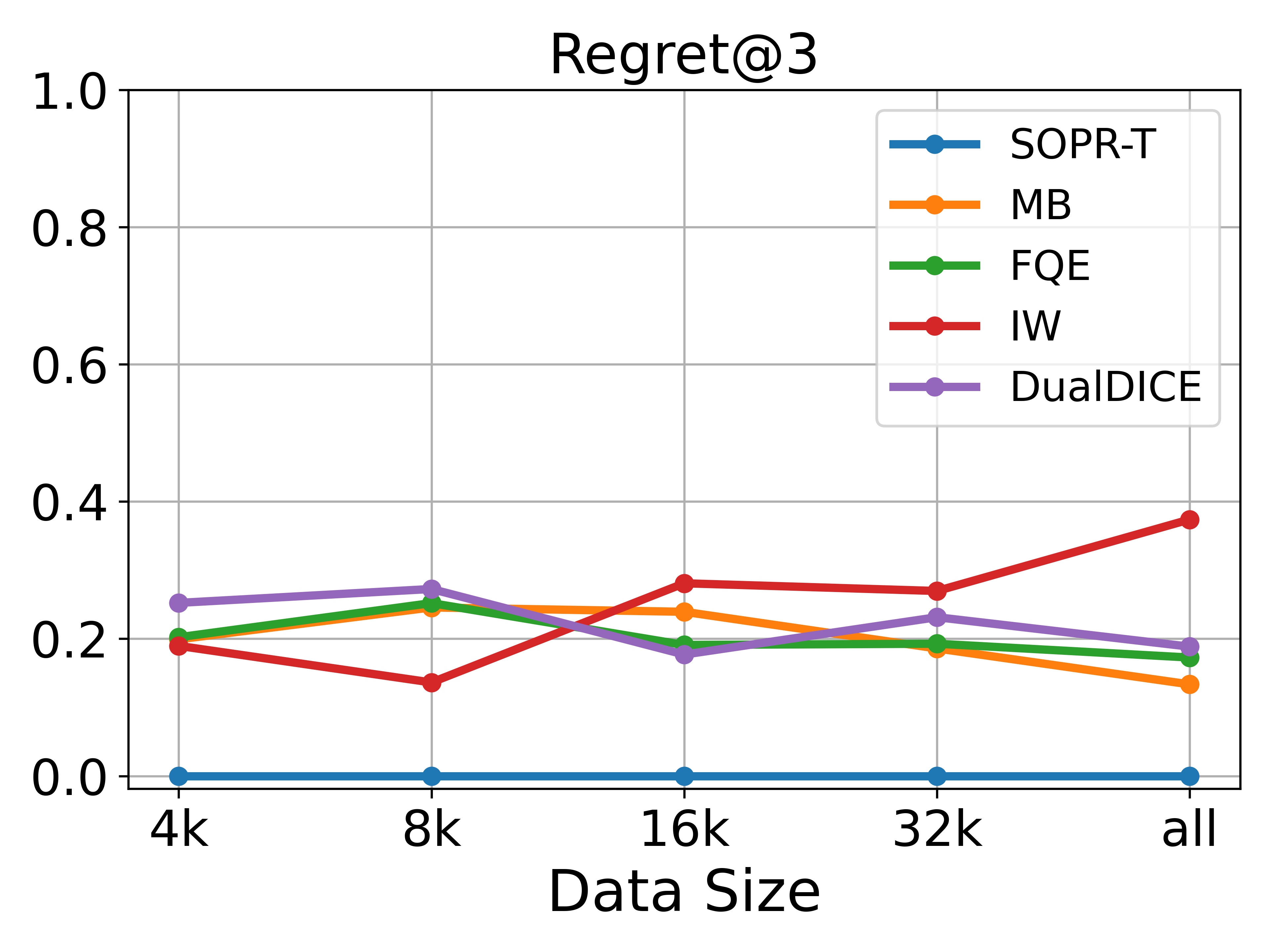}
    \end{minipage}
}
\caption {
Performance comparison with different data size.
Top row: average results on ranking the policies in Test Set I.
Bottom row: average results on ranking the policies in Test Set II.
Left: rank correlation.
Right: regret@3.
}
\label{less_data_trend}
\end{figure}

Figure~\ref{less_data_trend} shows the average results over all tasks of each algorithm with different data size. The results demonstrate that SOPR-T is relatively robust to data size, although with a slight drop as dataset decreases in size.
Remarkably, SOPR-T outperforms baseline OPE algorithms consistently.

\paragraph{Effect of Transformer Encoder}
We investigate the effect of using Transformer encoder to encode state-action pairs.
To this end, we replace Transformer encoder with MLP encoder.
We name the MLP-based SOPR as SOPR-MLP.
Both the number of hidden layers and the number of units of each hidden layer in the MLP encoder are aligned with the Transformer encoder used in SOPR-T.
Details of training and inference methods are also the same as SOPR-T.

\begin{figure}[H]
\centering
\subfigure[]
{
    \begin{minipage}[t]{0.47\linewidth}
	\centering
    \includegraphics[width = \linewidth]{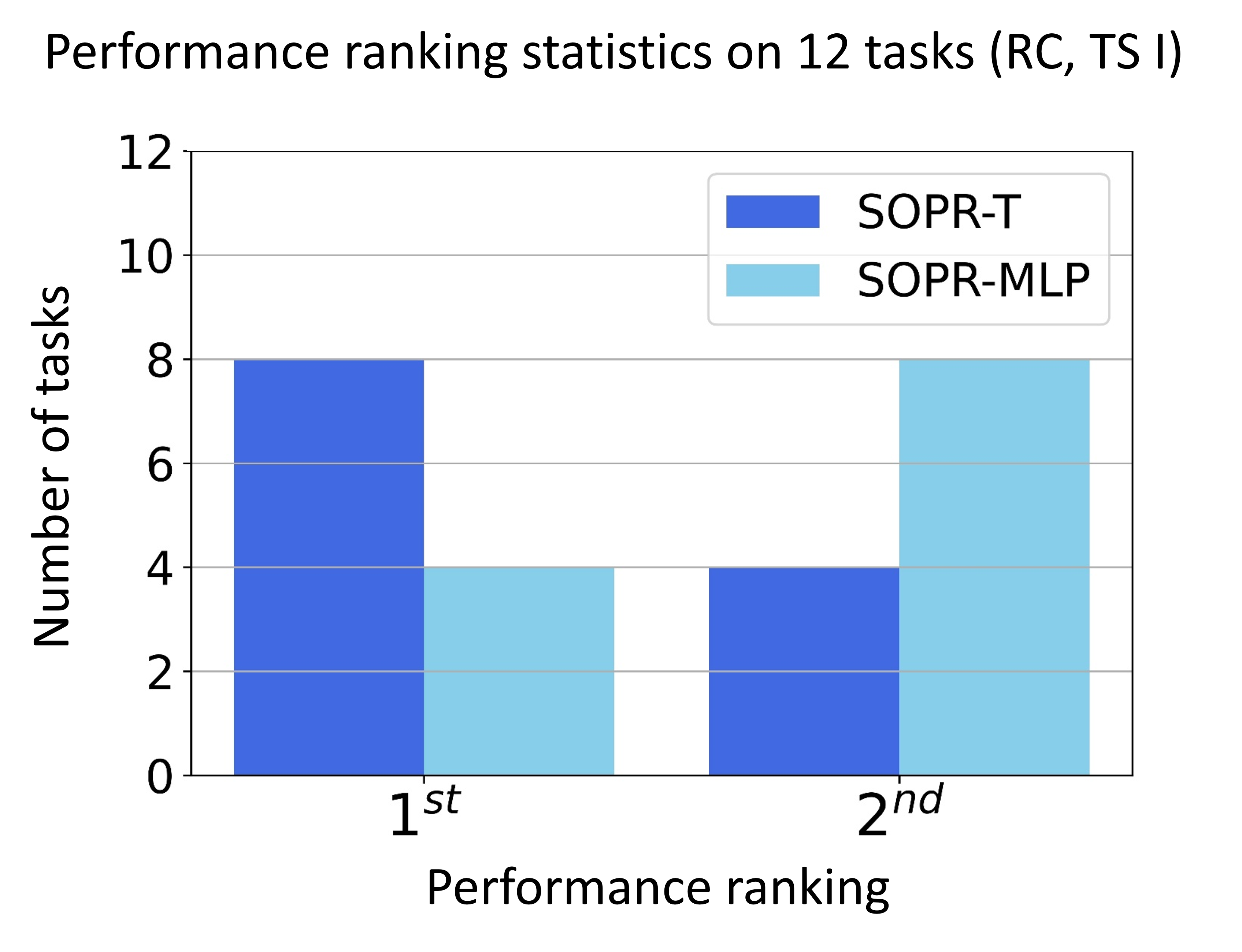}
    \end{minipage}
}
\subfigure[]
{
    \begin{minipage}[t]{0.47\linewidth}
	\centering
    \includegraphics[width = \linewidth]{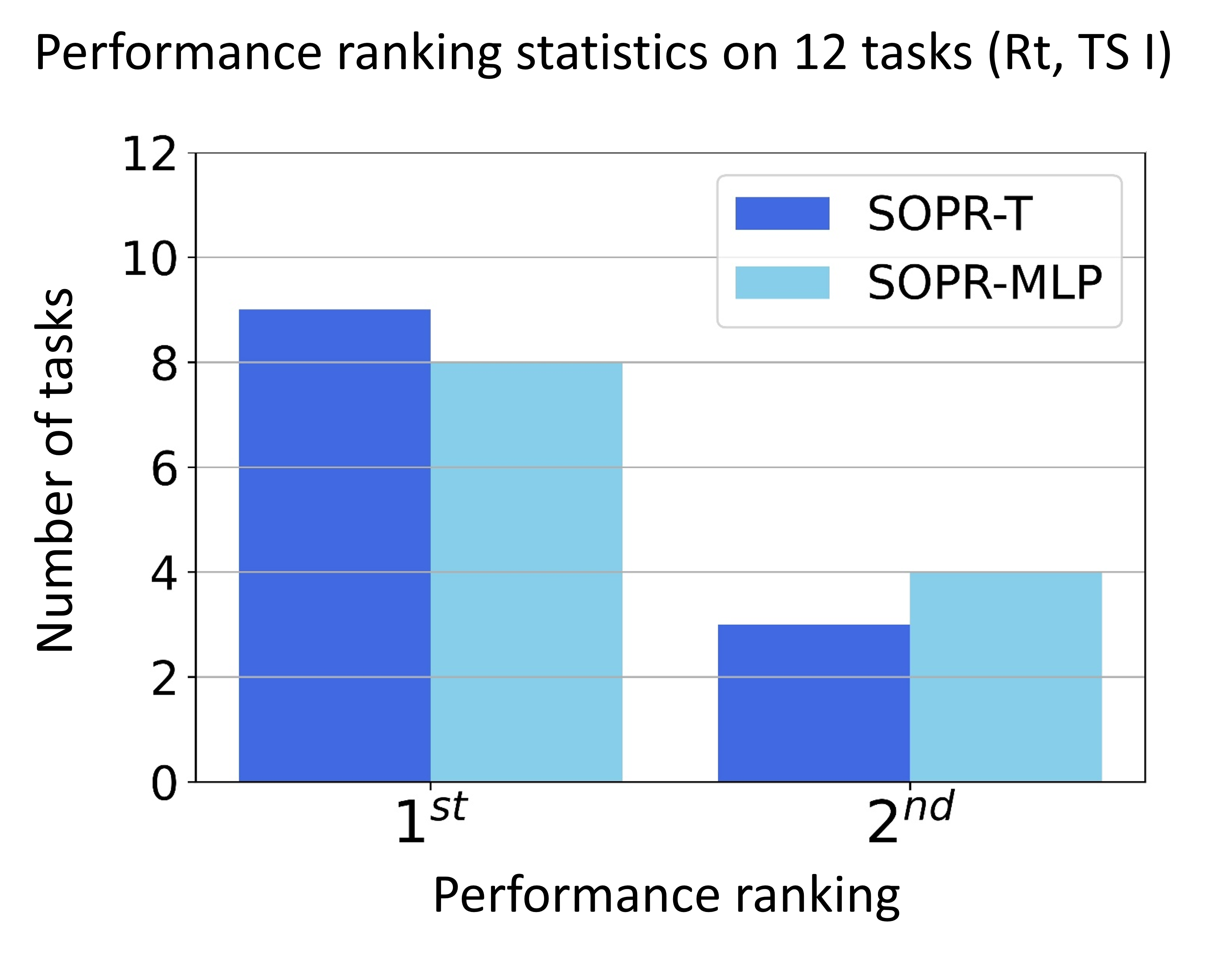}
    \end{minipage}
}

\subfigure[]
{
    \begin{minipage}[t]{0.47\linewidth}
	\centering
    \includegraphics[width = \linewidth]{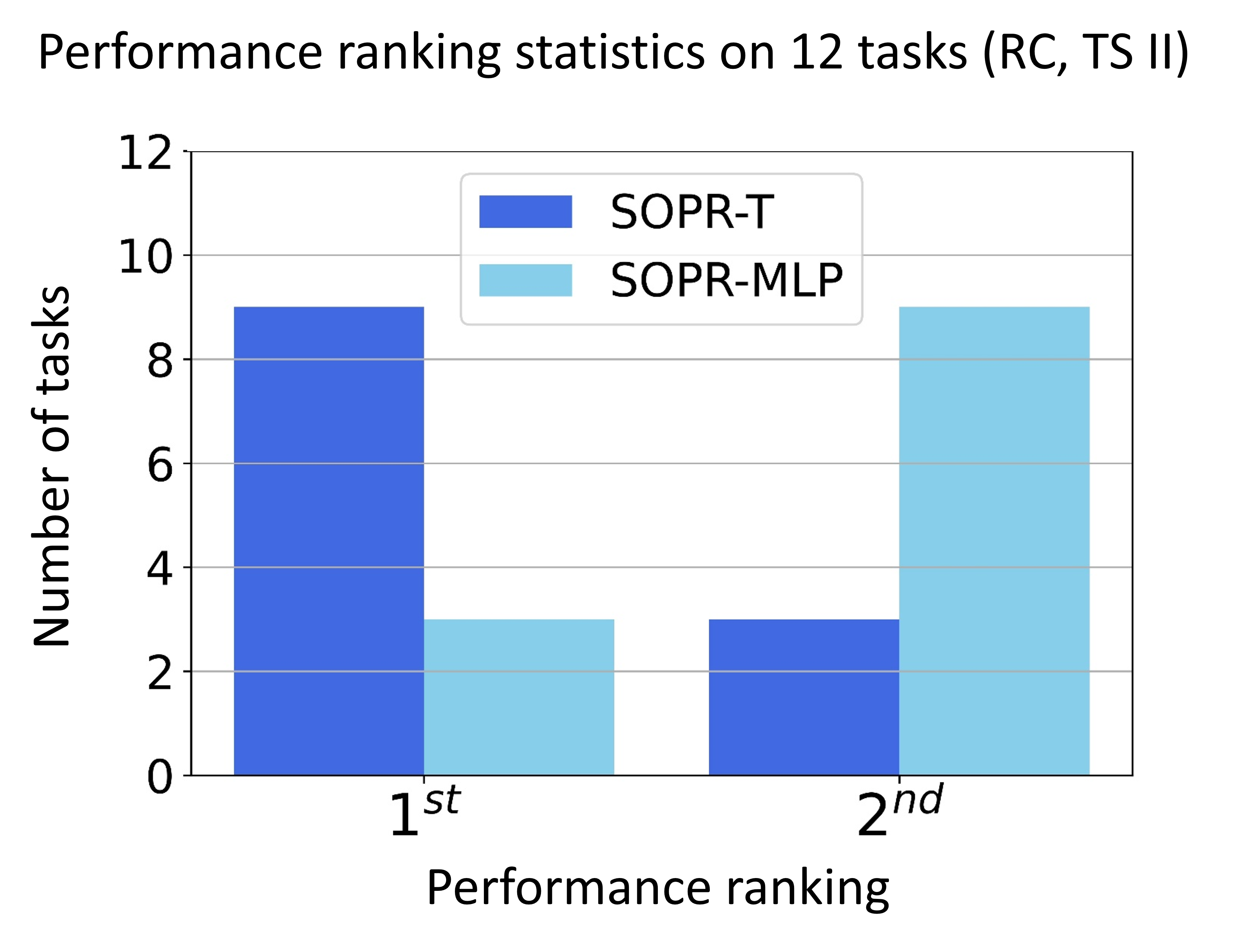}
    \end{minipage}
}
\subfigure[]
{
    \begin{minipage}[t]{0.47\linewidth}
	\centering
    \includegraphics[width = \linewidth]{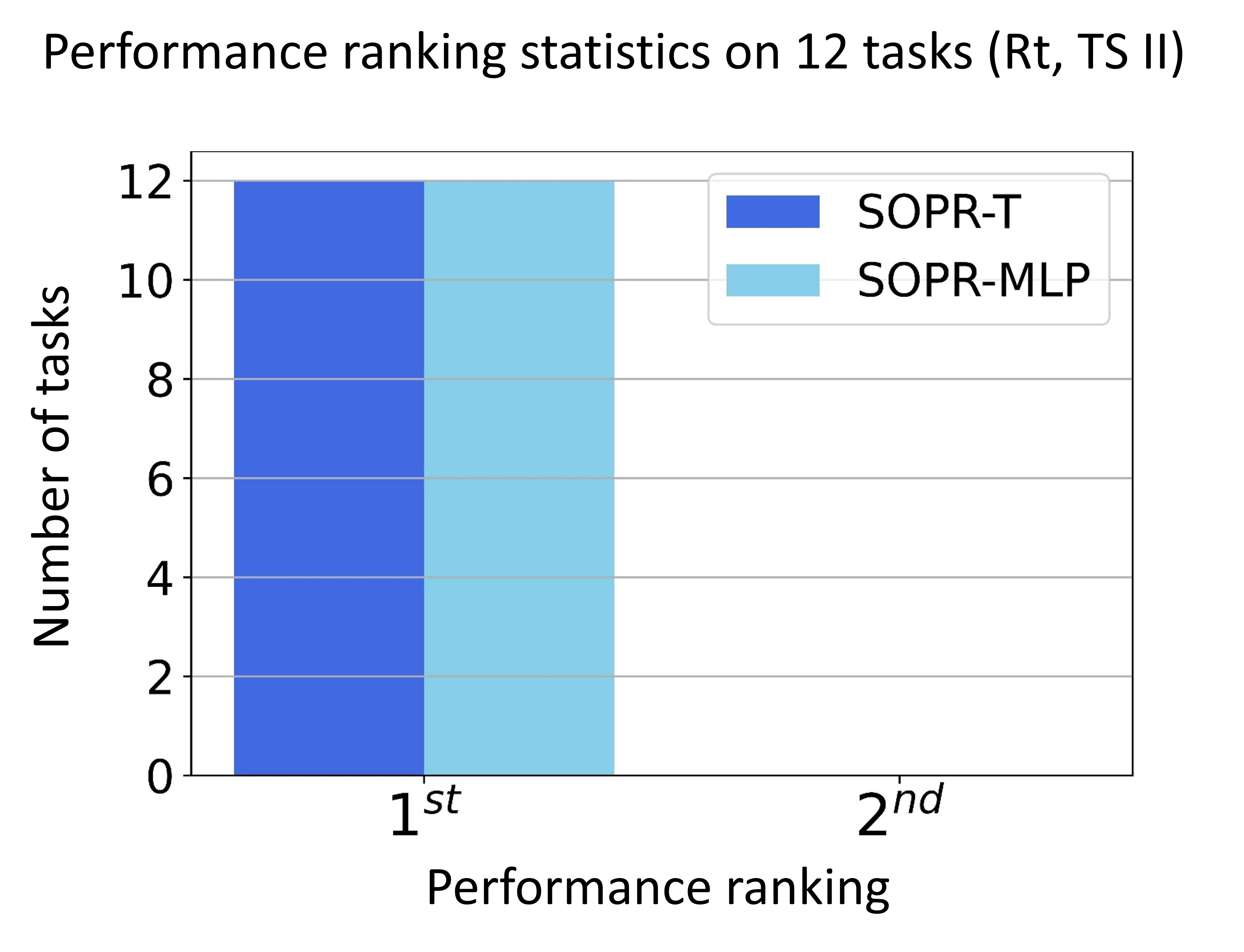}
    \end{minipage}
}
\caption { Performance ranking statistics between SOPR-T and SOPR-MLP on 12 tasks. 
Top row: Test Set I. Bottom row: Test Set II.
Left: rank correlation.
Right: regret@3.
}
\label{TE_vs_MLP}
\end{figure}

We compare the performance of SOPR-T and SOPR-MLP also with two test policy sets.
Figure~\ref{TE_vs_MLP} presents the performance ranking statistics between SOPR-MLP and SOPR-T on 12 tasks. 
As can be seen from the results, SOPR-T outperforms SOPR-MLP in most of the tasks.
Specifically, for the results on Test Set I, as shown in the top row of Figure~\ref{TE_vs_MLP}, SOPR-T achieves higher rank correlation in 8 out of 12 tasks and lower or the same (zero) regret value in 9 out of 12 tasks.
For Test Set II, as shown in the bottom row of Figure~\ref{TE_vs_MLP}, the results demonstrate that SOPR-T achieves higher rank correlation in 9 out of 12 tasks, and the same (zero) regret value in all the tasks.
The results indicate that the Transformer encoder is superior to MLP encoder in terms of encoding state-action pairs and representing policies in our SOPR method.
Detailed performance of each algorithm on each individual task can be found in Figure~\ref{5_3_supp_2.1} and Figure~\ref{5_3_supp_2.3} in Appendix \ref{A.5}.


\paragraph{Performance Variance }
In this part, we investigate the performance variance of SOPR-T in inference, especially when using small data size in inference.
In many real-world applications, the inference speed of policy ranking matters a lot.
Although our SOPR-T is efficient due to its end-to-end scoring model, we would also like to investigate whether SOPR-T can use less data in inference than that used in the training phase in order to reduce inference time.
To this end, in inference, we sample different amounts (1, 5, 10, 50, 100, and 200) of subsets from the whole dataset to calculate an average score for each policy, and then rank different policies with their scores.
The size of each subset is the same as the batch size of training, which is provided in Appendix \ref{A.1}.
Note that, in the above experiments regarding the Effect of Data Size, both training and testing are conducted on the same dataset. However, results in this part correspond to using less data (subsets) in inference, and using the original dataset in training.

\begin{figure}[!t]
\centering
\subfigure[]
{
    \begin{minipage}[t]{0.47\linewidth}
	\centering
    \includegraphics[width = \linewidth]{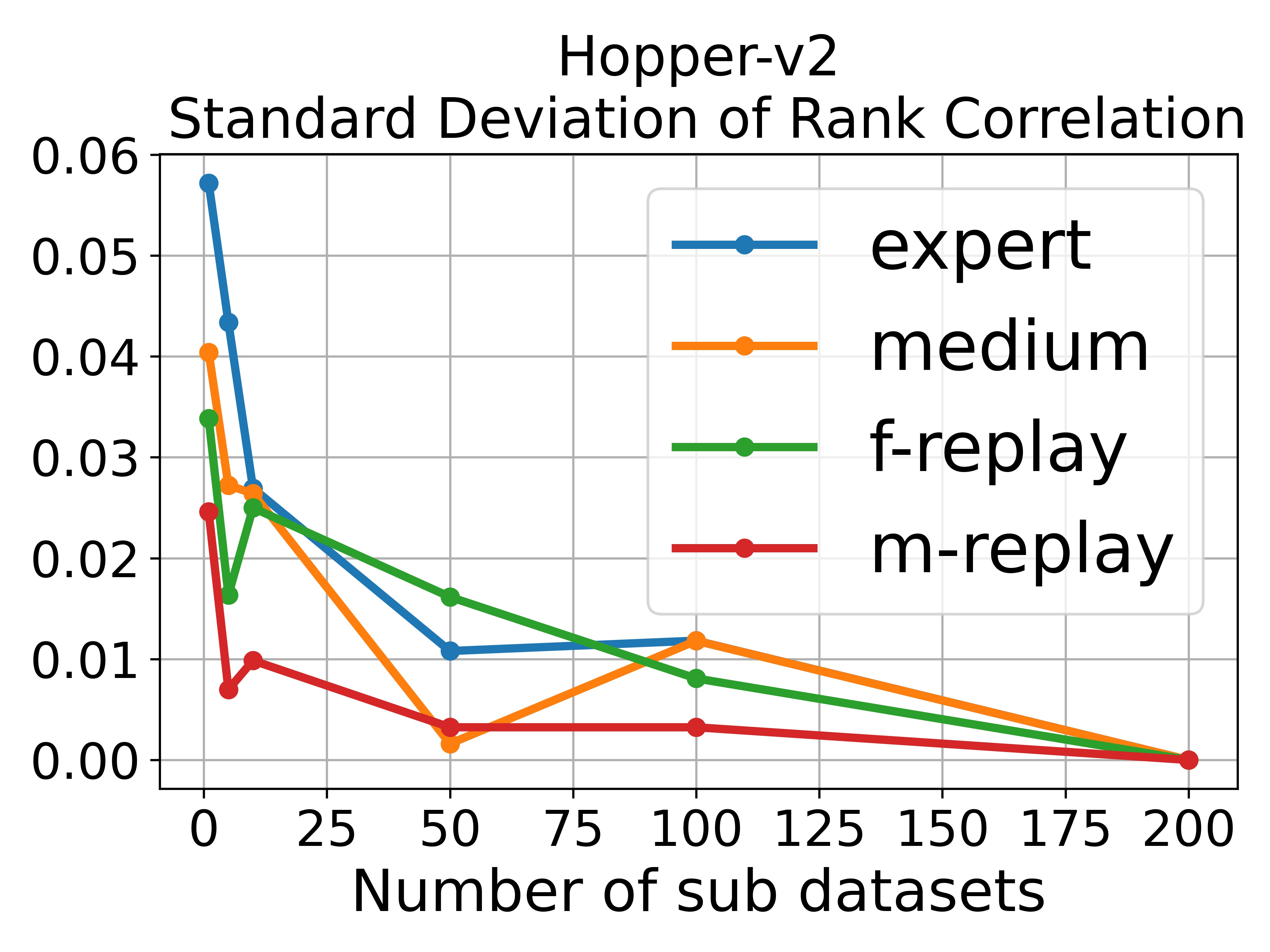}
    \end{minipage}
}
\subfigure[]
{
    \begin{minipage}[t]{0.47\linewidth}
	\centering
    \includegraphics[width = \linewidth]{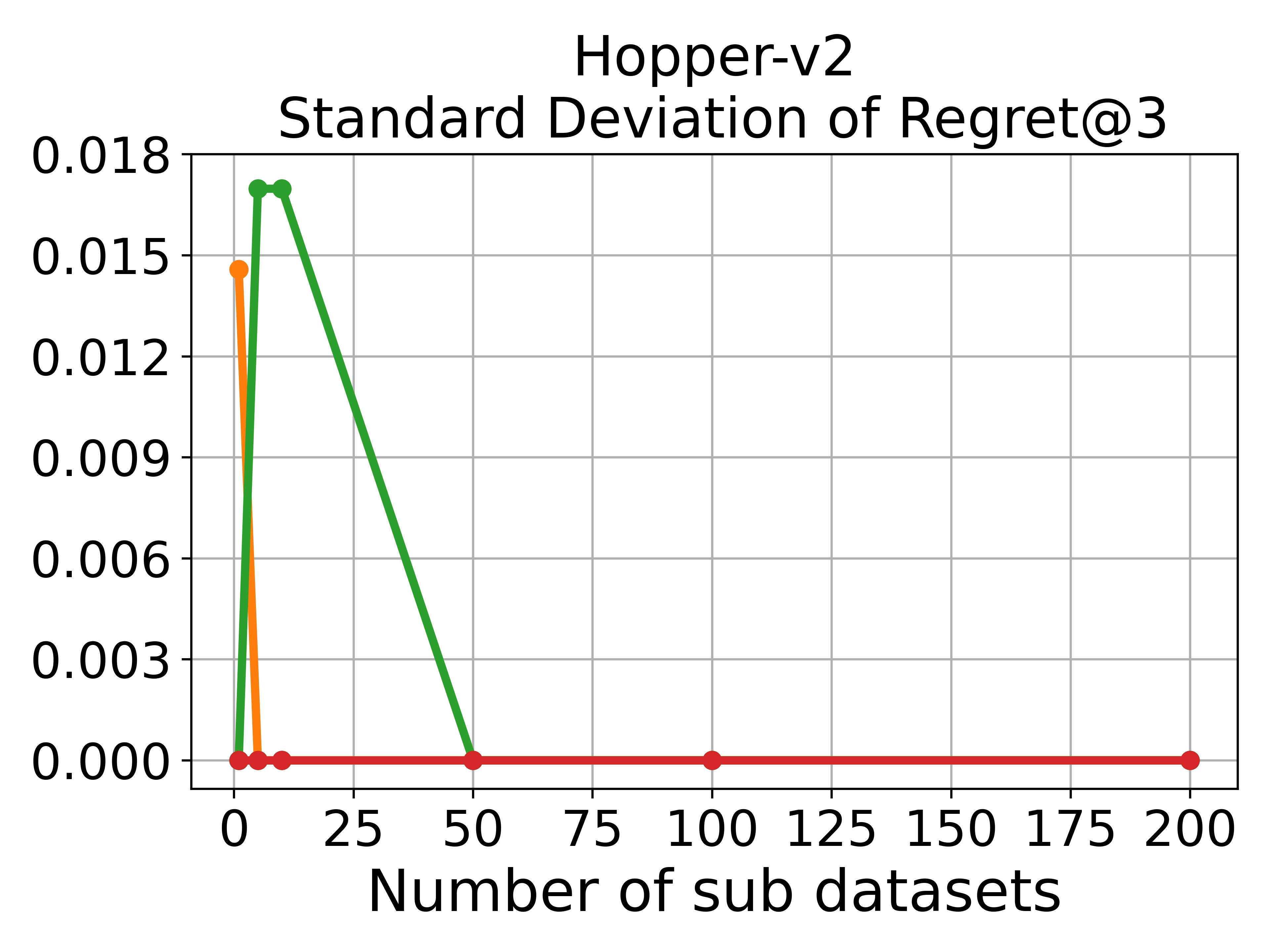}
    \end{minipage}
}
\caption{Standard deviation of SOPR-T with different number of subsets in inference. Left: rank correlation. Right: regret@3.}
\label{var}
\end{figure}

Given the number of sampled subsets, we repeat the inference process five times, and compute the variance of the five results regarding rank correlation or regret value.
The relationship between the standard deviation of the results and the number of sampled subsets is shown in Figure~\ref{var} (Hopper game) and Appendix \ref{A.5} (Figure~\ref{5_3_supp_3.1} and Figure~\ref{5_3_supp_3.2}).
As can be seen from the figures, the standard deviations of rank correlation and regret value of SOPR-T are both small consistently for different amounts of subsets.
As the number of subsets increases, the standard deviation decreases gradually, which indicates SOPR-T achieves more stable performance.

\section{Conclusions and Future Work}
\label{conclusion}
In this work, we defined a new problem, supervised off-policy ranking (SOPR), which aims to score target policies for the purpose of correct ranking them instead of precise return estimation.
We leveraged a set of training policies with known performance ranking in addition to the off-policy data. 
We proposed a hierarchical Transformer encoder-based scoring model, which represents a policy by a set of state-action pairs and then maps such a set to a score indicating the relative performance of the policy. 
The scoring model is trained by minimizing a pairwise ranking loss. 
Experiments demonstrate that our SOPR method outperforms four representative baseline OPE algorithms in terms of rank correlation, regret value, and stability.

As SOPR is a newly defined problem, our work is just a first and preliminary step and far away from solving it. Many possible future directions are left to explore. First, beyond the state-action pair-based representations used in this work, how to better represent a policy? For example, considering the existence of many OPE methods, the values of a policy estimated by those methods can be used to represent the policy, and one can conduct supervised learning to combine the estimations of those methods. Furthermore, the policy values estimated by those OPE methods can serve as features and enhance the representations of a policy adopted in this work. Second, we employed a Transformer based scoring model in this work. It is interesting to explore other possibilities for the scoring model. Third, we adopted the pairwise ranking loss in our algorithm. There are other ranking loss functions (e.g., listwise loss~\cite{cao2007learning}) which are worth trying. Fourth, since we conducted supervised learning, a limitation of our algorithm is that its effectiveness will be heavily impacted by the quality of training policies. We will study and quantify the impact of the quality of training policies in our future work. Last but not least, as a supervised learning problem, there are many theoretical questions to answer~\cite{vapnik1999overview, vapnik2013nature}. For example, is a learning algorithm consistent for supervised OPE or supervised OPR? What is the rate of convergence of a learning algorithm? What is the generalization bound of a learning algorithm?  
We hope that our work can inspire more research along the direction of supervised OPE/OPR.

\bibliographystyle{icml2022}
\bibliography{main}

\newpage
\appendix
\onecolumn
\section{Appendix}
\subsection{Model and Training Configurations} \label{A.1}

Table \ref{hyperparameter} lists the configurations of our model and training process.
\begin{table}[!h]
\renewcommand\arraystretch{1.3}
\caption{Model and training configurations for SOPR-T.}
\centering
\begin{tabular}{l l }  
\toprule [1pt]
Hyperparameter & Value \\
\hline
Input linear projection layer & ((dim\_s+dim\_a), 64)  \\
Low-level encoder & n\_layers=2, n\_head=2, dim\_feedforward=128, dropout=0.1   \\
High-level encoder & n\_layers=6, n\_head=8, dim\_feedforward=512, dropout=0.1\\
Output linear projection layer & (256, 1)  \\
Optimizer        &Adam \\
Learning rate    &0.001 \\
Batch size       &$|D_s|=16\text{k}$ \\
Number of clusters  &$K=256$ \\
\bottomrule [1pt]
\end{tabular} \label{hyperparameter}
\end{table}

\subsection{Computational Resource and Time Cost}
Our experiments are run with a Nvidia Tesla P100 GPU.
Table~\ref{training_t} and Table~\ref{test_t_sopr} present training and inference time cost of SOPR-T regarding different random seeds, respectively.
Table~\ref{training_t_compare}, Table~\ref{test_t_compare_1}, and Table~\ref{test_t_compare_2} present comparison of average time cost among different algorithms.
Note that, the time cost of SOPR-T in inference depends on the number of subsets used to calculate an average score for a policy.
Unless otherwise specified, in inference, we use all ($200$) subsets that are used in training.
We observe that, as shown in the last part of Section~\ref{exp_further} and the last part of Appendix (Figure~\ref{5_3_supp_3.1}), SOPR-T with 5 subsets achieves nearly the same performance as using 200 subsets.
Approximately, to score and rank 10 policies in each task shown in Table~\ref{test_t_sopr}, SOPR-T takes less than {\bf{25.24}} (1009.6/40) seconds to get a good ranking, which is comparable to IW and much faster than FQE and DualDICE.

\begin{table}[H]
\caption{Training time cost of SOPR-T. (seconds)}
\centering
\begin{tabular}{lllll}
\toprule [1pt]
Task Name   & Seed 0   & Seed 1   & Seed 2   & Average \\
\hline
halfcheetah-expert        & 3937.8 & 3987.3 & 3993.0 & \textbf{3972.7} \\
halfcheetah-medium        & 3816.1 & 3973.8 & 3923.8 & \textbf{3904.6} \\
halfcheetah-medium-replay & 4653.5 & 4987.4 & 4969.4 & \textbf{4870.1} \\
halfcheetah-full-replay   & 5283.5 & 5151.5 & 6300.0 & \textbf{5578.3} \\
\hline
hopper-expert             & 3204.7 & 3214.5 & 3228.9 & \textbf{3216.0} \\
hopper-medium             & 3216.3 & 3177.5 & 3298.5 & \textbf{3230.8} \\
hopper-medium-replay      & 3799.9 & 3811.7 & 4123.5 & \textbf{3911.7} \\
hopper-full-replay        & 3837.4 & 3795.7 & 3909.6 & \textbf{3847.6} \\
\hline
walker2d-expert           & 3702.3 & 4030.1 & 3678.7 & \textbf{3803.7} \\
walker2d-medium           & 3601.0 & 3740.9 & 3647.4 & \textbf{3663.1} \\
walker2d-medium-replay    & 4687.4 & 4719.2 & 4606.0 & \textbf{4670.9} \\
walker2d-full-replay      & 4683.7 & 4566.4 & 4648.7 & \textbf{4632.9} \\
\bottomrule [1pt]
\end{tabular}
\label{training_t}
\end{table}

\begin{table}[H]
\caption{ Comparison of training time cost. (seconds)}
\centering
\begin{tabular}{llccll}
\toprule [1pt]
Task Name                 & \textbf{SOPR-T} & \multicolumn{1}{l}{FQE} & \multicolumn{1}{l}{DualDICE} & MB        & IW \\
\hline
halfcheetah-expert        & \textbf{3972.7} & /                       & /                            & 6610.0 & 3841.2 \\
halfcheetah-medium        & \textbf{3904.6} & /                       & /                            & 4386.1 & 4000.8 \\
halfcheetah-medium-replay & \textbf{4870.1} & /                       & /                            & 4682.8 & 2762.3 \\
halfcheetah-full-replay   & \textbf{5578.3} & /                       & /                            & 5037.2 & 3370.9 \\
\hline
hopper-expert             & \textbf{3216.0} & /                       & /                            & 7614.4 & 3695.9 \\
hopper-medium             & \textbf{3230.8} & /                       & /                            & 4764.0 & 3684.9 \\
hopper-medium-replay      & \textbf{3911.7} & /                       & /                            & 5104.8 & 3034.7 \\
hopper-full-replay        & \textbf{3847.6} & /                       & /                            & 4776.6 & 3523.0 \\
\hline
walker2d-expert           & \textbf{3803.7} & /                       & /                            & 4582.4 & 3371.0 \\
walker2d-medium           & \textbf{3663.1} & /                       & /                            & 4629.5 & 3542.5 \\
walker2d-medium-replay    & \textbf{4670.9} & /                       & /                            & 4780.0 & 2933.2 \\
walker2d-full-replay      & \textbf{4632.9} & /                       & /                            & 4732.3 & 3532.7 \\
\bottomrule [1pt]
\end{tabular}
\label{training_t_compare}
\end{table}

\begin{table}[H]
\caption{
Inference time cost of SOPR-T (ranking 10 policies). (seconds)}
\centering
\begin{tabular}{lllllllll}
\toprule [1pt]
\multicolumn{1}{c}{\multirow{2}{*}{Task   Name}} & \multicolumn{4}{c}{Test Set I}    & \multicolumn{4}{c}{Test Set II} \\
\multicolumn{1}{c}{}                             & Seed 0  & Seed 1  & Seed 2 & Average & Seed 0 & Seed 1 & Seed 2 & Average \\ \cline{1-1}
\hline
halfcheetah-expert                                 & 799.9  & 770.9  & 775.5 & \textbf{782.1}   & 600.4 & 515.0 & 523.6 & \textbf{546.3}   \\
halfcheetah-medium                                 & 741.7  & 775.9  & 746.5 & \textbf{754.7}   & 532.5 & 543.0 & 539.8 & \textbf{538.4}   \\
halfcheetah-medium-replay                          & 854.3  & 864.5  & 849.6 & \textbf{856.1}   & 658.6 & 660.1 & 668.9 & \textbf{662.5}   \\
halfcheetah-full-replay                            & 1038.1 & 1050.7 & 940.1 & \textbf{1009.6}  & 657.7 & 664.7 & 658.8 & \textbf{660.4}   \\
\hline
hopper-expert                                      & 843.4  & 841.4  & 833.6 & \textbf{839.4}   & 490.3 & 502.2 & 494.7 & \textbf{495.7}   \\
hopper-medium                                      & 696.6  & 701.1  & 761.8 & \textbf{719.8}   & 481.9 & 488.1 & 500.6 & \textbf{490.2}   \\
hopper-medium-replay                               & 772.9  & 790.1  & 803.7 & \textbf{788.9}   & 586.5 & 583.7 & 591.2 & \textbf{587.1}   \\
hopper-full-replay                                 & 737.2  & 735.0  & 745.6 & \textbf{739.3}   & 586.0 & 643.9 & 650.3 & \textbf{626.7}   \\
\hline
walker2d-expert                                    & 814.1  & 809.8  & 785.0 & \textbf{803.0}   & 544.5 & 518.8 & 531.3 & \textbf{531.5}   \\
walker2d-medium                                    & 906.5  & 874.7  & 893.9 & \textbf{891.7}   & 544.7 & 547.8 & 520.3 & \textbf{537.6}   \\
walker2d-medium-replay                             & 781.7  & 781.8  & 779.9 & \textbf{781.2}   & 652.7 & 663.5 & 659.0 & \textbf{658.4}   \\
walker2d-full-replay                               & 946.9  & 939.6  & 946.7 & \textbf{944.4}   & 659.3 & 671.7 & 656.3 & \textbf{662.4}  \\
\bottomrule [1pt]
\end{tabular}
\label{test_t_sopr}
\end{table}

\begin{table}[H]
\caption{ Comparison of inference time cost (ranking 10 policies in Test Set I). (seconds)}
\centering
\begin{tabular}{llllll}
\toprule [1pt]
Task   Name               & \textbf{SOPR-T} & FQE      & DualDICE & MB    & IW   \\
\hline
halfcheetah-expert        & \textbf{782.1}  & 63262.7  & 70975.3  & 142.4 & 27.8 \\
halfcheetah-medium        & \textbf{754.7}  & 39271.1  & 48123.2  & 121.3 & 19.0 \\
halfcheetah-medium-replay & \textbf{856.1}  & 38574.8  & 44506.9  & 96.7  & 14.3 \\
halfcheetah-full-replay   & \textbf{1009.6} & 39221.6  & 48345.0  & 141.0 & 21.1 \\
\hline
hopper-expert             & \textbf{839.4}  & 39055.9  & 50544.5  & 135.0 & 19.2 \\
hopper-medium             & \textbf{719.8}  & 72422.3  & 79971.6  & 163.9 & 26.0 \\
hopper-medium-replay      & \textbf{788.9}  & 39408.0  & 53415.2  & 137.6 & 27.1 \\
hopper-full-replay        & \textbf{739.3}  & 299882.7 & 299237.4 & 446.2 & 48.8 \\
\hline
walker2d-expert           & \textbf{803.0}  & 38092.9  & 58832.3  & 125.0 & 19.8 \\
walker2d-medium           & \textbf{891.7}  & 39518.1  & 47013.5  & 134.2 & 20.6 \\
walker2d-medium-replay    & \textbf{781.2}  & 39394.2  & 62608.0  & 142.2 & 20.3 \\
walker2d-full-replay      & \textbf{944.4}  & 47131.7  & 51518.7  & 214.9 & 26.6 \\
\bottomrule [1pt]
\end{tabular}
\label{test_t_compare_1}
\end{table}

\begin{table}[H]
\caption{ Comparison of inference time cost (ranking 10 policies in Test Set II). (seconds)}
\centering
\begin{tabular}{llllll}
\toprule [1pt]
Task   Name               & \textbf{SOPR-T} & FQE      & DualDICE & MB    & IW   \\
\hline
halfcheetah-expert        & \textbf{546.3}  & 48336.0 & 47690.6  & 125.8 & 19.6 \\
halfcheetah-medium        & \textbf{538.4}  & 45660.3 & 52757.4  & 105.4 & 19.7 \\
halfcheetah-medium-replay & \textbf{662.5}  & 42467.1 & 56625.4  & 82.9 & 14.6 \\
halfcheetah-full-replay   & \textbf{660.4}  & 43887.4 & 47438.6  & 110.1 & 21.1 \\
\hline
hopper-expert             & \textbf{495.7}  & 44551.3 & 48764.1  & 103.4 & 19.3 \\
hopper-medium             & \textbf{490.2}  & 51706.9 & 59532.8  & 152.1 & 25.2 \\
hopper-medium-replay      & \textbf{587.1}  & 44676.7 & 58953.9  & 122.5 & 25.9 \\
hopper-full-replay        & \textbf{626.7}  & 60112.8 & 90679.3  & 180.8 & 37.2 \\
\hline
walker2d-expert           & \textbf{531.5}  & 52014.5 & 49146.8  & 112.1 & 19.6 \\
walker2d-medium           & \textbf{537.6}  & 49727.6 & 53583.4  & 115.2 & 20.7 \\
walker2d-medium-replay    & \textbf{658.4}  & 44017.6 & 49982.8  & 108.2 & 20.2 \\
walker2d-full-replay      & \textbf{662.4}  & 44859.8 & 56993.5  & 129.3 & 27.3 \\
\bottomrule [1pt]
\end{tabular}
\label{test_t_compare_2}
\end{table}






\subsection{Performance on Offline Learned Policies (Corresponding to Section \ref{exp_testset_1})} \label{A.3}

\begin{figure}[H]
\centering
\subfigure[]
{
    \begin{minipage}[t]{0.3\linewidth}
	\centering
    \includegraphics[width = \linewidth]{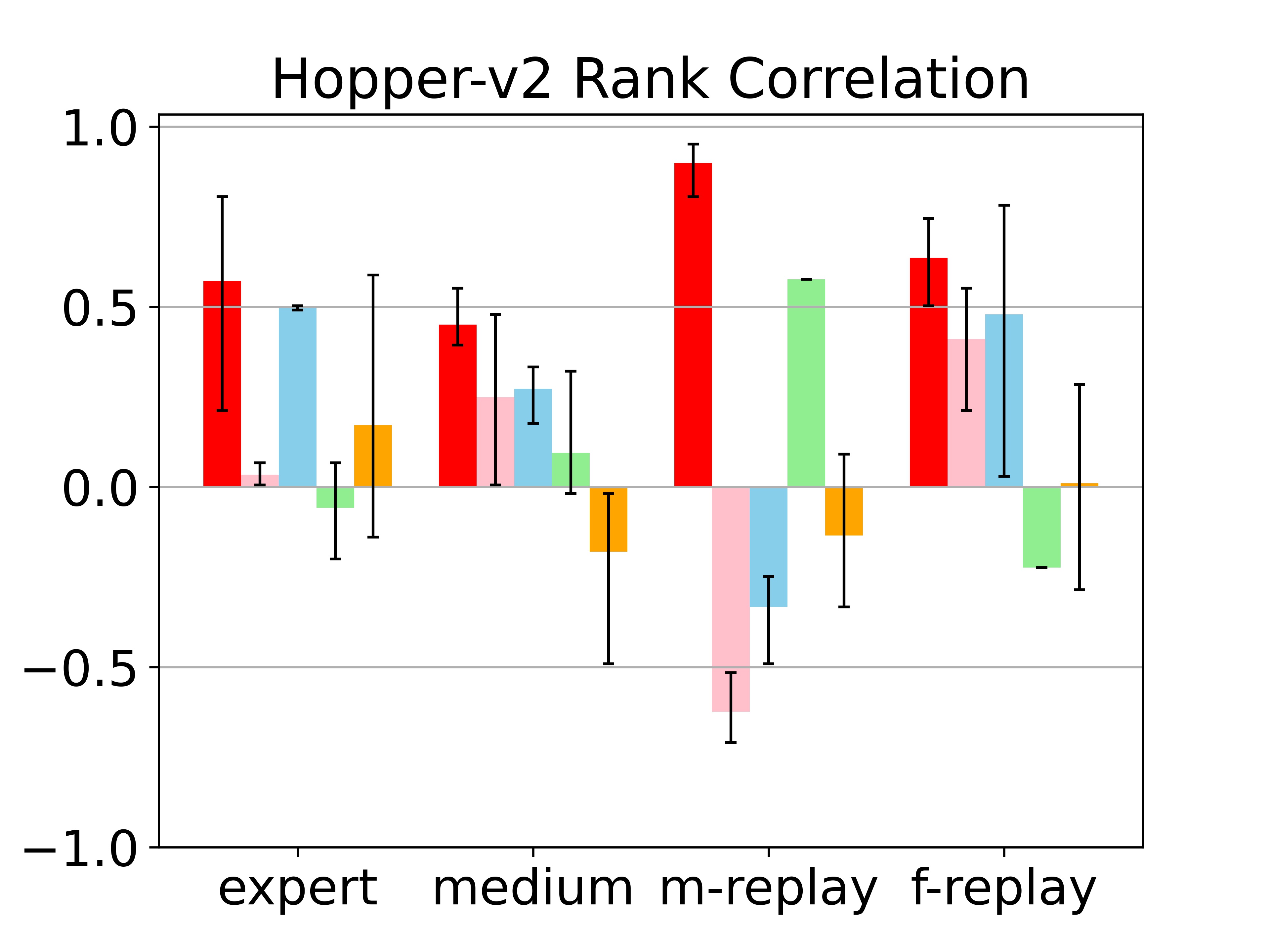}
    \end{minipage}
}
\subfigure[]
{
    \begin{minipage}[t]{0.3\linewidth}
	\centering
    \includegraphics[width = \linewidth]{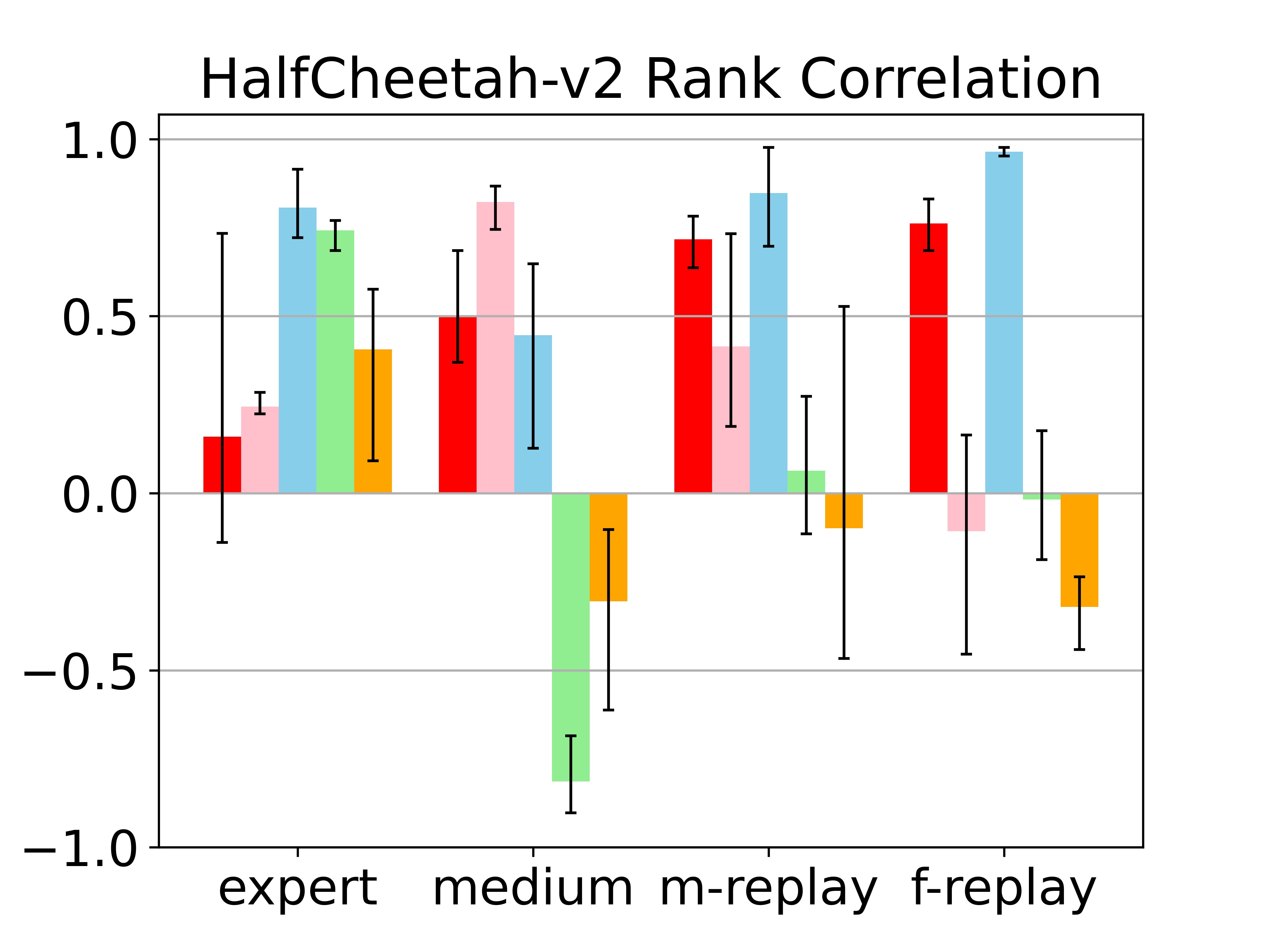}
    \end{minipage}
}
\subfigure[]
{
    \begin{minipage}[t]{0.3\linewidth}
	\centering
    \includegraphics[width = \linewidth]{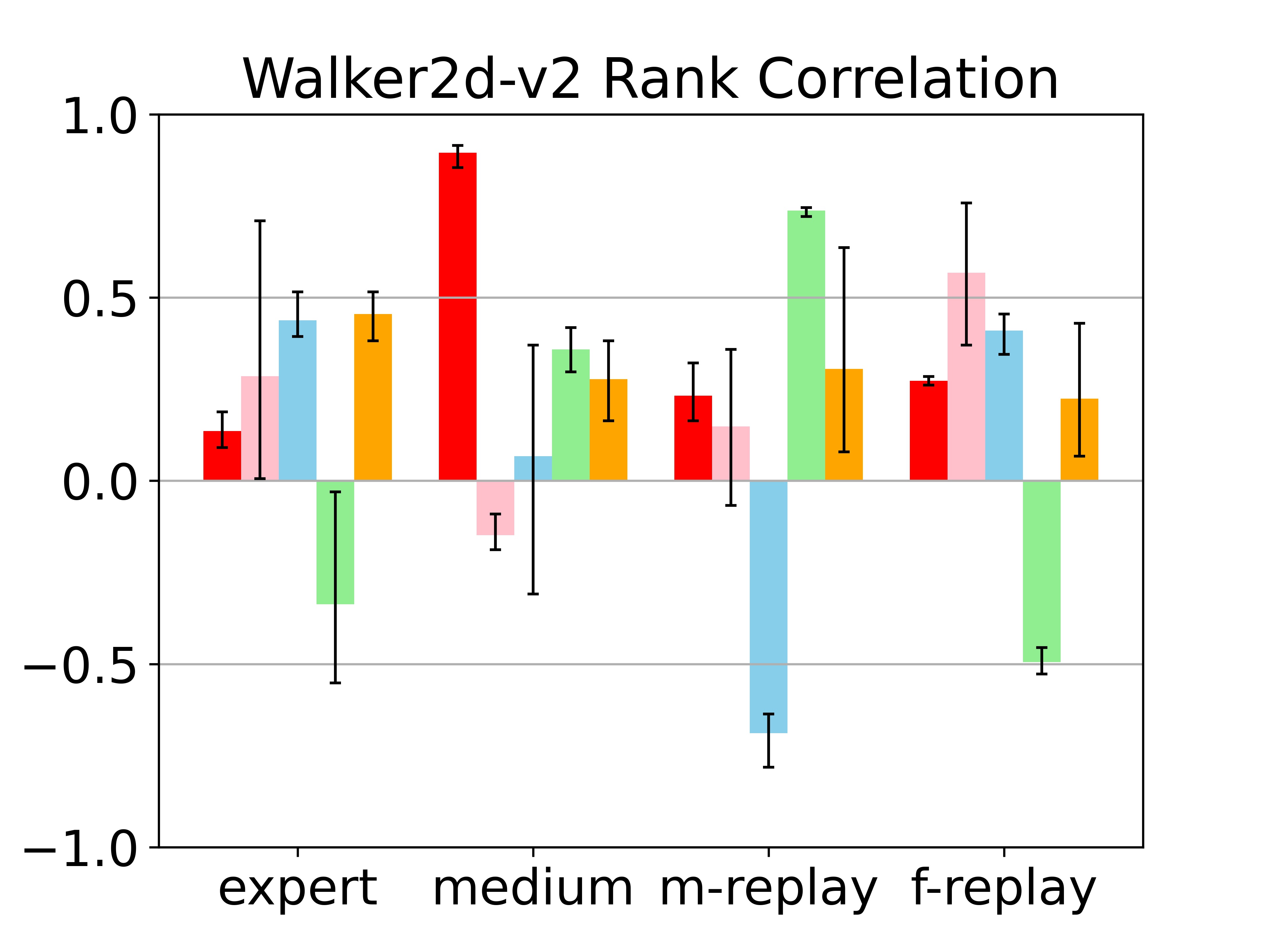}
    \end{minipage}
}

\subfigure[]
{
    \begin{minipage}[t]{0.3\linewidth}
	\centering
    \includegraphics[width = \linewidth]{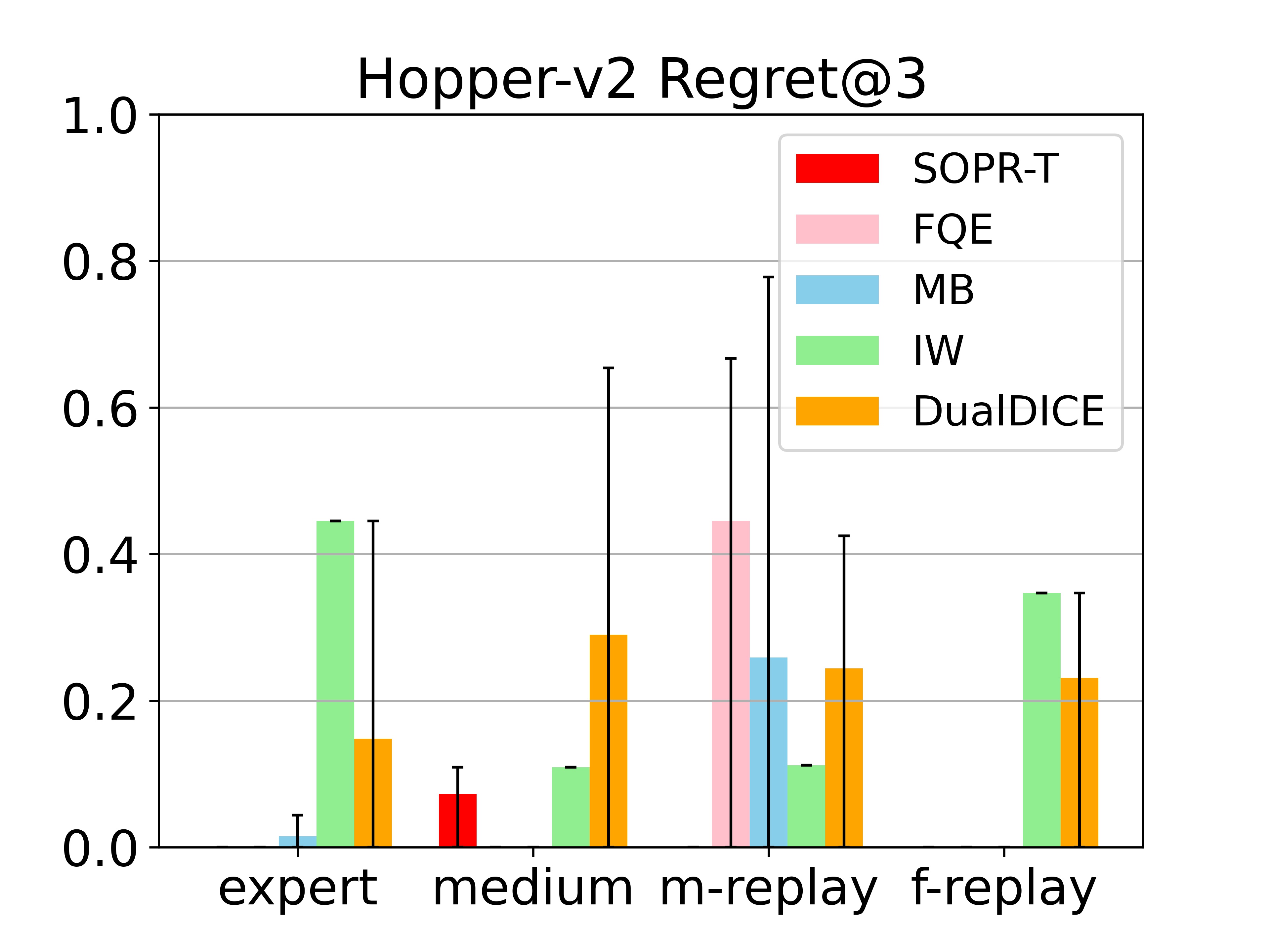}
    \end{minipage}
}
\subfigure[]
{
    \begin{minipage}[t]{0.3\linewidth}
	\centering
    \includegraphics[width = \linewidth]{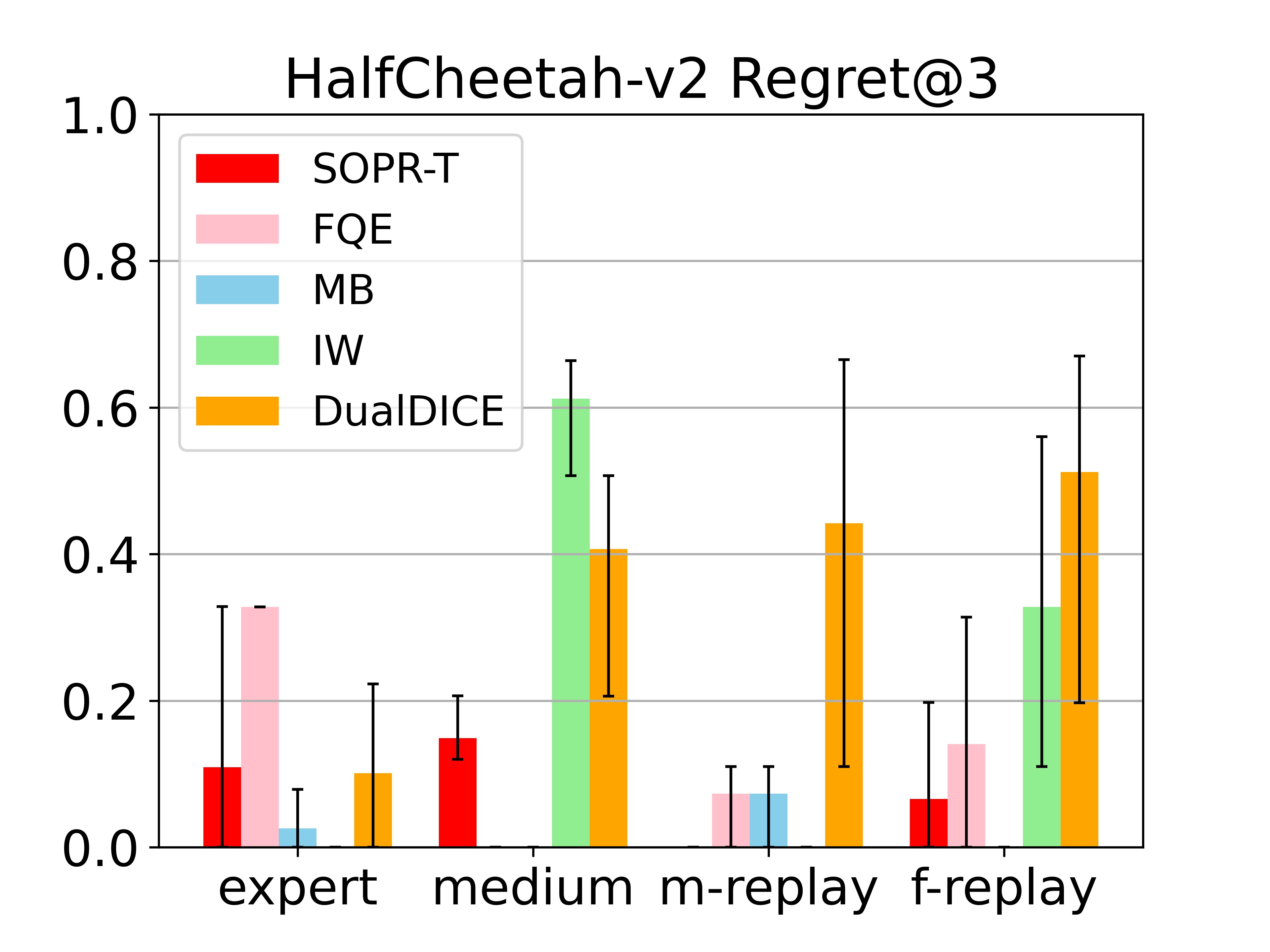}
    \end{minipage}
}
\subfigure[]
{
    \begin{minipage}[t]{0.3\linewidth}
	\centering
    \includegraphics[width = \linewidth]{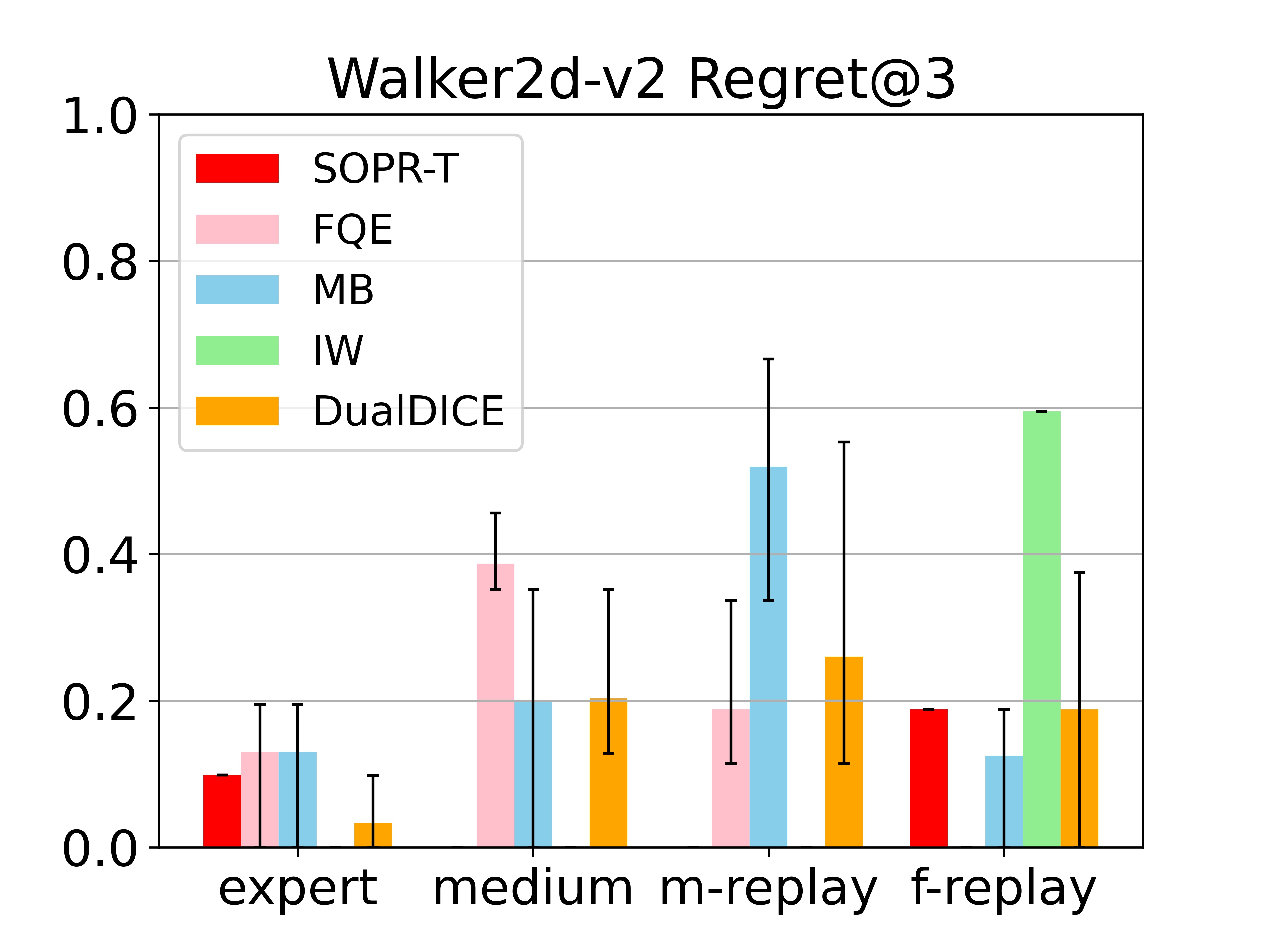}
    \end{minipage}
}
\caption {
Performance comparison (ranking the policies in Test Set I).
Top row: rank correlation.
Bottom row: regret@3.
}
\label{5_1_supp}
\end{figure}

In Section \ref{exp_testset_1}, we presented the performance of each algorithm on offline learned policies (Test Set I) in the Hopper game.
Here, Figure~~\ref{5_1_supp} shows all the results in three games.
As can be seen from the results, SOPR-T outperforms baseline OPE algorithms consistently on four tasks of the Hopper game.
Though in Walker2d and Halfcheetah, SOPR-T does not hold consistent superiority, it performs the most stably. That is, SOPR-T does not have negative rank correlation results in all the tasks, whereas all the baseline OPE algorithms have one or more negative correlation results.

\newpage
\subsection{Performance on Online Learned Policies (Corresponding to Section \ref{exp_testset_2})} \label{A.4}

\begin{figure}[H]
\centering
\subfigure[]
{
    \begin{minipage}[t]{0.3\linewidth}
	\centering
    \includegraphics[width = \linewidth]{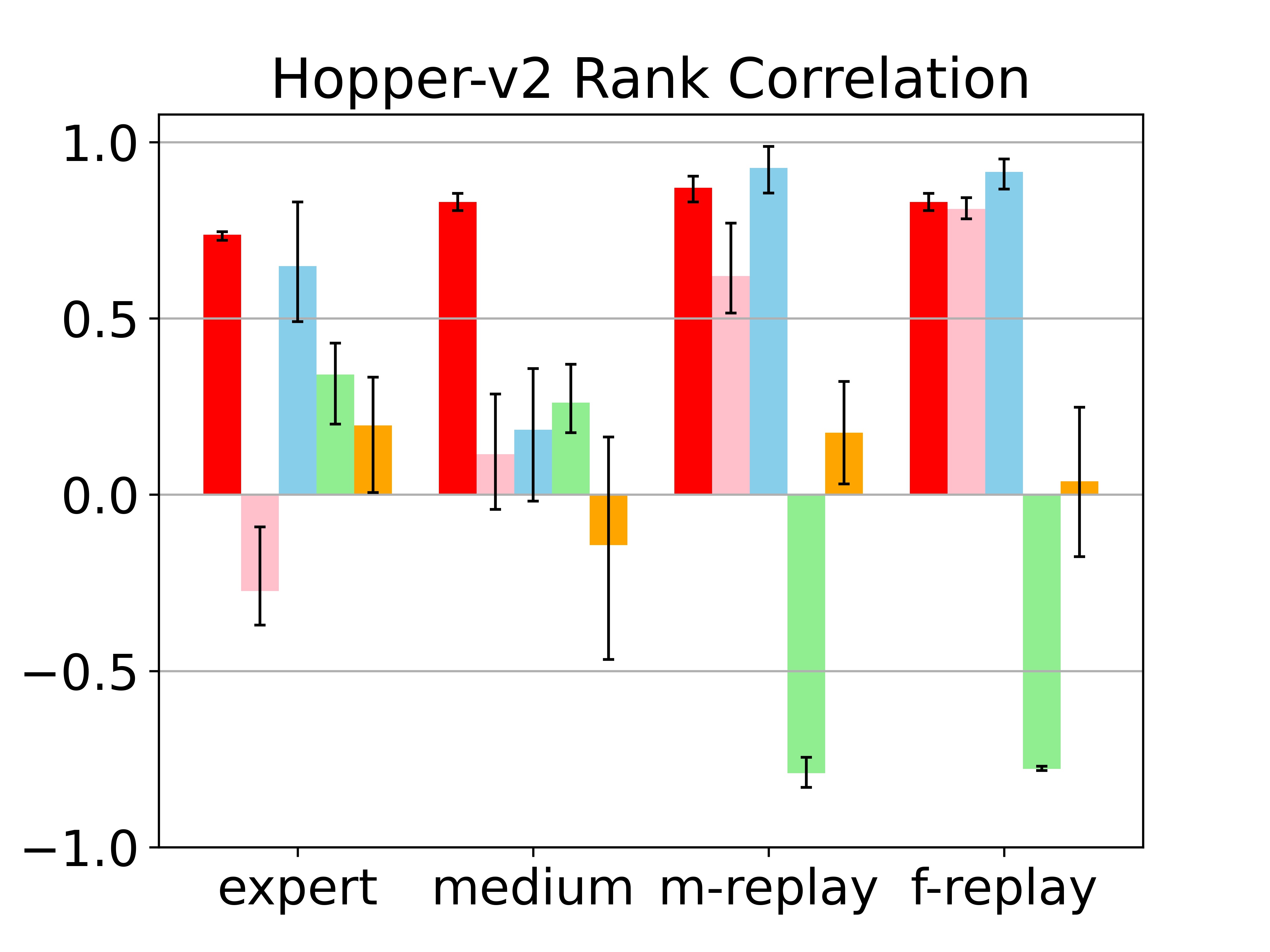}
    \end{minipage}
}
\subfigure[]
{
    \begin{minipage}[t]{0.3\linewidth}
	\centering
    \includegraphics[width = \linewidth]{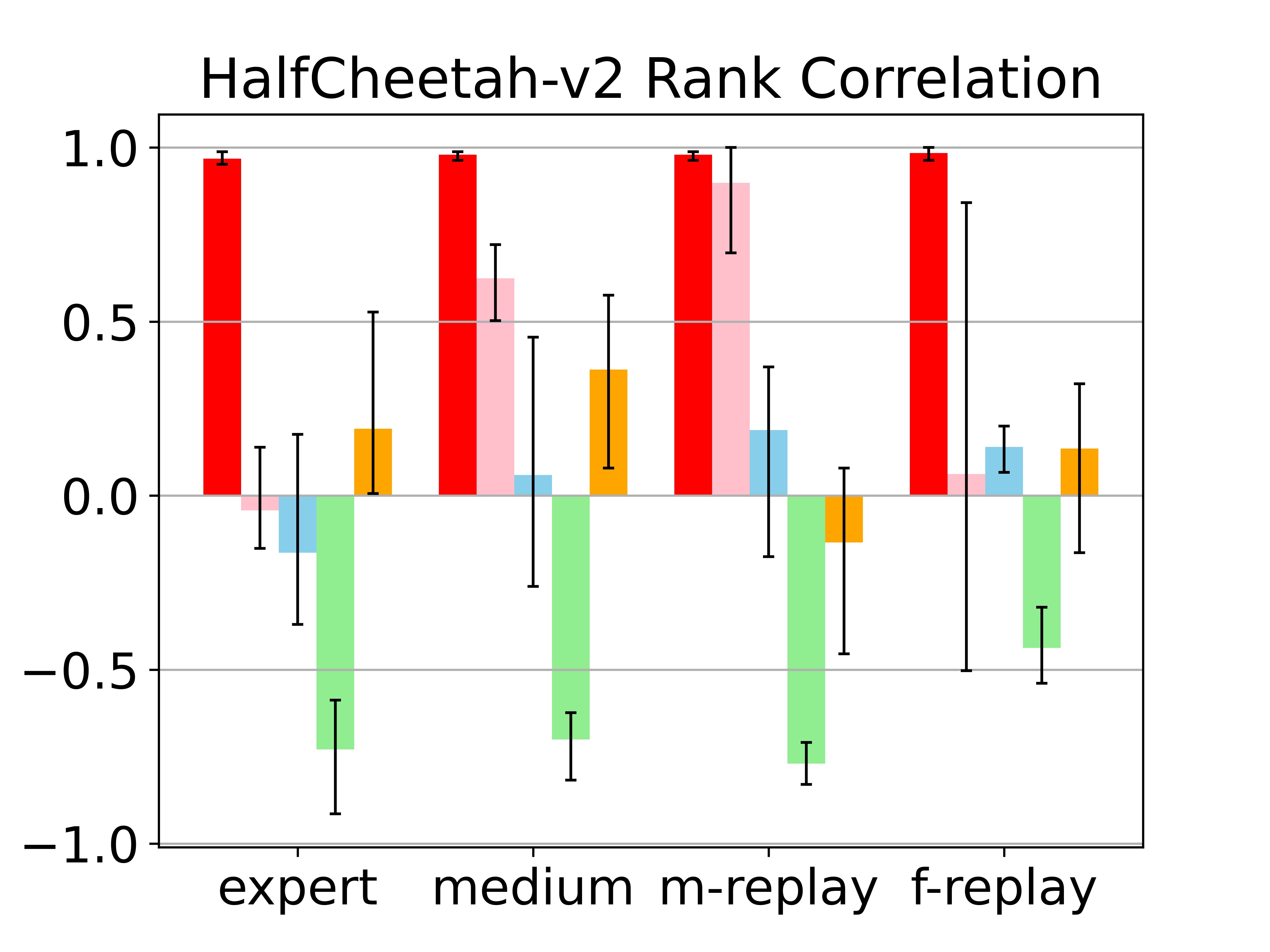}
    \end{minipage}
}
\subfigure[]
{
    \begin{minipage}[t]{0.3\linewidth}
	\centering
    \includegraphics[width = \linewidth]{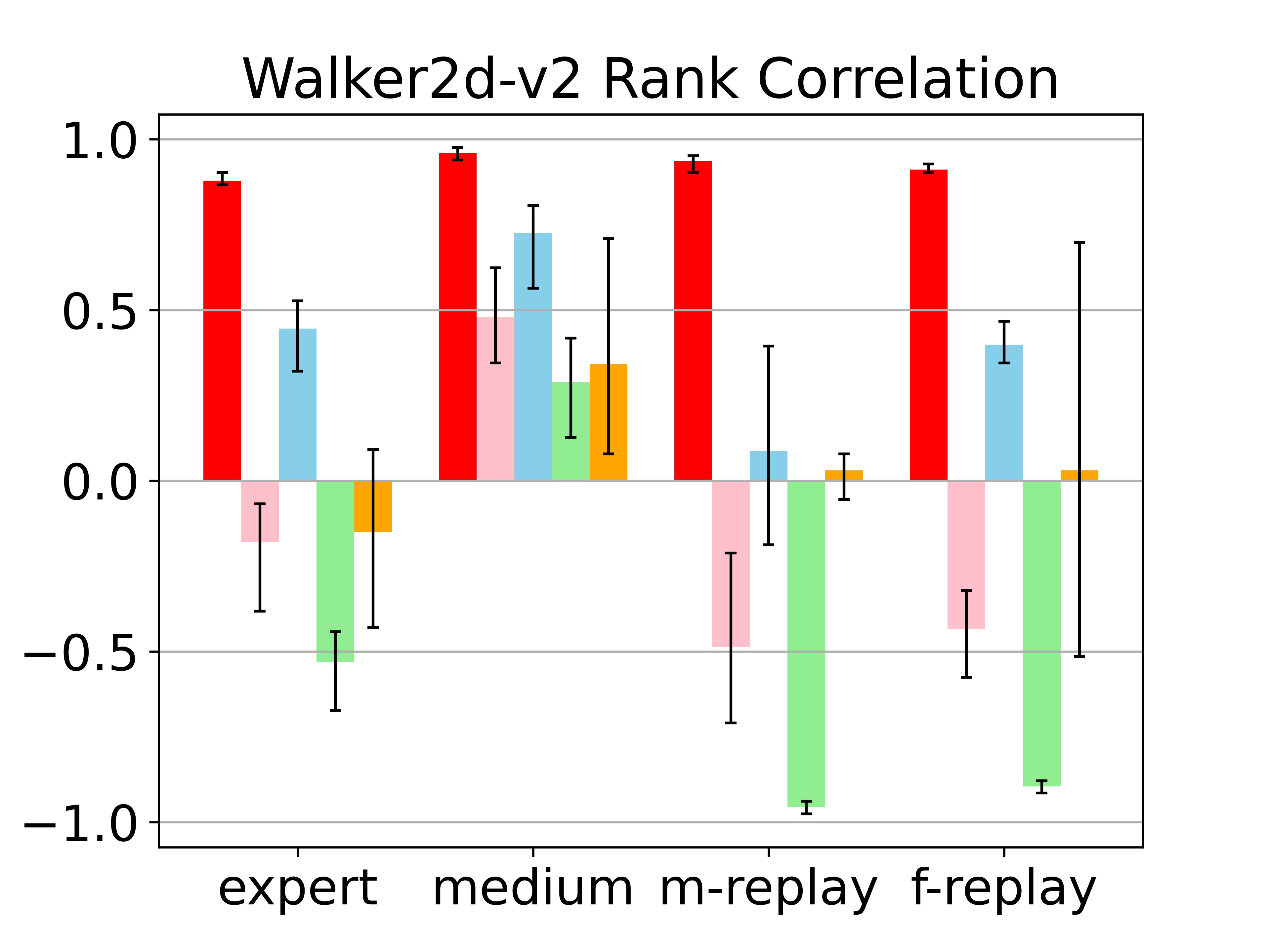}
    \end{minipage}
}

\subfigure[]
{
    \begin{minipage}[t]{0.3\linewidth}
	\centering
    \includegraphics[width = \linewidth]{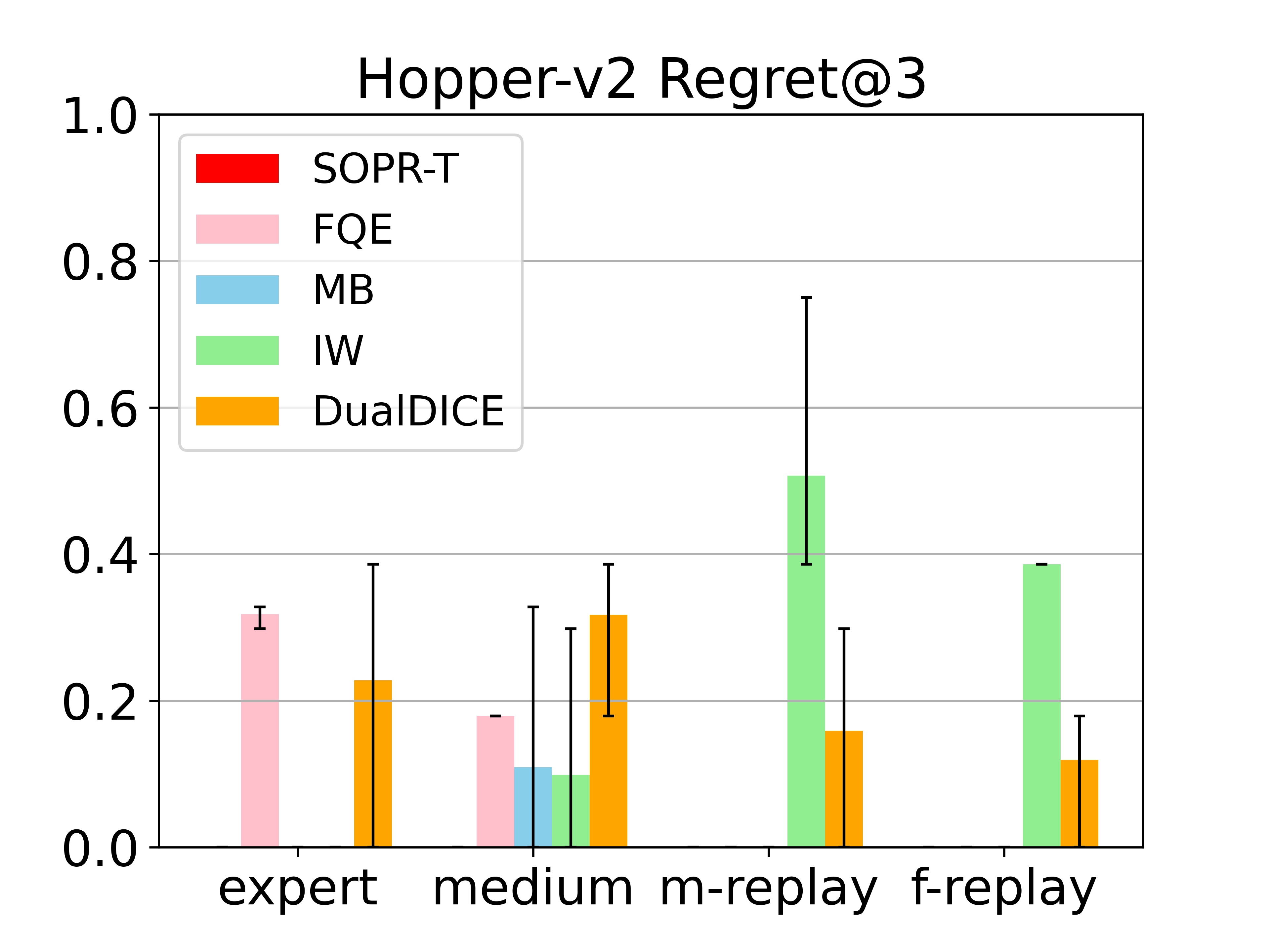}
    \end{minipage}
}
\subfigure[]
{
    \begin{minipage}[t]{0.3\linewidth}
	\centering
    \includegraphics[width = \linewidth]{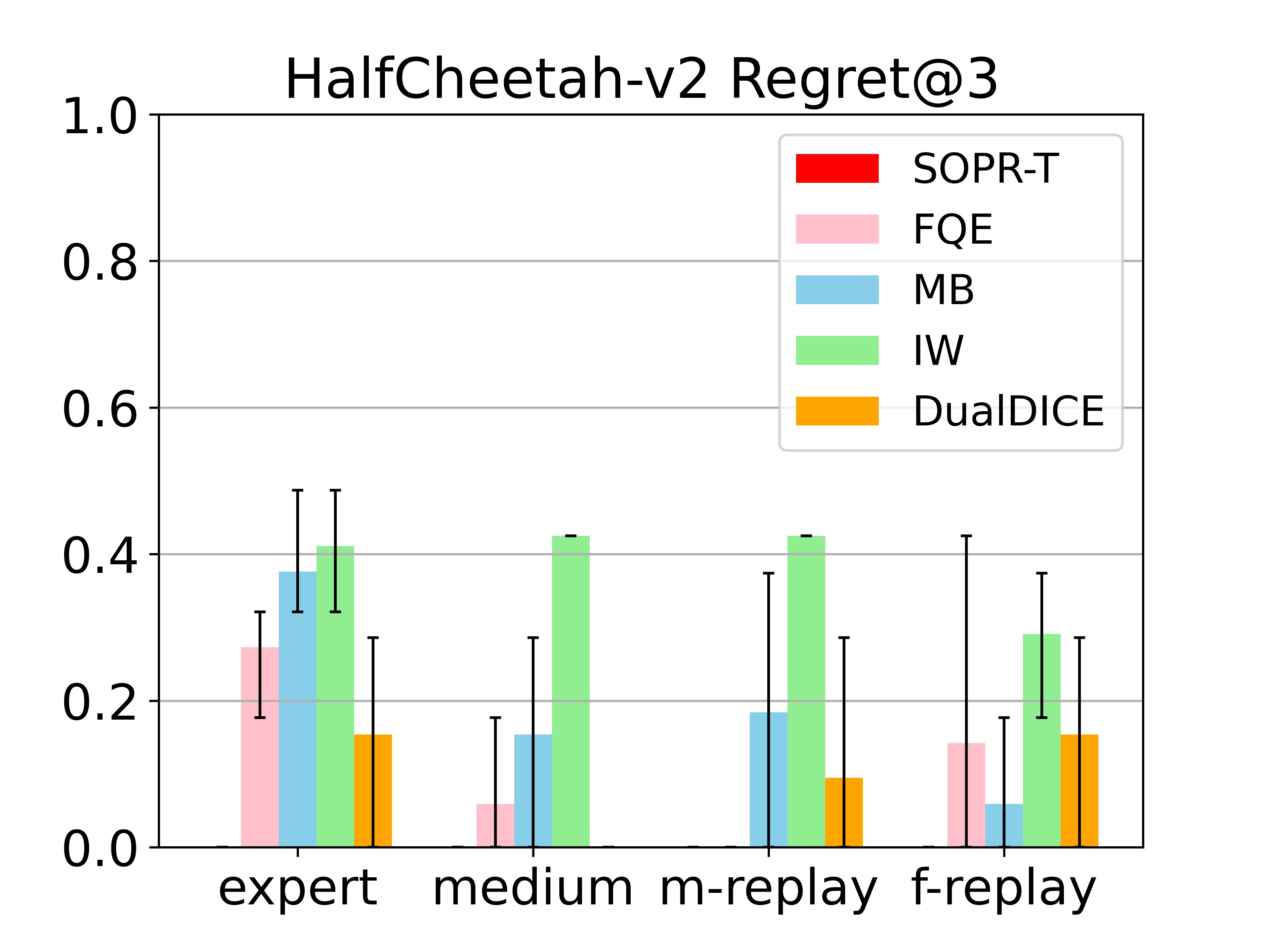}
    \end{minipage}
}
\subfigure[]
{
    \begin{minipage}[t]{0.3\linewidth}
	\centering
    \includegraphics[width = \linewidth]{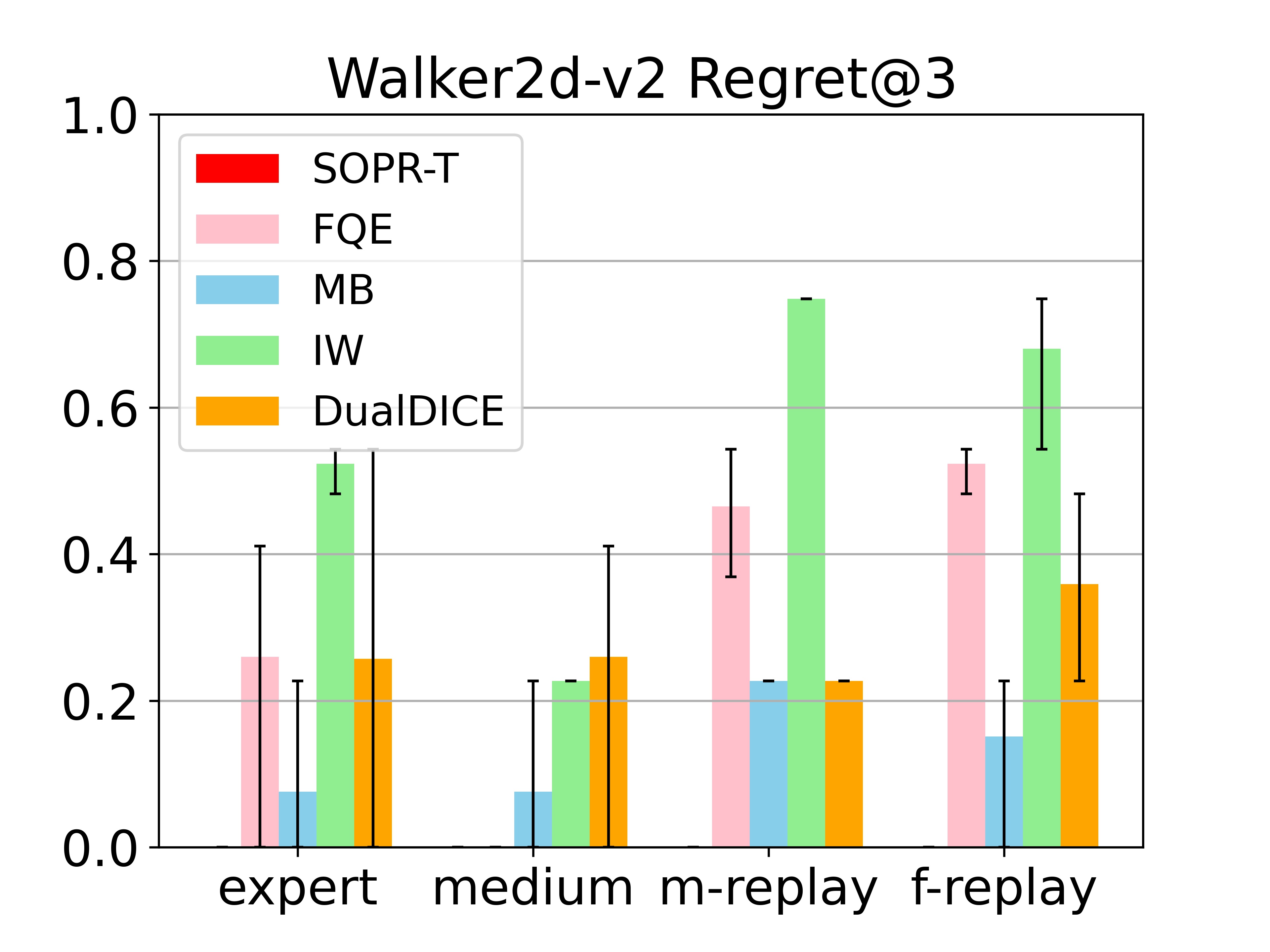}
    \end{minipage}
}
\caption {
Performance comparison (ranking the policies in Test Set II).
Top row: rank correlation.
Bottom row: regret@3.
}
\label{5_2_supp_1}
\end{figure}

In Section \ref{exp_testset_2}, we presented the performance of each algorithm on online learned policies (Test Set II) in the Hopper game.
Here, Figure~\ref{5_2_supp_1} shows all the results in three games.
As can be seen from the results, SOPR-T achieves very high rank correlation and zero regret values in all the tasks.
In addition, the variance caused by random seeds is very small.


\begin{figure}[!h]
\centering
\subfigure[]
{
    \begin{minipage}[t]{0.3\linewidth}
	\centering
    \includegraphics[width = \linewidth]{fig/Hopper_mse.jpg}
    \end{minipage}
}
\subfigure[]
{
    \begin{minipage}[t]{0.3\linewidth}
	\centering
    \includegraphics[width = \linewidth]{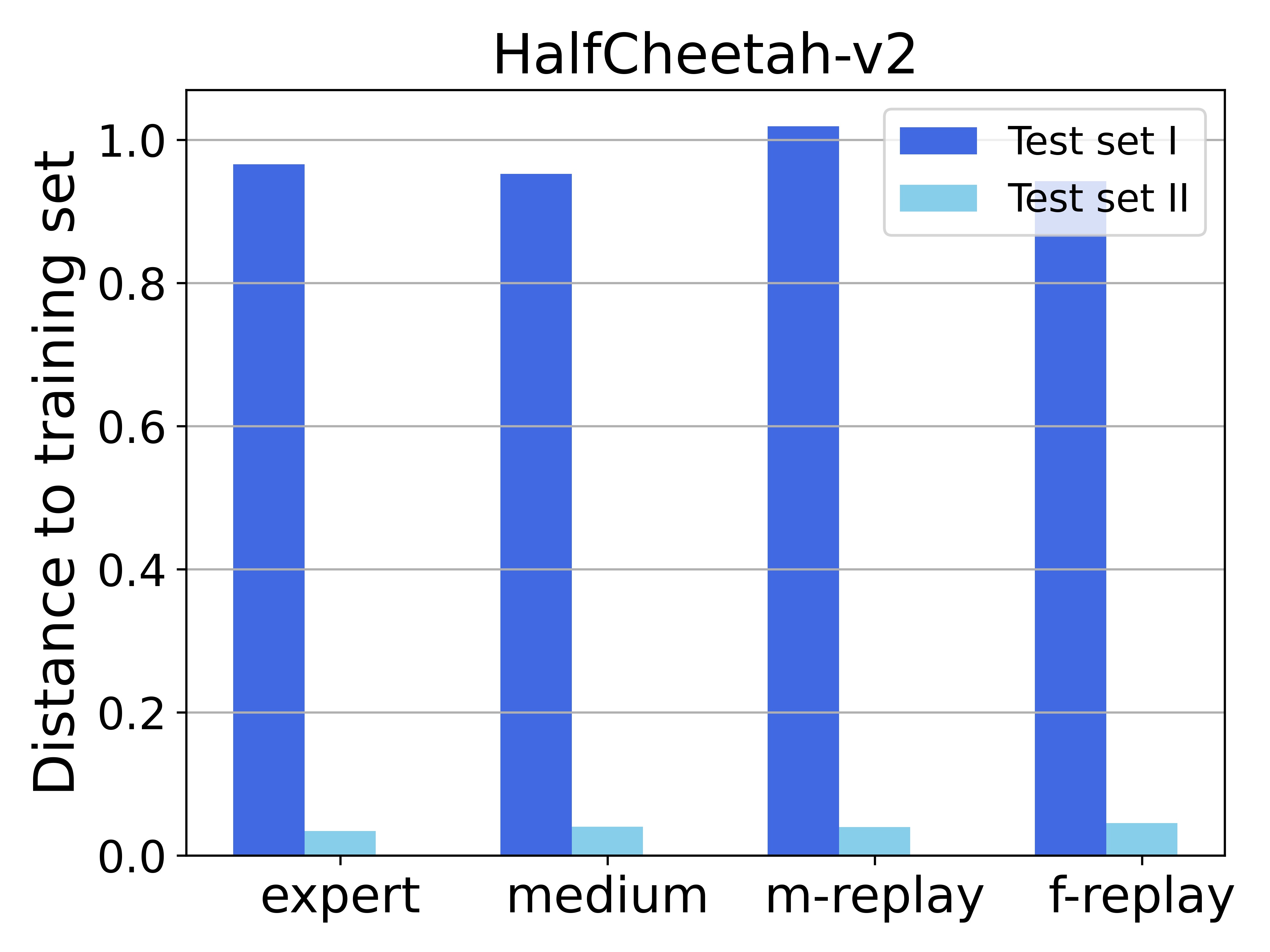}
    \end{minipage}
}
\subfigure[]
{
    \begin{minipage}[t]{0.3\linewidth}
	\centering
    \includegraphics[width = \linewidth]{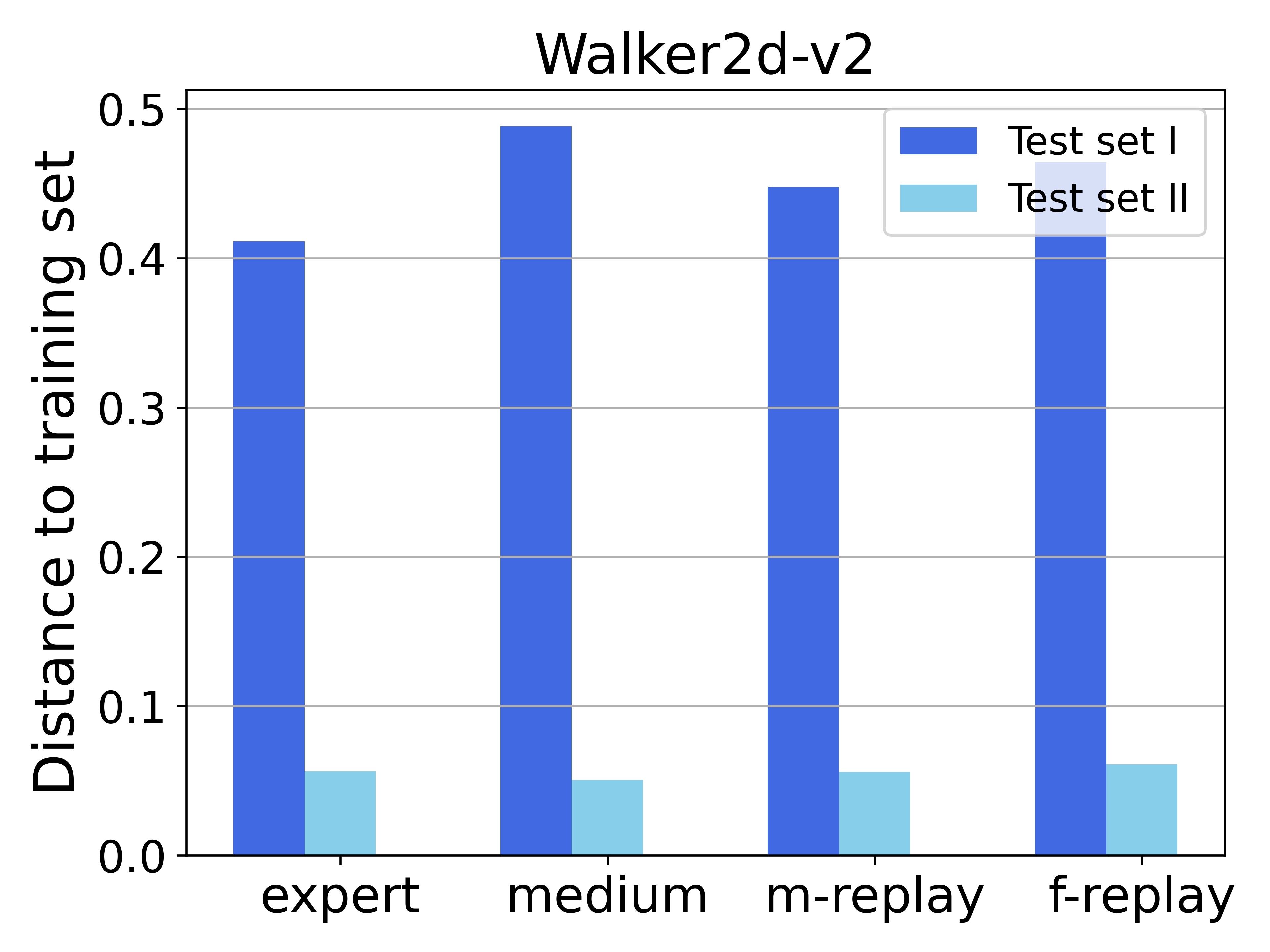}
    \end{minipage}
}
\caption {
Distance between training and test policy sets.
}
\label{5_2_supp_2}
\end{figure}

Figure~\ref{5_2_supp_2} presents the distribution distance between the training and test policy sets measured by Eq.~(\ref{eq_distance}).
As can be seen from the results, in different dataset cases, the distance between Test Set I and the training policy set is larger than the distance between Test Set II and the training policy set. The results reflect the influence of distribution distance between training and test policy sets on the performance of our algorithm. 


\newpage
\subsection{Additional Results of Section \ref{exp_further}} \label{A.5}

{\bf{Effect of Data Size}}

\begin{figure}[H]
\centering
\subfigure[]
{
    \begin{minipage}[t]{0.25\linewidth}
	\centering
    \includegraphics[width = \linewidth]{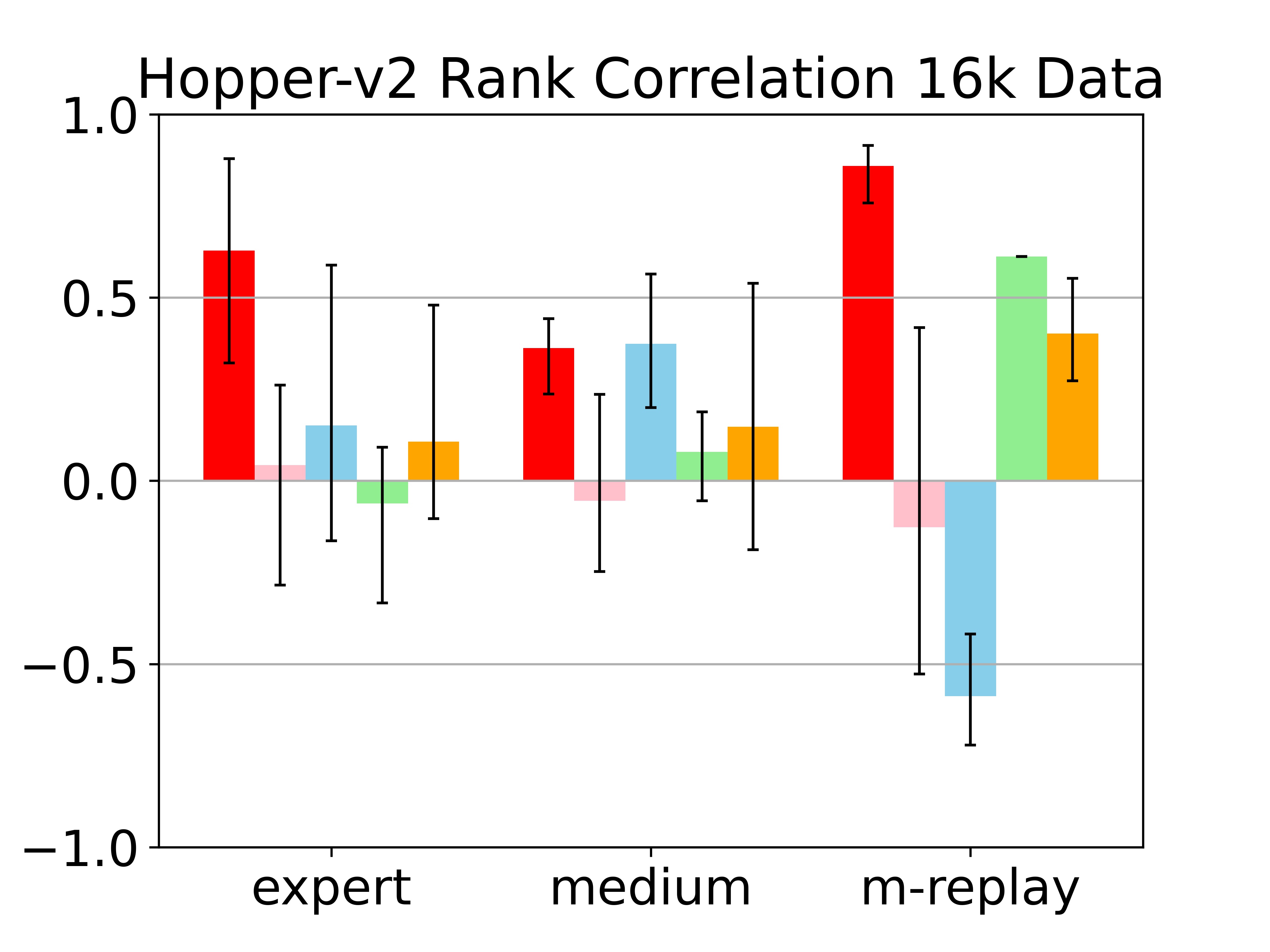}
    \end{minipage}
}
\subfigure[]
{
    \begin{minipage}[t]{0.25\linewidth}
	\centering
    \includegraphics[width = \linewidth]{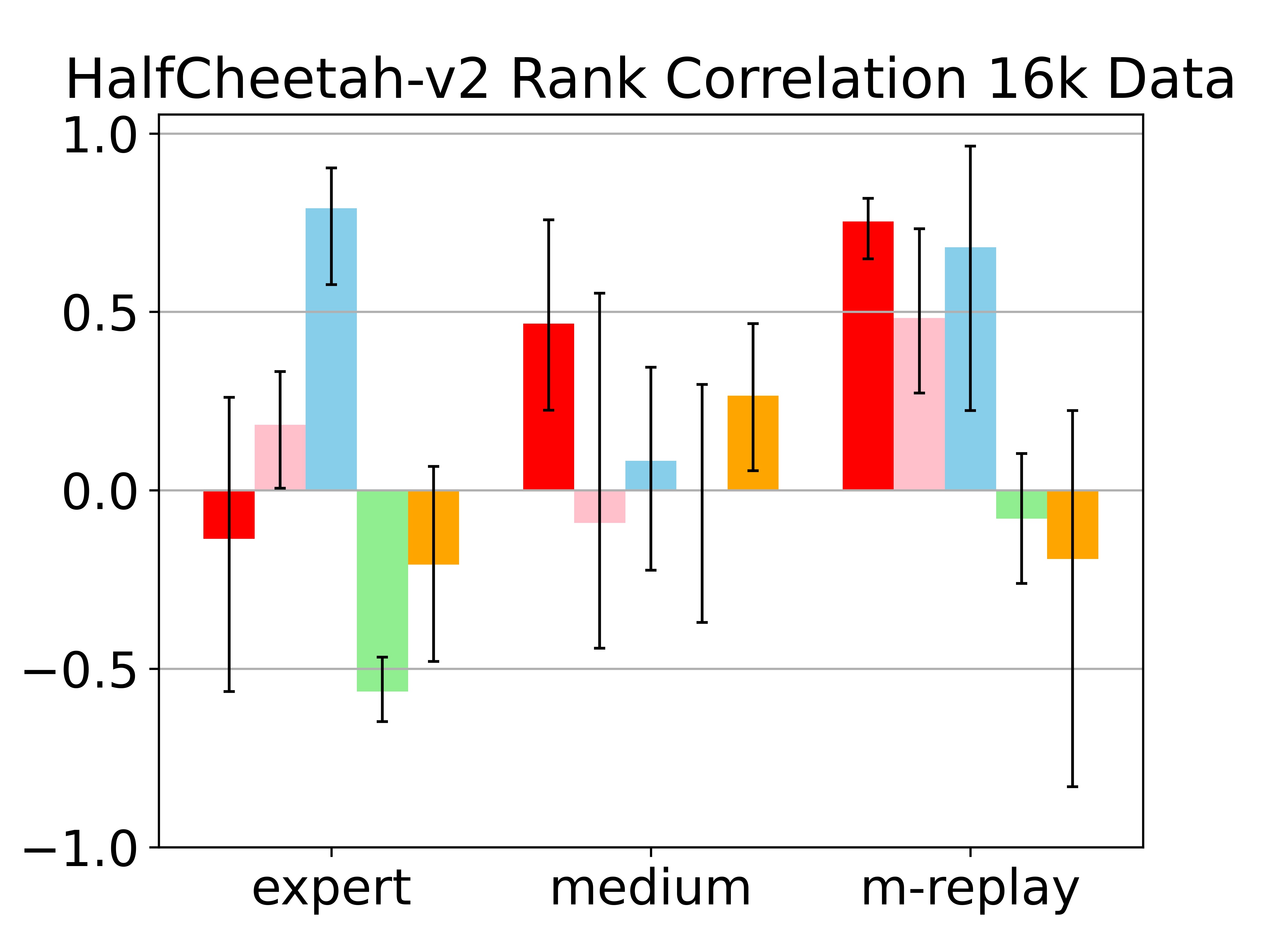}
    \end{minipage}
}
\subfigure[]
{
    \begin{minipage}[t]{0.25\linewidth}
	\centering
    \includegraphics[width = \linewidth]{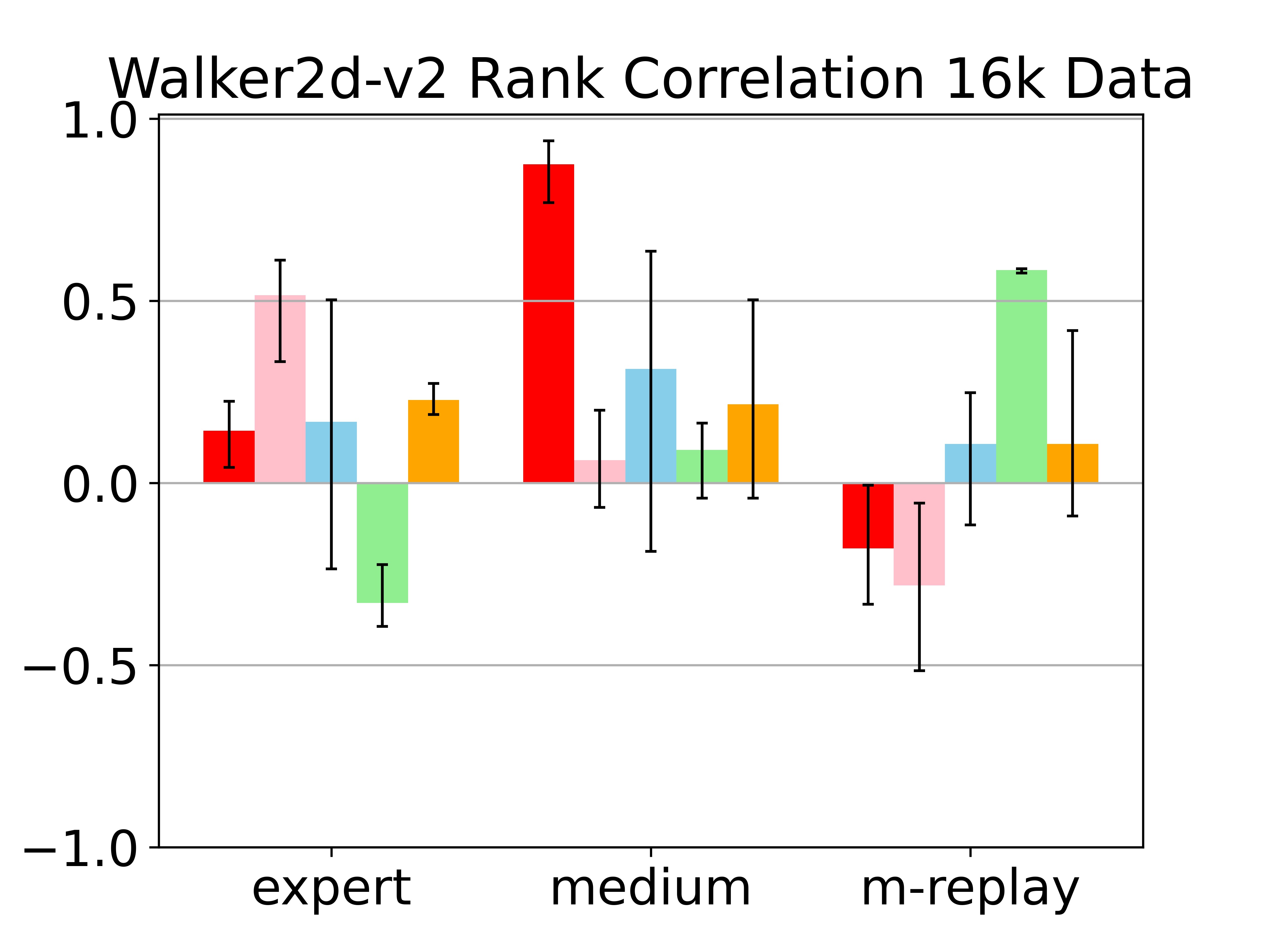}
    \end{minipage}
}

\subfigure[]
{
    \begin{minipage}[t]{0.25\linewidth}
	\centering
    \includegraphics[width = \linewidth]{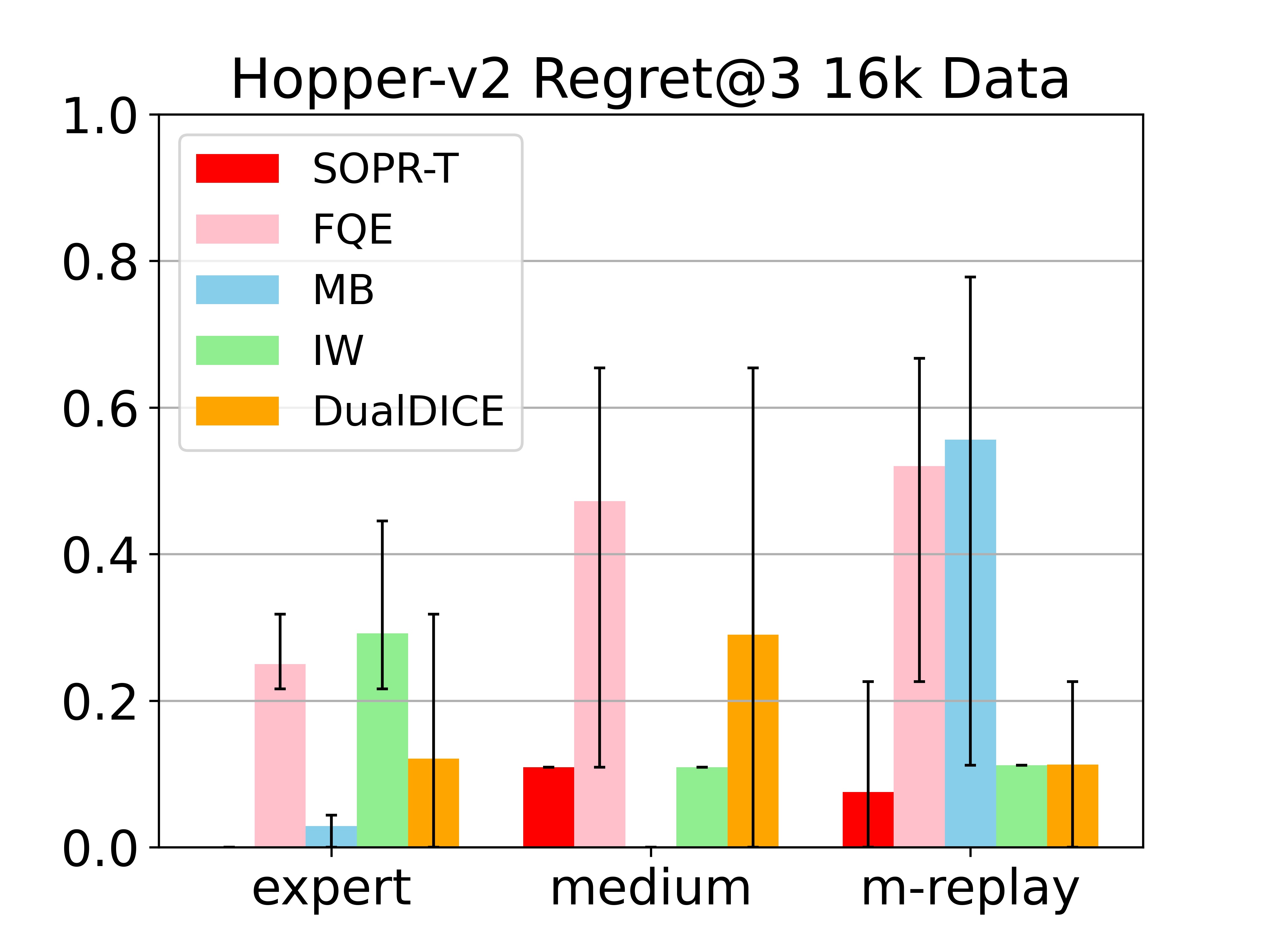}
    \end{minipage}
}
\subfigure[]
{
    \begin{minipage}[t]{0.25\linewidth}
	\centering
    \includegraphics[width = \linewidth]{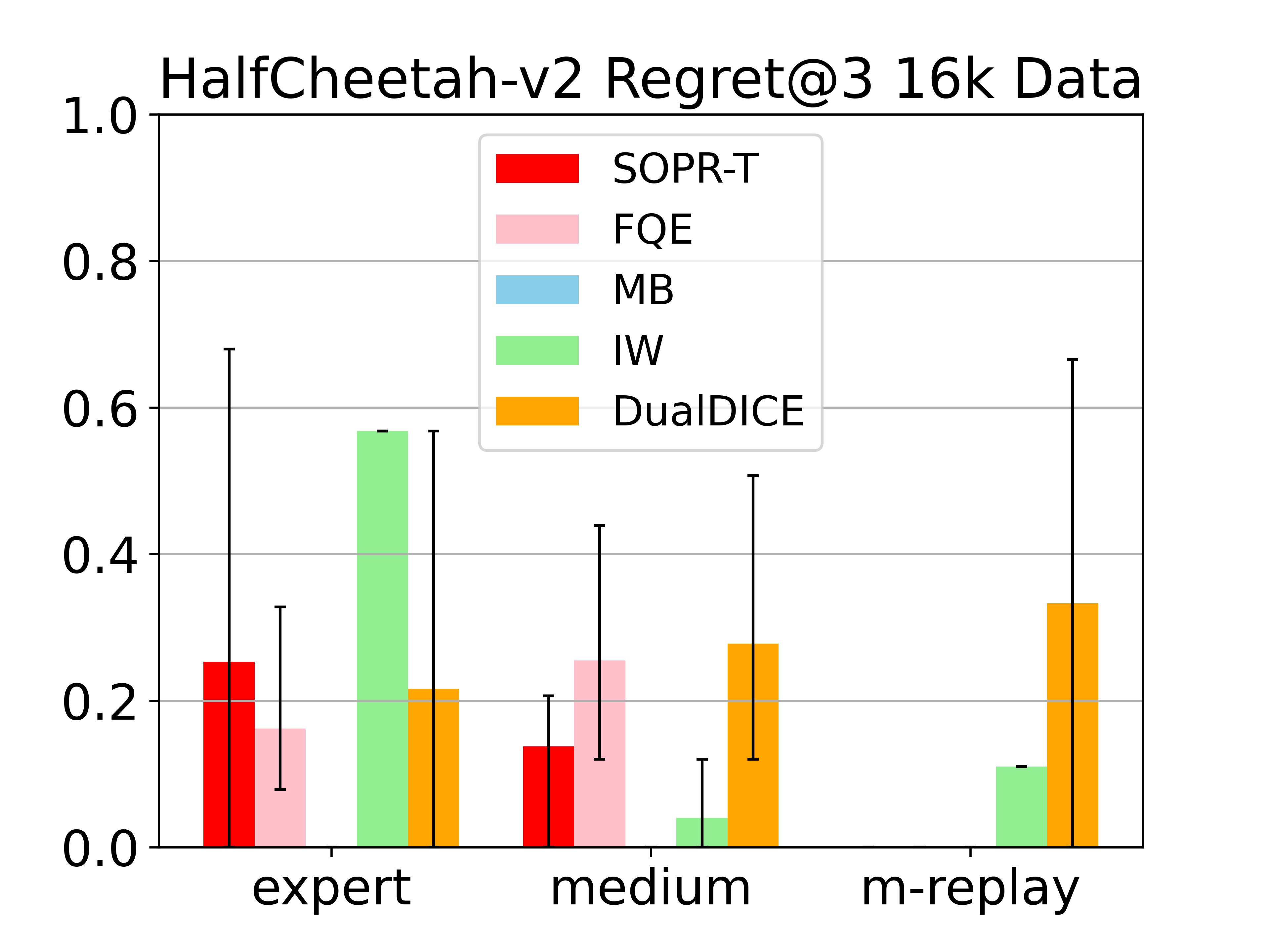}
    \end{minipage}
}
\subfigure[]
{
    \begin{minipage}[t]{0.25\linewidth}
	\centering
    \includegraphics[width = \linewidth]{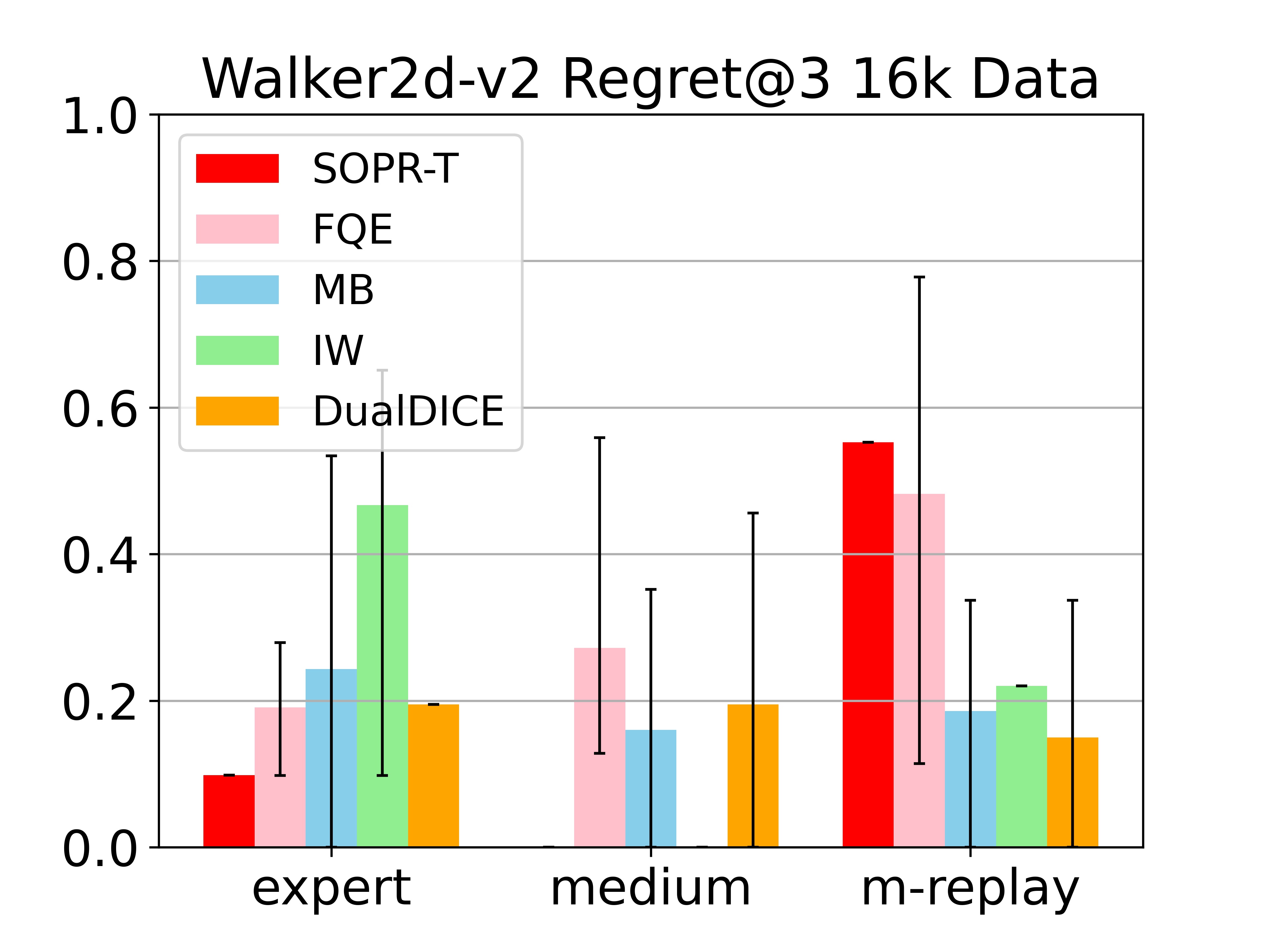}
    \end{minipage}
}
\caption {
Performance comparison (ranking the policies in Test Set I) when the size of dataset is 16k.
Top row: rank correlation.
Bottom row: regret@3.
}
\label{5_3_supp_16k_1}
\end{figure}

\begin{figure}[H]
\centering
\subfigure[]
{
    \begin{minipage}[t]{0.25\linewidth}
	\centering
    \includegraphics[width = \linewidth]{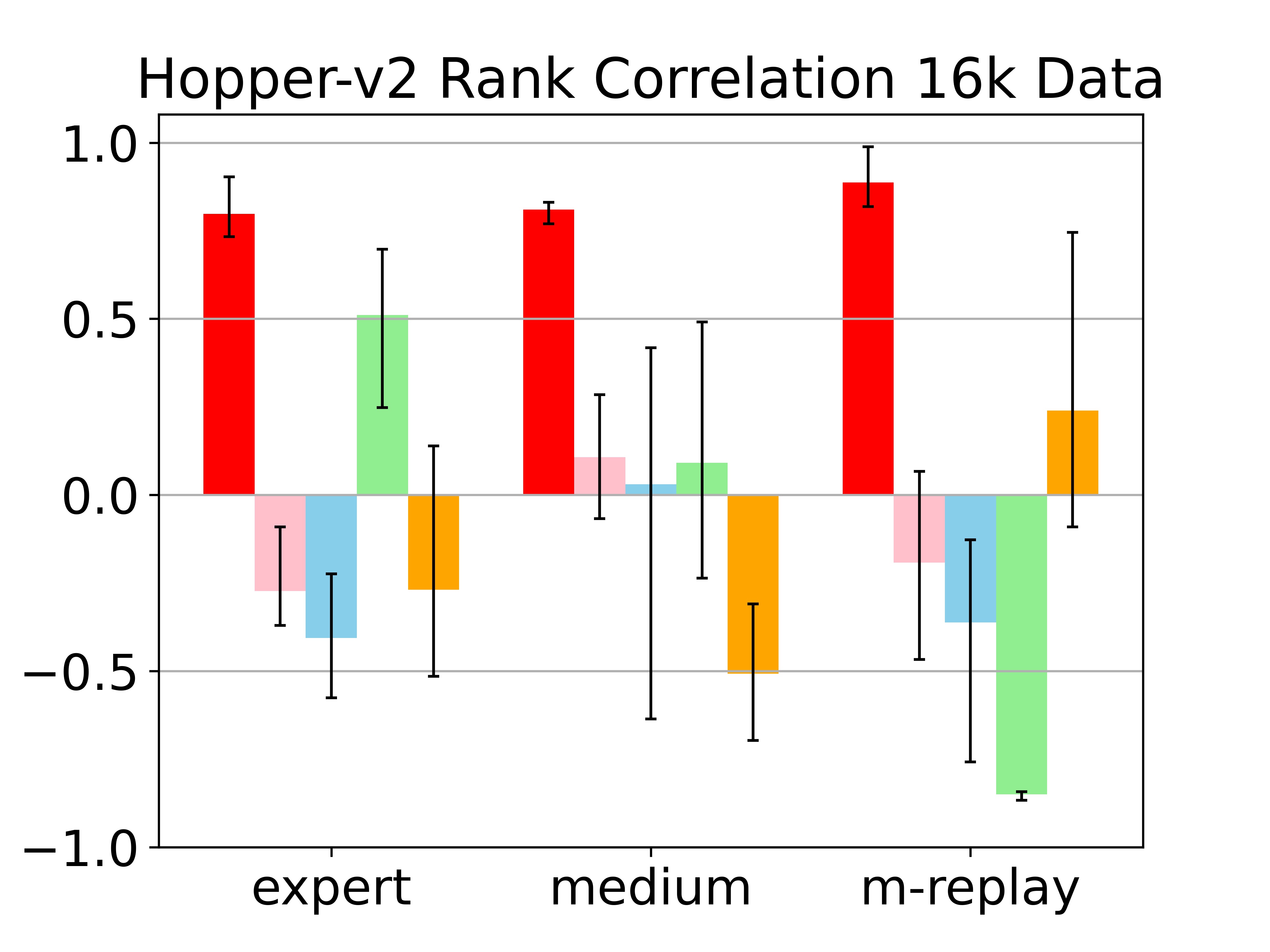}
    \end{minipage}
}
\subfigure[]
{
    \begin{minipage}[t]{0.25\linewidth}
	\centering
    \includegraphics[width = \linewidth]{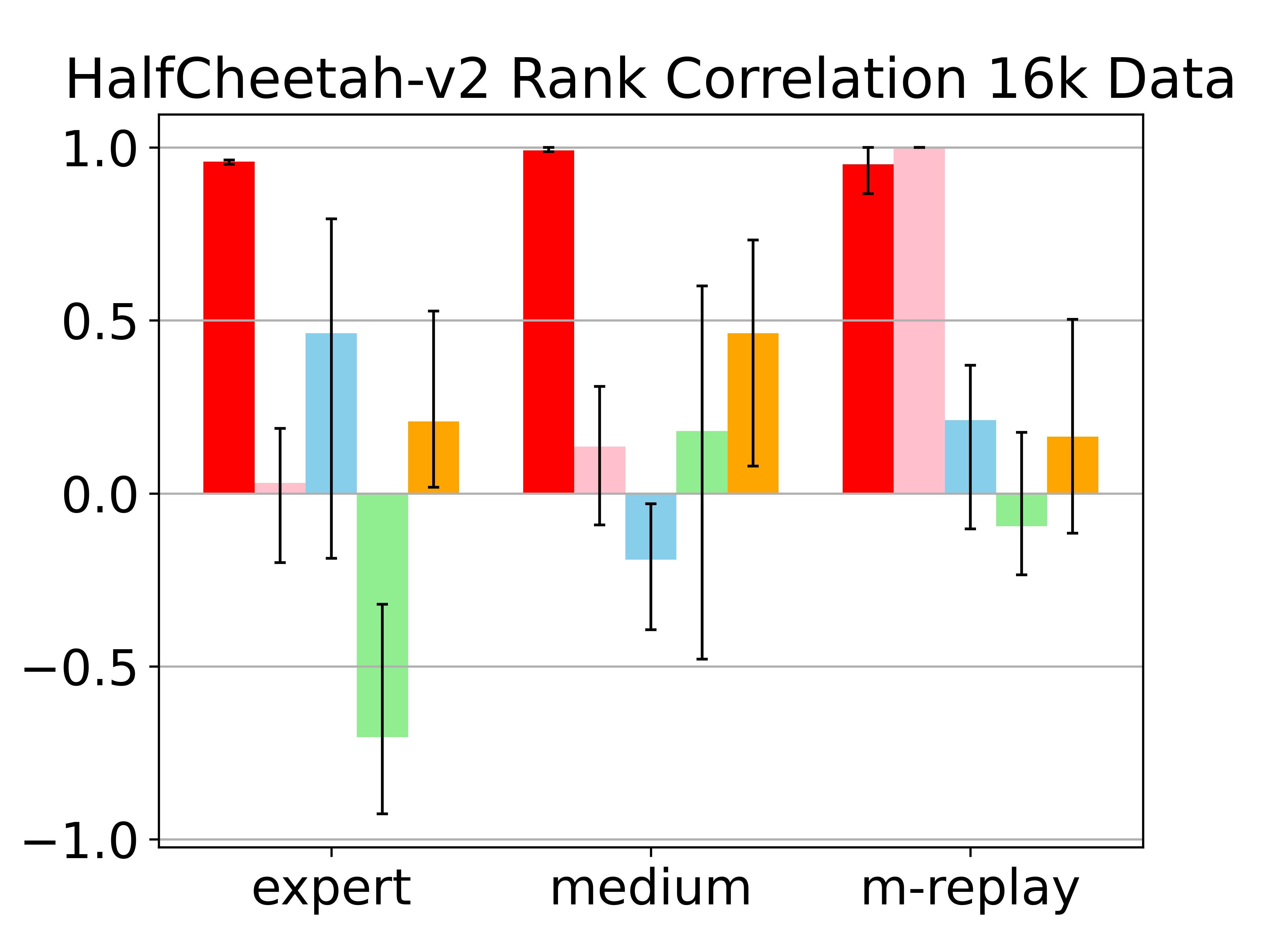}
    \end{minipage}
}
\subfigure[]
{
    \begin{minipage}[t]{0.25\linewidth}
	\centering
    \includegraphics[width = \linewidth]{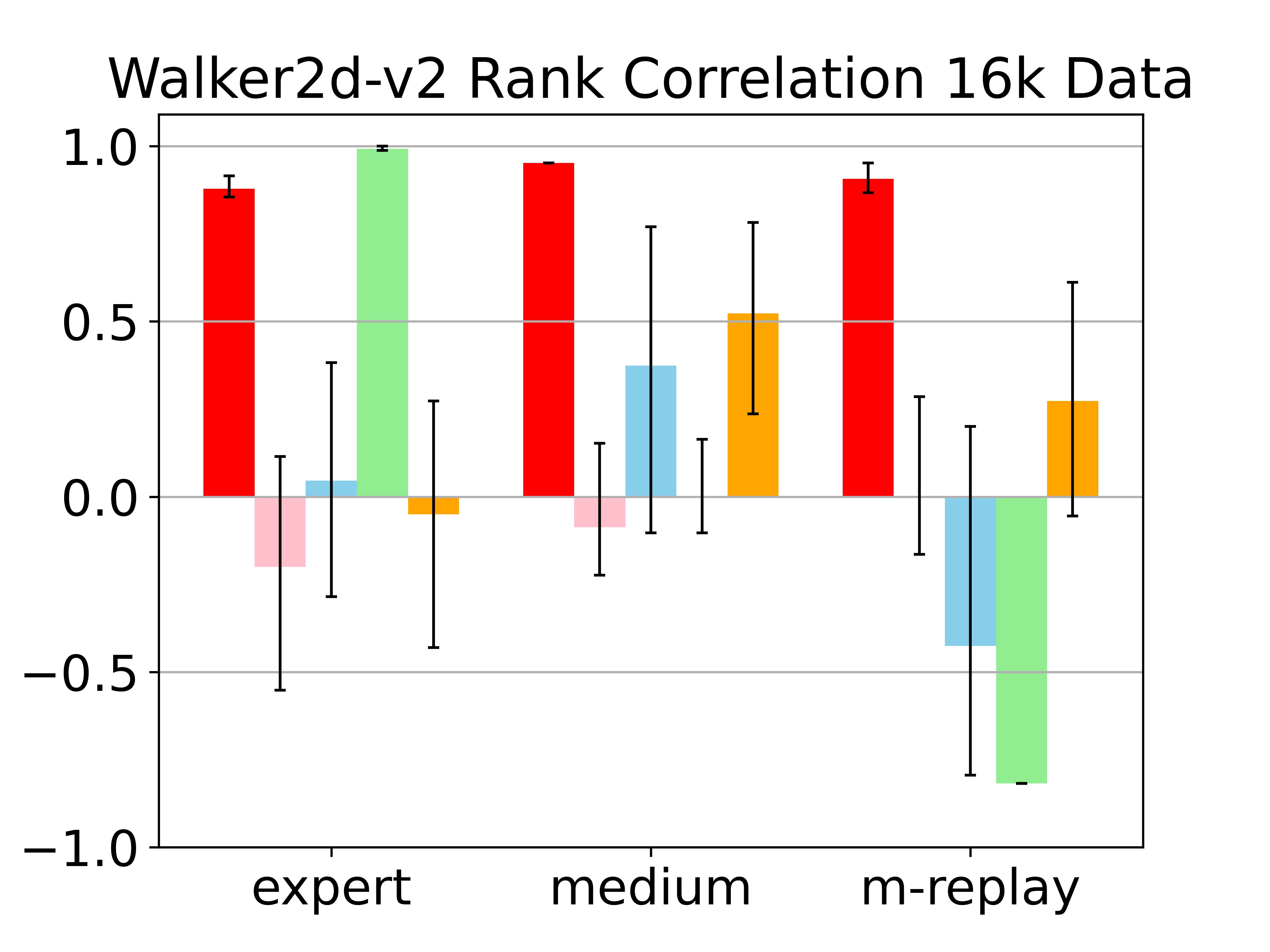}
    \end{minipage}
}

\subfigure[]
{
    \begin{minipage}[t]{0.25\linewidth}
	\centering
    \includegraphics[width = \linewidth]{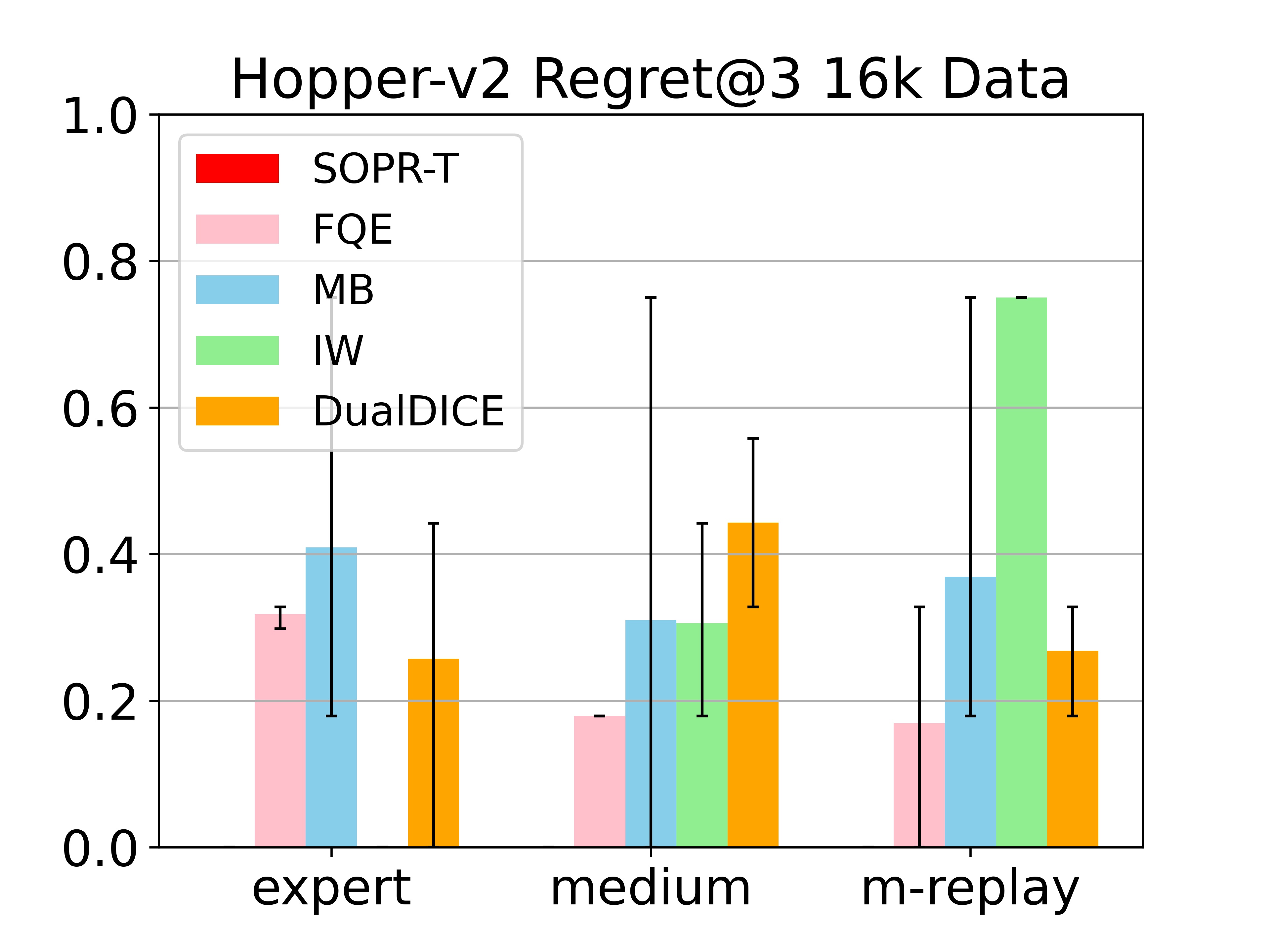}
    \end{minipage}
}
\subfigure[]
{
    \begin{minipage}[t]{0.25\linewidth}
	\centering
    \includegraphics[width = \linewidth]{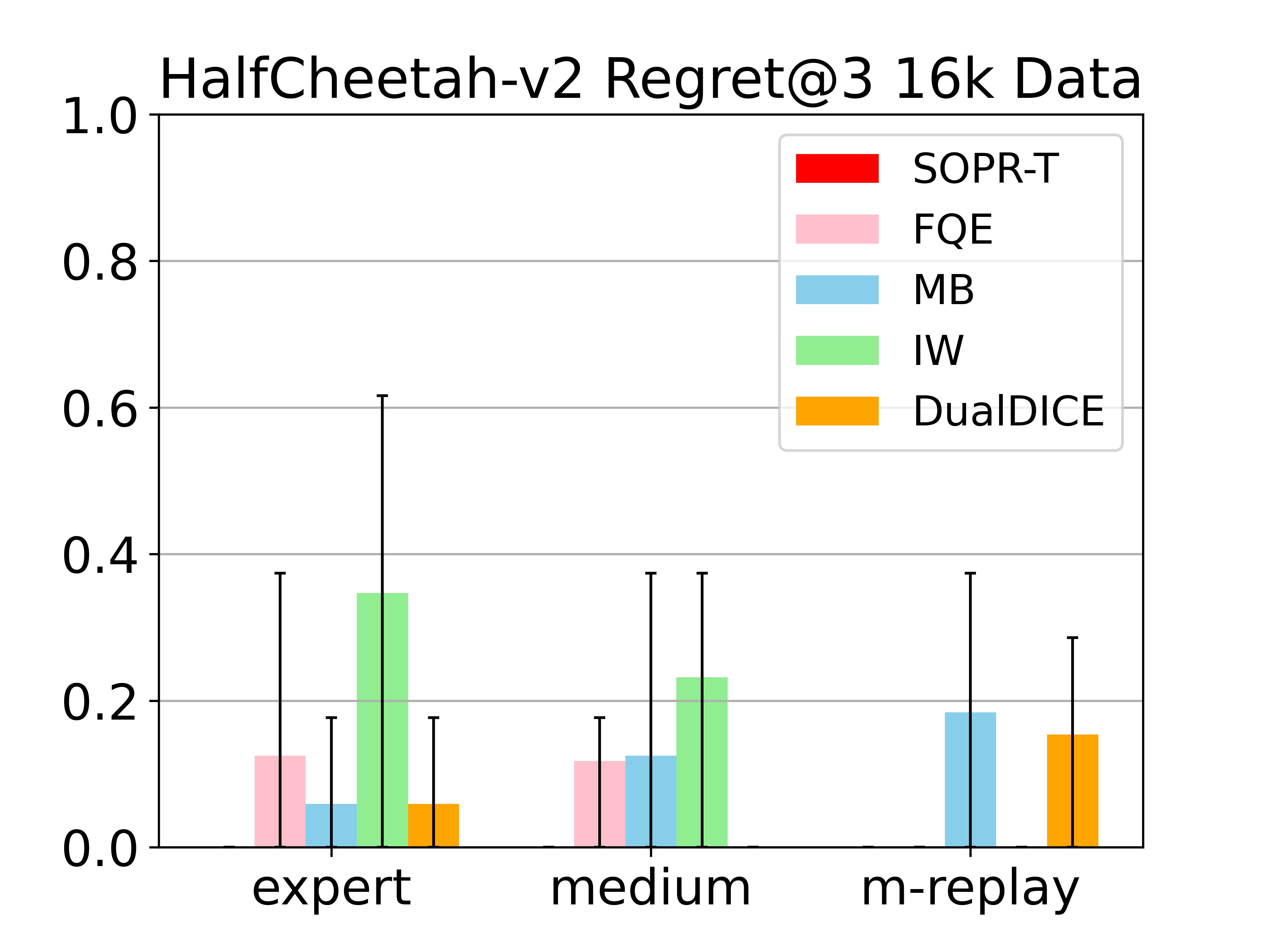}
    \end{minipage}
}
\subfigure[]
{
    \begin{minipage}[t]{0.25\linewidth}
	\centering
    \includegraphics[width = \linewidth]{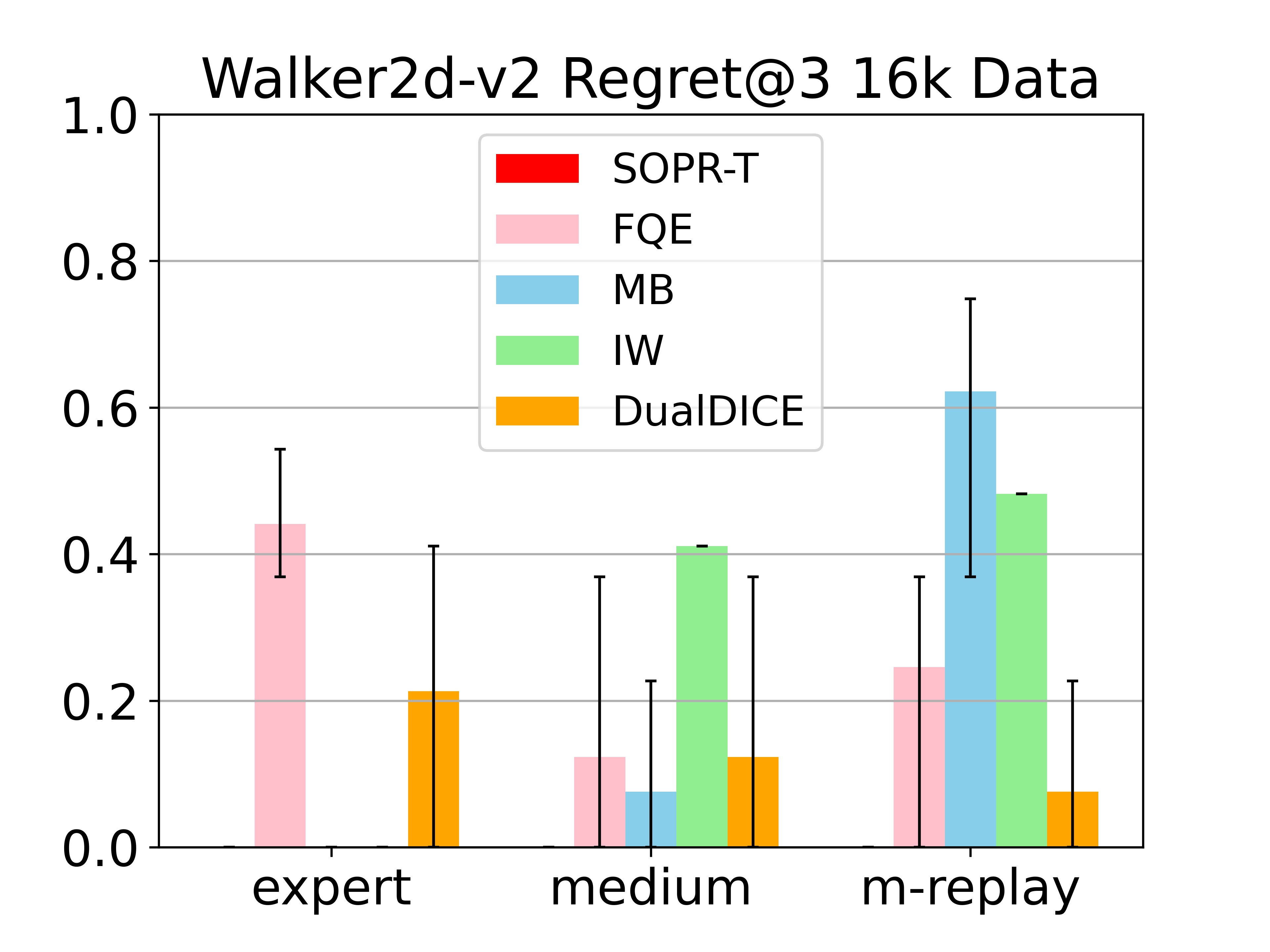}
    \end{minipage}
}
\caption {
Performance comparison (ranking the policies in Test Set II) when the size of dataset is 16k.
Top row: rank correlation.
Bottom row: regret@3.
}
\label{5_3_supp_16k_2}
\end{figure}

In Section \ref{exp_further}, we presented the performance ranking statistics of 5 algorithms on all tasks when the size of off-policy dataset is 16k.
Here, Figure~\ref{5_3_supp_16k_1} and Figure~\ref{5_3_supp_16k_2} show detailed performance of each algorithm on each individual task.
As shown in Figure~\ref{5_3_supp_16k_1} and Figure~\ref{5_3_supp_16k_2}, SOPR-T outperforms baseline OPE algorithms in most of the tasks.

{\bf{Transformer Encoder vs. MLP Encoder}}

\begin{figure}[H]
\centering
\subfigure[]
{
    \begin{minipage}[t]{0.27\linewidth}
	\centering
    \includegraphics[width = \linewidth]{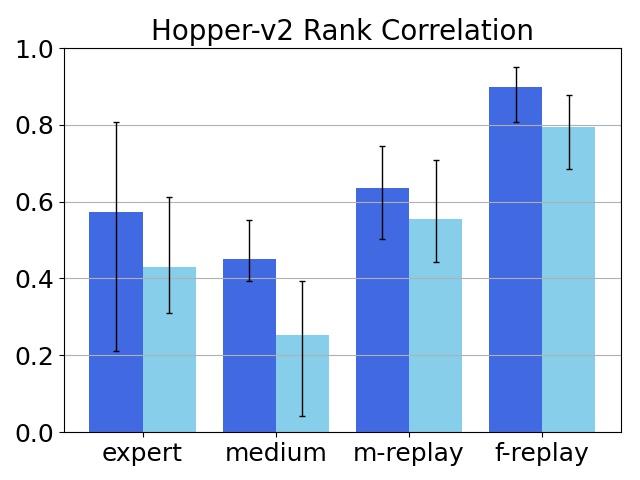}
    \end{minipage}
}
\subfigure[]
{
    \begin{minipage}[t]{0.27\linewidth}
	\centering
    \includegraphics[width = \linewidth]{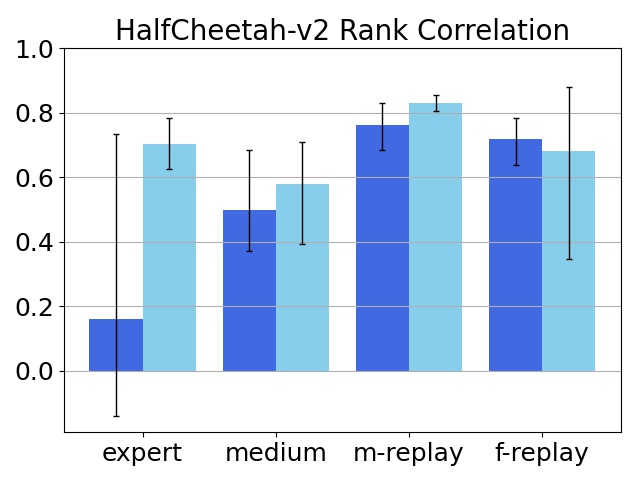}
    \end{minipage}
}
\subfigure[]
{
    \begin{minipage}[t]{0.27\linewidth}
	\centering
    \includegraphics[width = \linewidth]{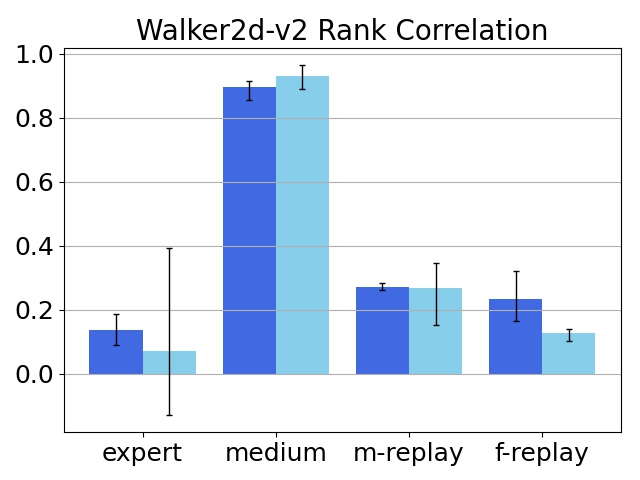}
    \end{minipage}
}

\subfigure[]
{
    \begin{minipage}[t]{0.27\linewidth}
	\centering
    \includegraphics[width = \linewidth]{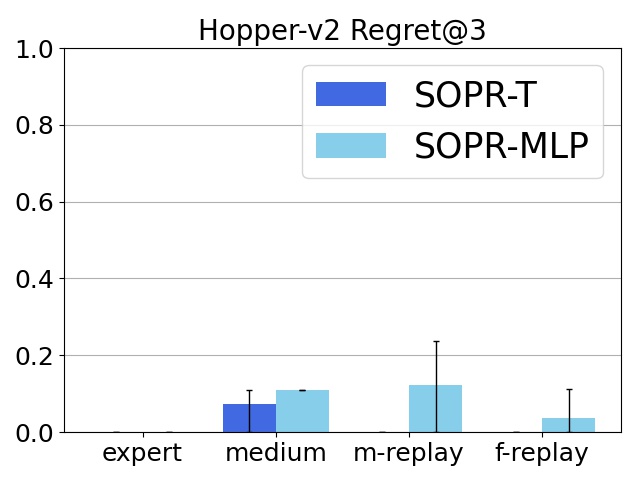}
    \end{minipage}
}
\subfigure[]
{
    \begin{minipage}[t]{0.27\linewidth}
	\centering
    \includegraphics[width = \linewidth]{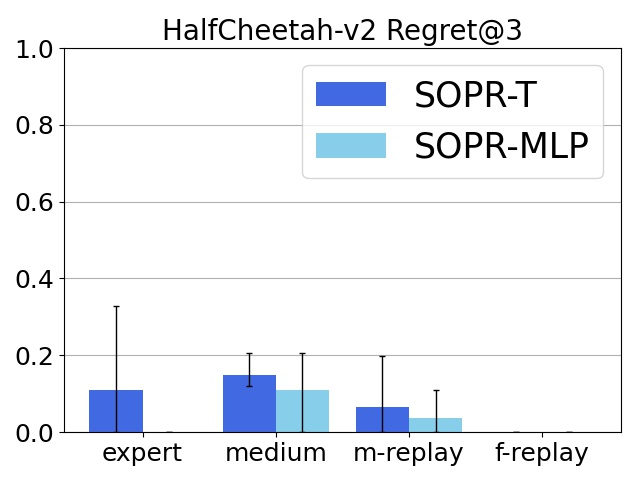}
    \end{minipage}
}
\subfigure[]
{
    \begin{minipage}[t]{0.27\linewidth}
	\centering
    \includegraphics[width = \linewidth]{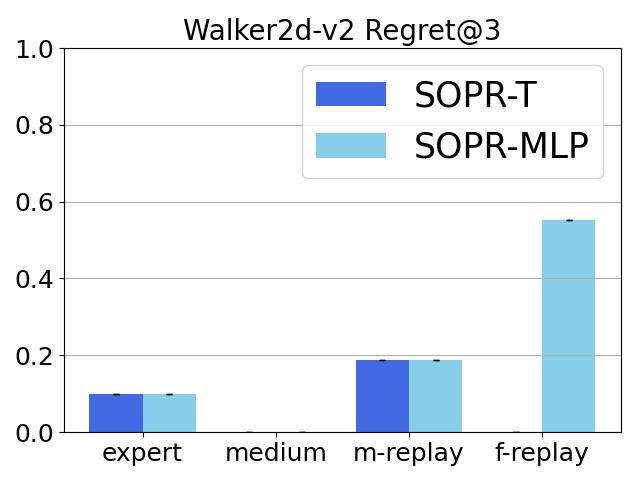}
    \end{minipage}
}
\caption {
Performance comparison between SOPR-T and SOPR-MLP (Test Set I).
Top row: rank correlation.
Bottom row: regret@3.
}
\label{5_3_supp_2.1}
\end{figure}

\begin{figure}[H]
\centering
\subfigure[]
{
    \begin{minipage}[t]{0.27\linewidth}
	\centering
    \includegraphics[width = \linewidth]{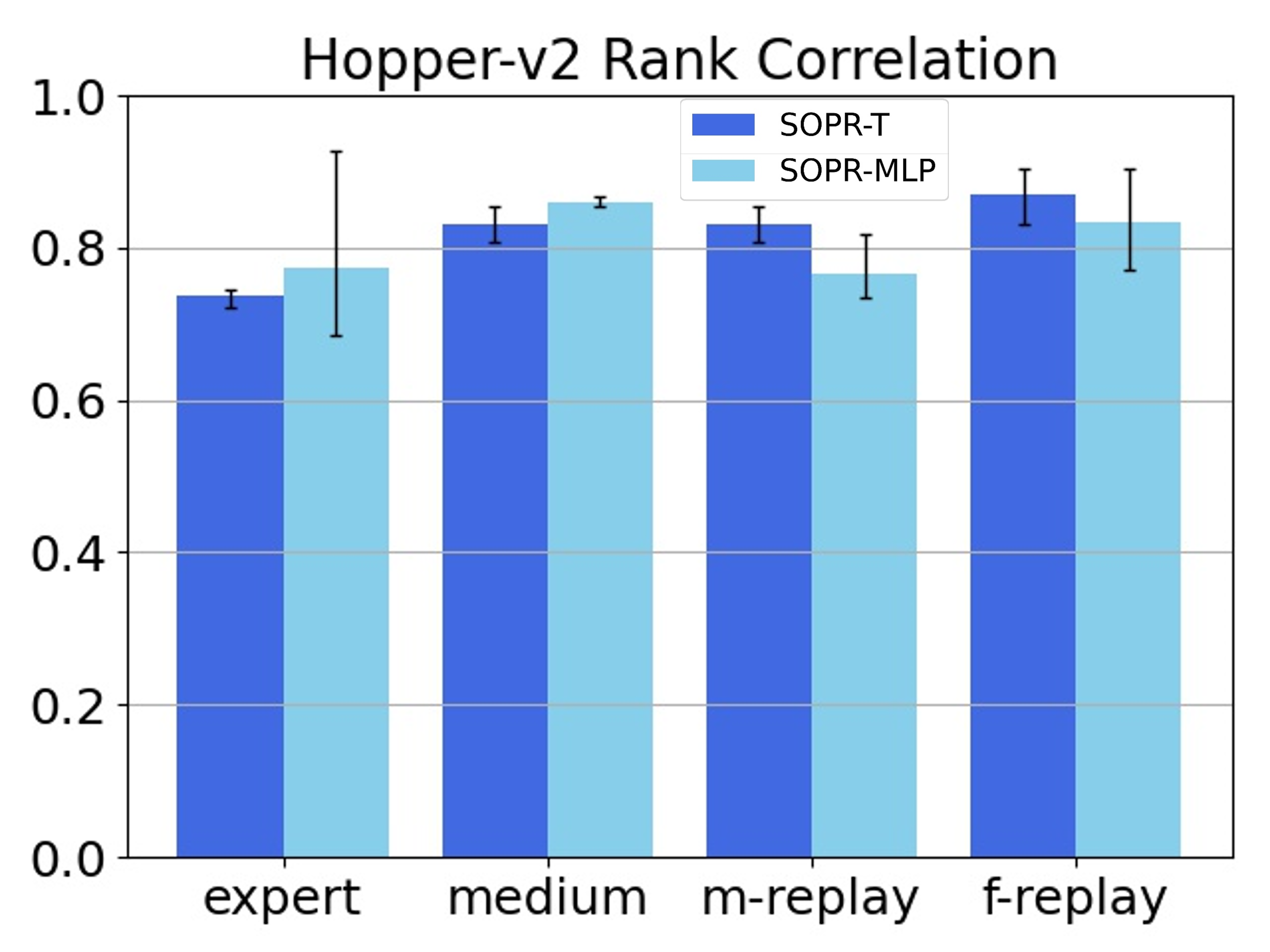}
    \end{minipage}
}
\subfigure[]
{
    \begin{minipage}[t]{0.27\linewidth}
	\centering
    \includegraphics[width = \linewidth]{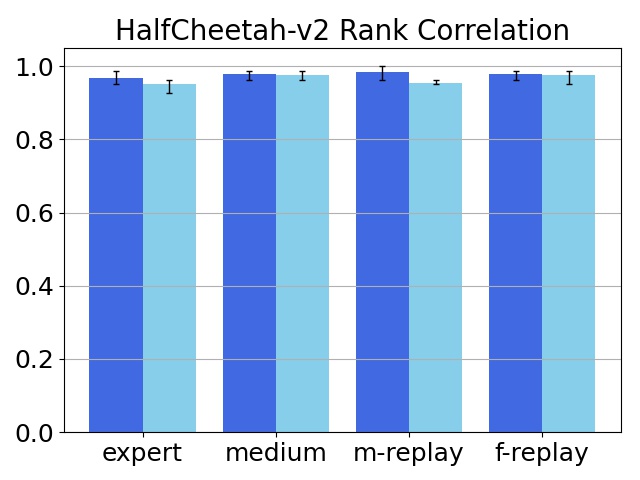}
    \end{minipage}
}
\subfigure[]
{
    \begin{minipage}[t]{0.27\linewidth}
	\centering
    \includegraphics[width = \linewidth]{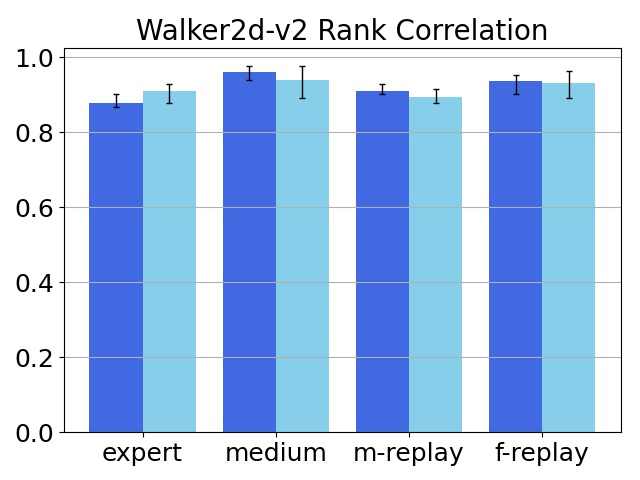}
    \end{minipage}
}
\caption { Performance comparison between SOPR-T and SOPR-MLP (Test Set II, rank correlation).
}
\label{5_3_supp_2.3}
\end{figure}

In Section \ref{exp_further}, we investigated the performance difference between SOPR-T and SOPR-MLP.
Here, the detailed performance of SOPR-T and SOPR-MLP on each individual task is shown in Figure~\ref{5_3_supp_2.1} (Test Set I) and Figure~\ref{5_3_supp_2.3} (Test Set II).
Because the regret values of both SOPR-T and SOPR-MLP in Test Set II in all the tasks are zero, the regret value results are not presented and we only show rank correlation results in Figure~\ref{5_3_supp_2.3}.
As can be seen from the results, SOPR-T outperforms SOPR-MLP in most tasks in terms of both rank correlation and regret value.

{\bf{Effect of Using Different Number of Subsets of Data in Inference}}

\begin{figure}[H]
\centering
\subfigure[]
{
    \begin{minipage}[t]{0.27\linewidth}
	\centering
    \includegraphics[width = \linewidth]{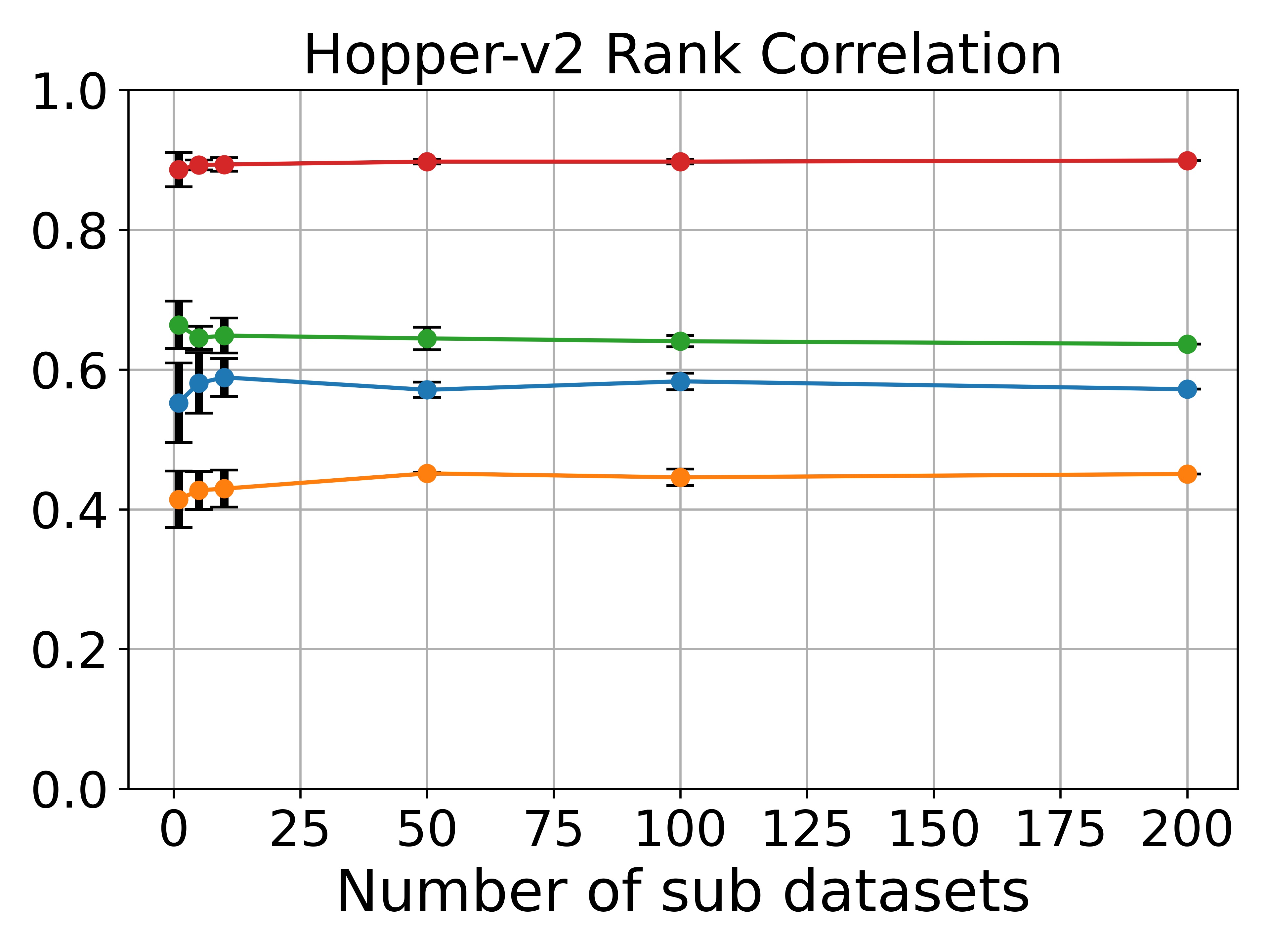}
    \end{minipage}
}
\subfigure[]
{
    \begin{minipage}[t]{0.27\linewidth}
	\centering
    \includegraphics[width = \linewidth]{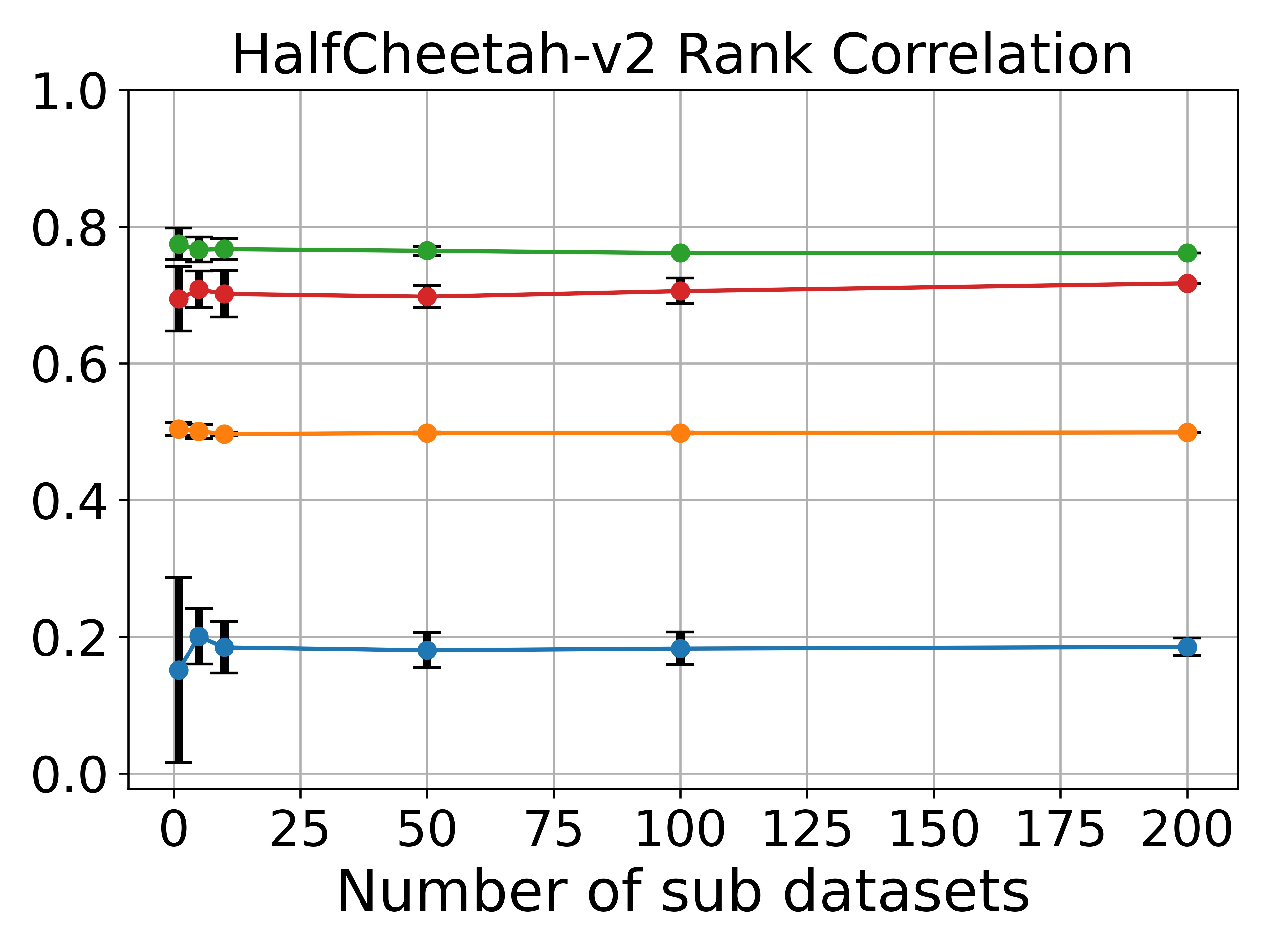}
    \end{minipage}
}
\subfigure[]
{
    \begin{minipage}[t]{0.27\linewidth}
	\centering
    \includegraphics[width = \linewidth]{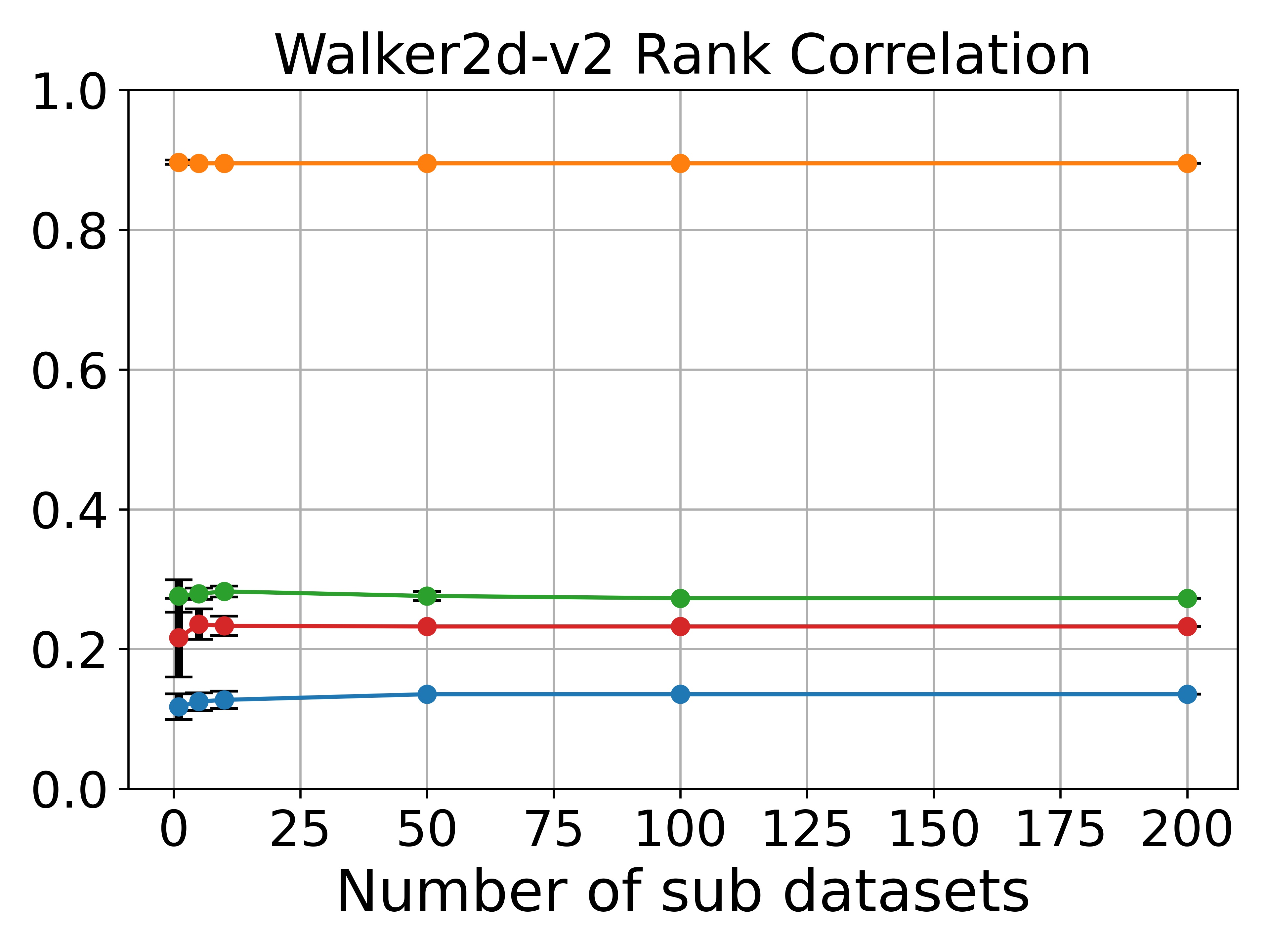}
    \end{minipage}
}

\subfigure[]
{
    \begin{minipage}[t]{0.27\linewidth}
	\centering
    \includegraphics[width = \linewidth]{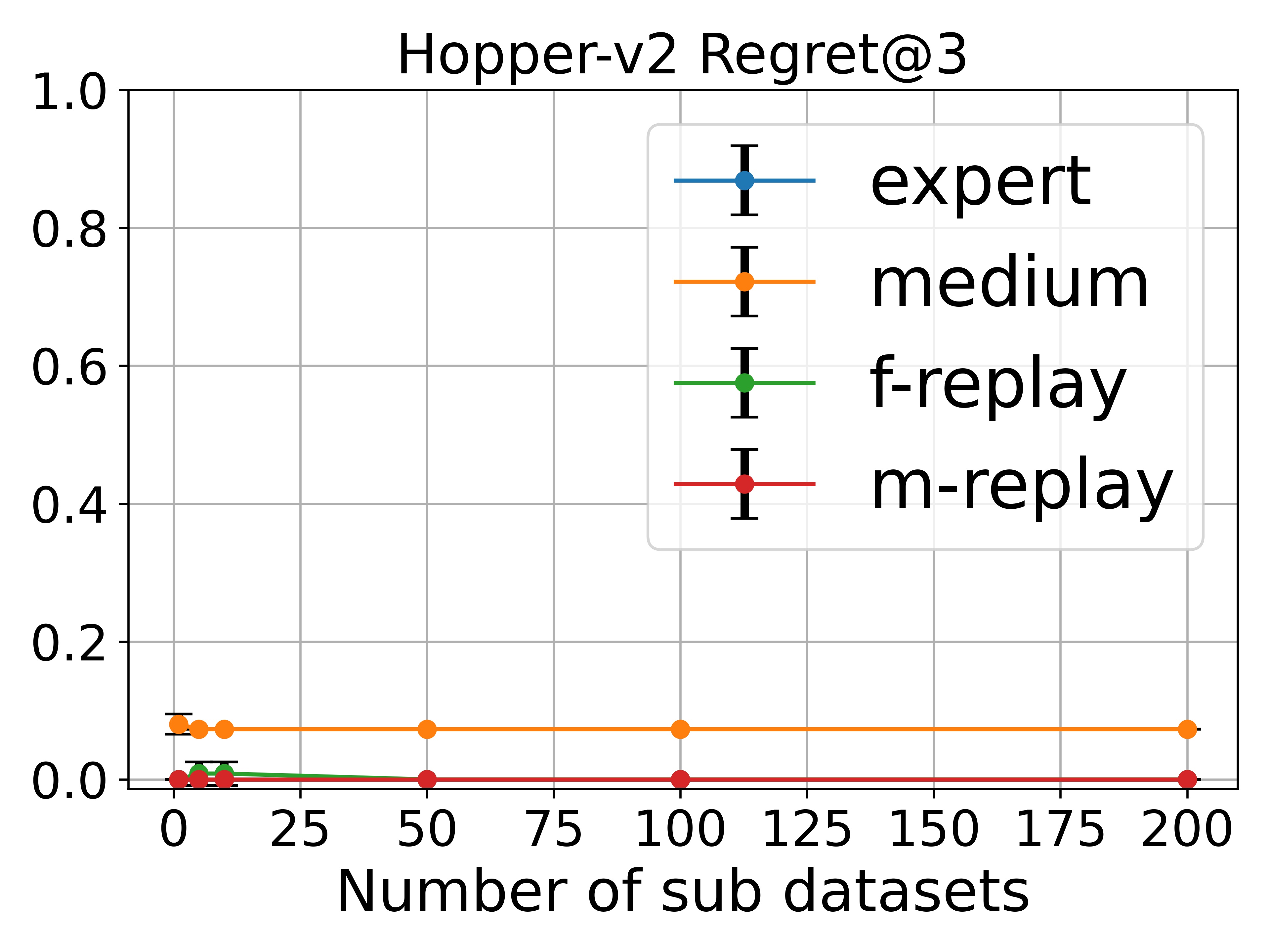}
    \end{minipage}
}
\subfigure[]
{
    \begin{minipage}[t]{0.27\linewidth}
	\centering
    \includegraphics[width = \linewidth]{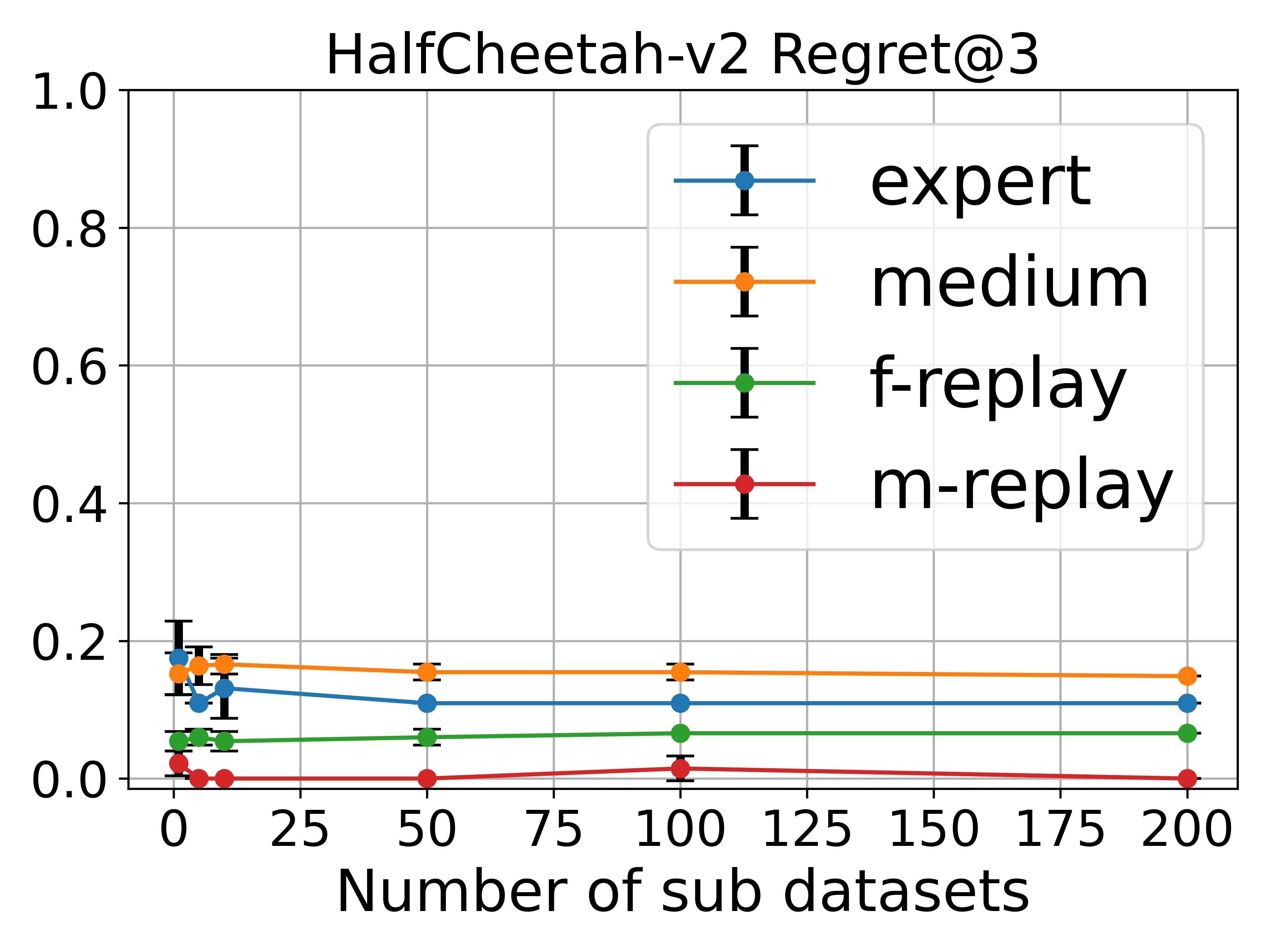}
    \end{minipage}
}
\subfigure[]
{
    \begin{minipage}[t]{0.27\linewidth}
	\centering
    \includegraphics[width = \linewidth]{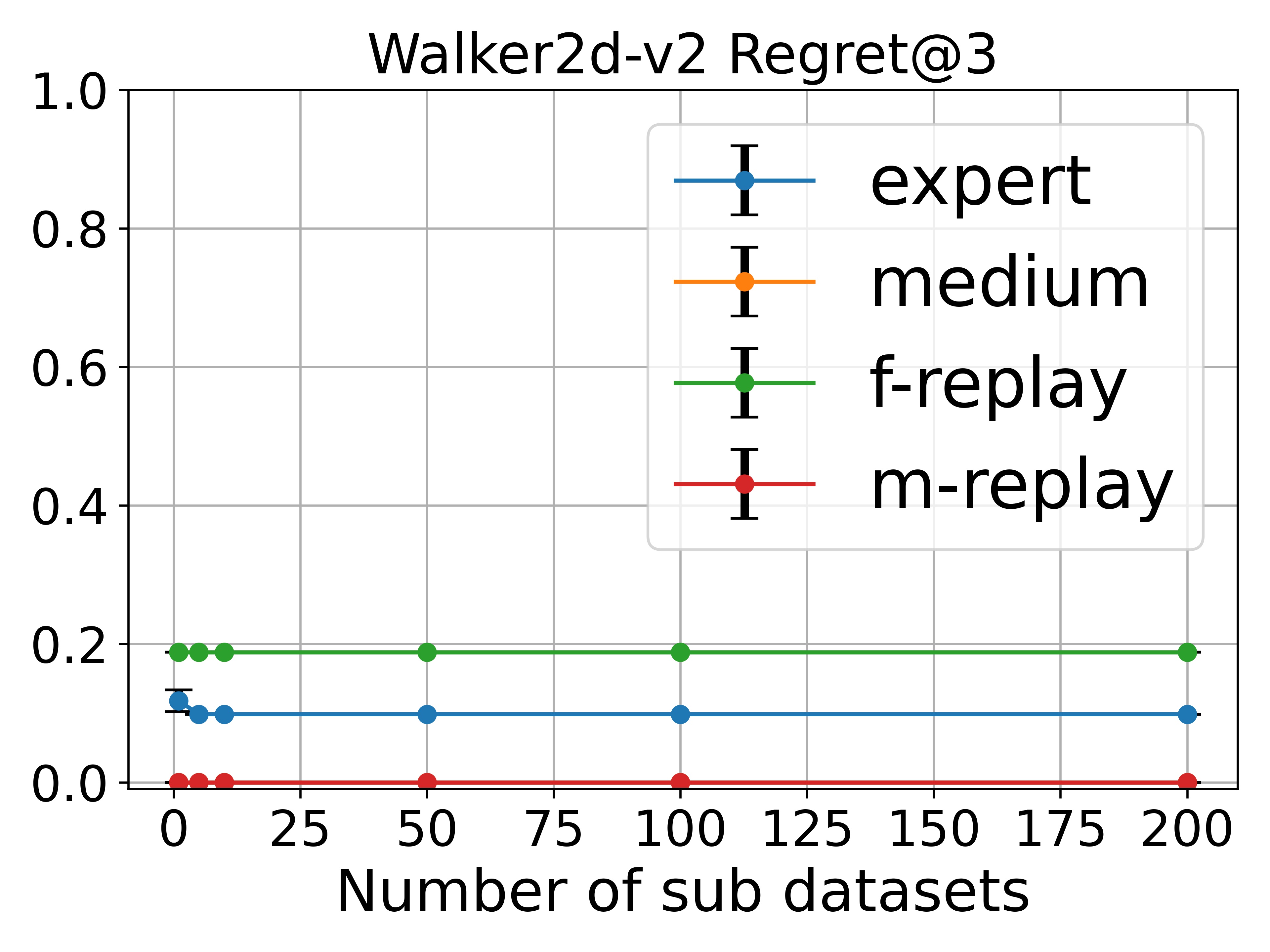}
    \end{minipage}
}
\caption {
Performance of SOPR-T with different numbers of subsets in inference.
Top row: rank correlation.
Bottom row: regret@3.
}
\label{5_3_supp_3.1}
\end{figure}

\begin{figure}[H]
\centering
\subfigure[]
{
    \begin{minipage}[t]{0.27\linewidth}
	\centering
    \includegraphics[width = \linewidth]{fig/Hopper_corr_var.jpg}
    \end{minipage}
}
\subfigure[]
{
    \begin{minipage}[t]{0.27\linewidth}
	\centering
    \includegraphics[width = \linewidth]{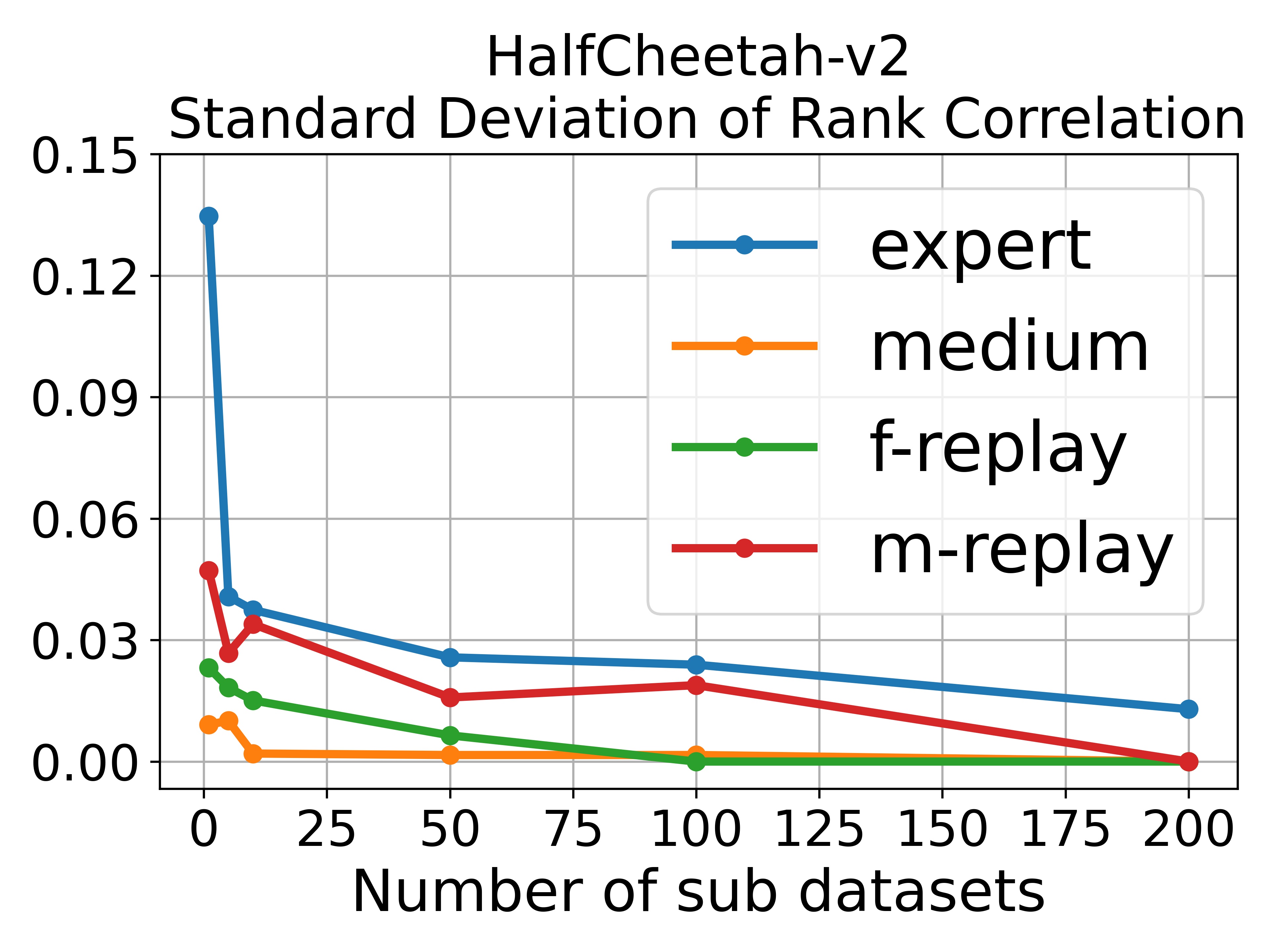}
    \end{minipage}
}
\subfigure[]
{
    \begin{minipage}[t]{0.27\linewidth}
	\centering
    \includegraphics[width = \linewidth]{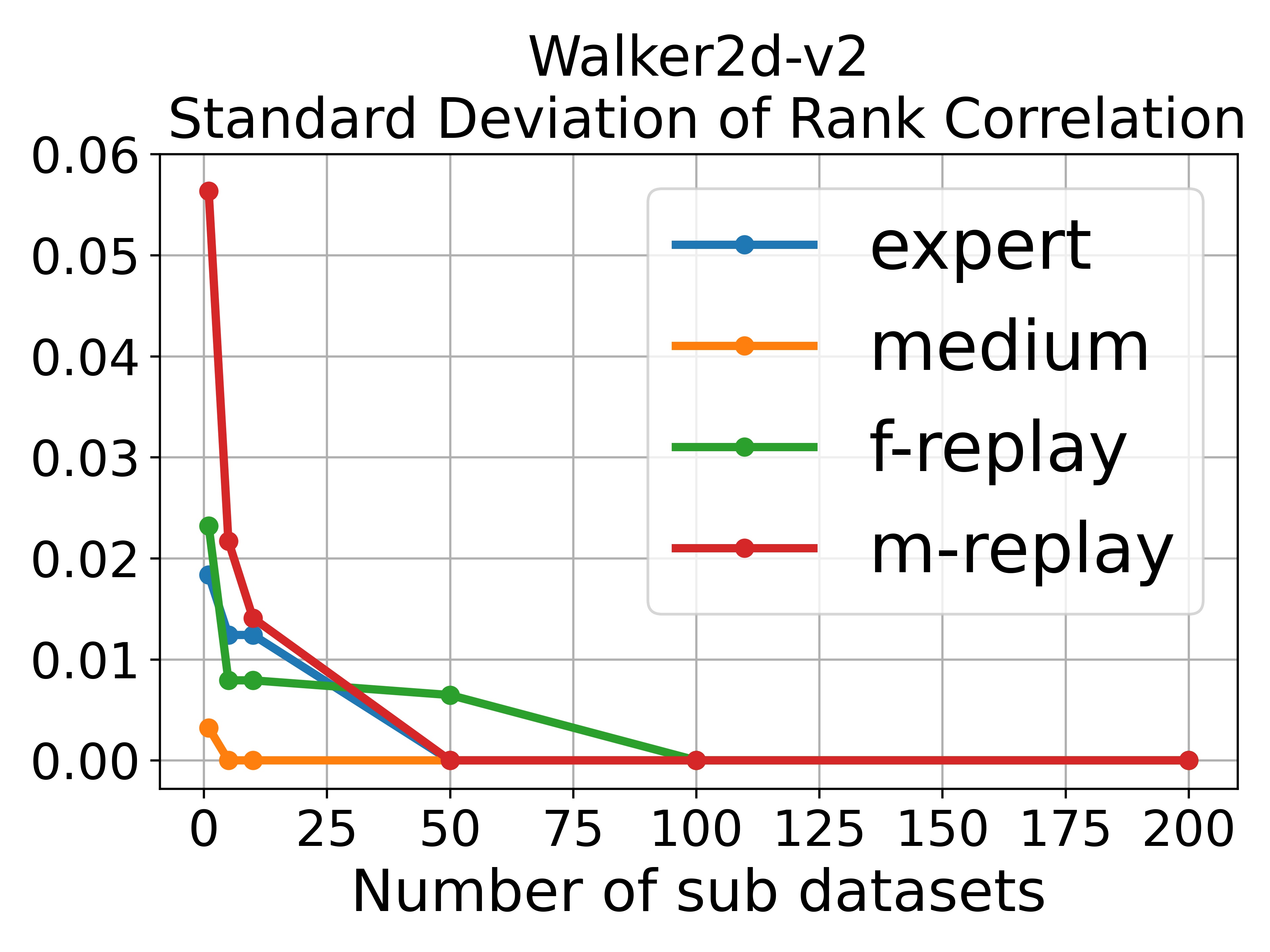}
    \end{minipage}
}

\subfigure[]
{
    \begin{minipage}[t]{0.27\linewidth}
	\centering
    \includegraphics[width = \linewidth]{fig/Hopper_regret_var.jpg}
    \end{minipage}
}
\subfigure[]
{
    \begin{minipage}[t]{0.27\linewidth}
	\centering
    \includegraphics[width = \linewidth]{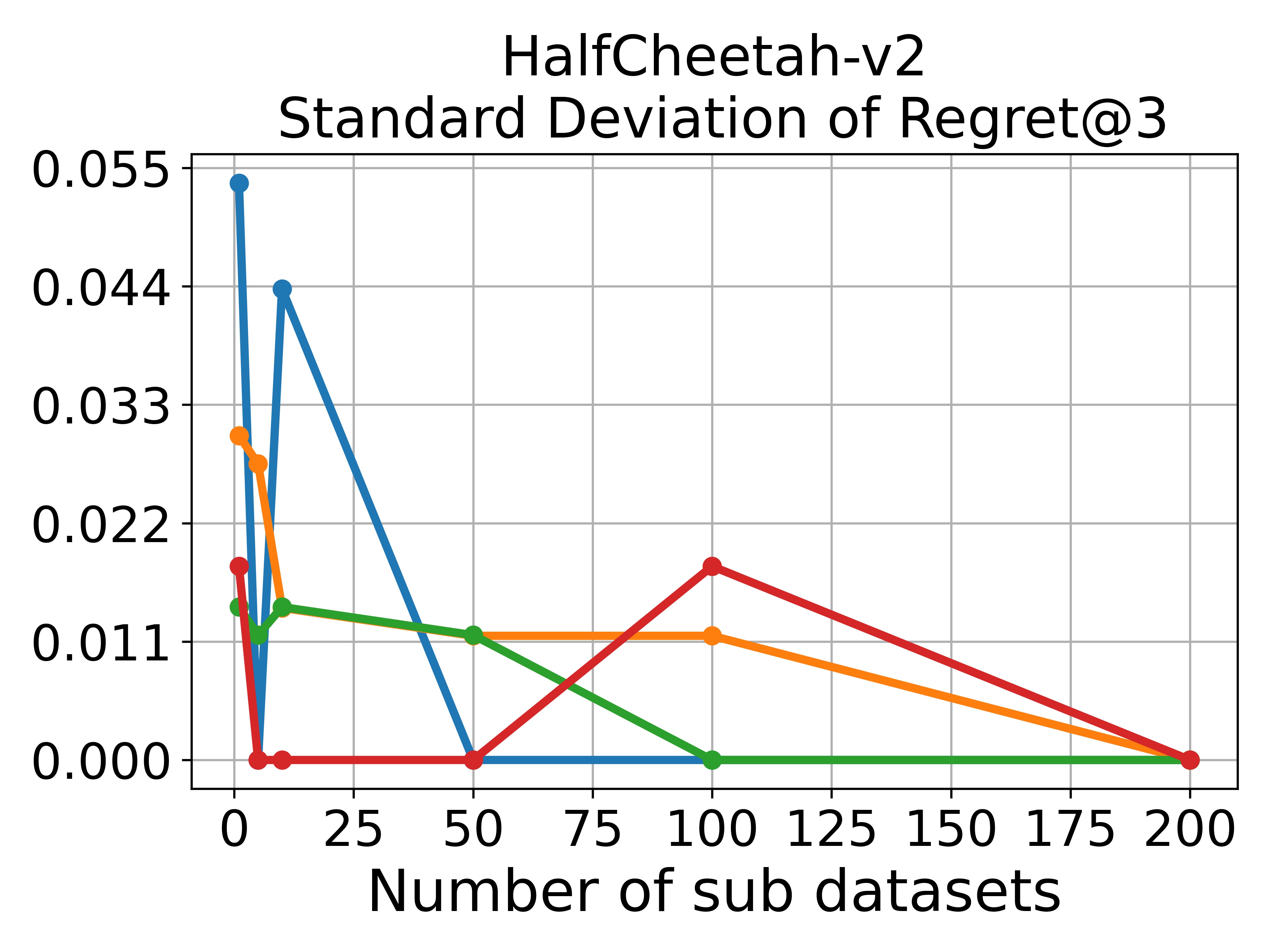}
    \end{minipage}
}
\subfigure[]
{
    \begin{minipage}[t]{0.27\linewidth}
	\centering
    \includegraphics[width = \linewidth]{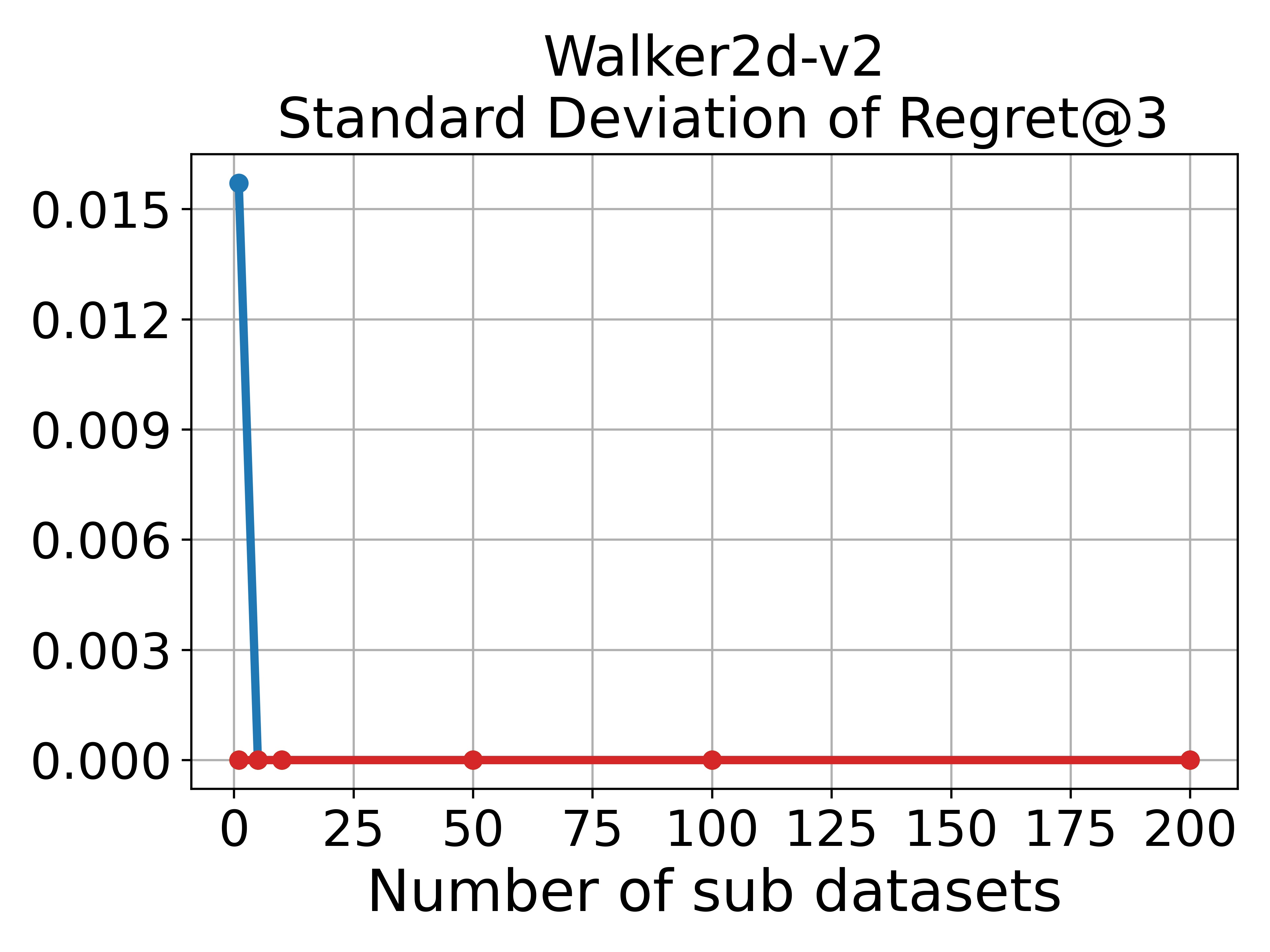}
    \end{minipage}
}
\caption {
Standard deviation of SOPR-T with different numbers of subsets in inference.
Top row: rank correlation.
Bottom row: regret@3.
}
\label{5_3_supp_3.2}
\end{figure}

In the last part of Section \ref{exp_further}, we investigated the performance variance of our SOPR-T algorithm and the effect of using different amounts of subsets in inference.
Here, Figure~\ref{5_3_supp_3.1} shows the average results with std bar on all the tasks with different amounts of subsets.
Figure~\ref{5_3_supp_3.2} shows the detailed standard deviation results.  

As can be seen from the results in Figure~\ref{5_3_supp_3.1}, the number of subsets makes little difference on the average performance of SOPR-T in all the tasks.
Figure~\ref{5_3_supp_3.2} indicates that the standard deviation decreases fast as the number of subsets increases.
In addition, in almost all the tasks, the standard deviation is very small.
Therefore, we can draw a conclusion that the performance of SOPR-T is stable even though it only uses small amount of data in inference.
Further, the inference time cost can be reduced by using a small number of subsets in practice.

\end{document}